\documentclass{article}

\usepackage{microtype}
\usepackage{graphicx}
\usepackage{subfig}
\usepackage{booktabs}

\usepackage{amsmath}
\usepackage{amssymb}
\DeclareMathOperator{\argmin}{argmin}
\DeclareMathOperator{\argmax}{argmax}
\usepackage{bbm}
\usepackage{mathtools}
\usepackage{cuted}
\usepackage{natbib}

\newcommand{\CV}{\text{CV}}
\newcommand{\DI}{\text{DI}}
\newcommand{\FNR}{\text{FNR}}
\usepackage{hyperref}

 \usepackage[accepted]{icml2020}

\icmltitlerunning{Accuracy and Fairness Trade-offs in Machine Learning: A Stochastic Multi-Objective Approach}

\begin{document}

\twocolumn[
\icmltitle{Accuracy and Fairness Trade-offs in Machine Learning: \\A Stochastic Multi-Objective Approach}

\icmlsetsymbol{equal}{*}

\begin{icmlauthorlist}
\icmlauthor{Suyun Liu}{equal,instit1}
\icmlauthor{Luis Nunes Vicente}{equal,instit1,instit2}
\end{icmlauthorlist}

\icmlaffiliation{instit1}{Department of Industrial and Systems Engineering, Lehigh University, Bethlehem, PA 18015, USA.}
\icmlaffiliation{instit2}{Support for
this author was partially provided by the Centre for Mathematics of the University of Coimbra under grant FCT/MCTES UIDB/MAT/00324/2020}

\icmlcorrespondingauthor{Suyun Liu}{sul217@lehigh.edu}
\icmlcorrespondingauthor{Luis Nunes Vicente}{lnv@lehigh.edu}



\vskip 0.3in
]

\printAffiliationsAndNotice{}

\begin{abstract}
In the application of machine learning to real-life decision-making systems, e.g., credit scoring and criminal justice, the prediction outcomes might discriminate against people with sensitive attributes, leading to unfairness. The commonly used strategy in fair machine learning is to include fairness as a constraint or a penalization term in the minimization of the prediction loss, which ultimately limits the information given to decision-makers. In this paper, we introduce a new approach to handle fairness by formulating a stochastic multi-objective optimization problem for which the corresponding Pareto fronts uniquely and comprehensively define the accuracy-fairness trade-offs. We have then applied a stochastic approximation-type method to efficiently obtain well-spread and accurate Pareto fronts, and by doing so we can handle training data arriving in a streaming way.
\end{abstract}

\section{Introduction}
\label{introduction}
Machine learning (ML) plays an increasingly significant role in data-driven decision making, e.g., credit scoring, college admission, hiring decisions, and criminal justice. As the learning models became more and more sophisticated, concern regarding fairness started receiving more and more attention. In 2014, the Obama Administration's Big Data Report~\cite{Executive_Office_2014} claimed that discrimination against individuals and groups might be the ``inadvertent outcome of the way big data technologies are structured and used''. Two years later, a White House report~\yrcite{Executive_Office_2016} on the challenges of big data emphasized the necessity of promoting fairness and called for equal opportunity in insurance, education, employment, and other sectors.

In supervised machine learning, training samples consist of pairs of feature vectors (containing a number of features that are descriptive of each instance) and target values/labels. One tries to determine an accurate predictor, seen as a function mapping feature vectors into target labels. Such a predictor is typically characterized by a number of parameters, and the process of identifying the optimal parameters is called training or learning. The trained predictor can then be used to predict labels for unlabeled instances.

If a ML predictor does inequitably treat people from different groups defined by \textit{sensitive} or \textit{protected} attributes, such as gender, race, country, or disability, we say that such a predictor is \textit{unfair}.
The sources of unfairness in supervised ML are twofold. Firstly, the ML predictors are trained on data collected by humans (or automated agents developed by humans), which may contain inherent biases. Hence, by learning from biased or prejudiced targets, the prediction results obtained from standard learning processes can hardly be unbiased. Secondly, even if the targets are unbiased, the learning process may sacrifice fairness, as the main goal of ML is to make predictions as accurate as possible. In fact, previous research work~\cite{DPedreshi_SRuggieri_FTurini_2008, RZemel_etal_2013} has showed that simply excluding sensitive attributes from features data (also called \textit{fairness through unawareness}) does not help due to the fact that the sensitive attributes can be inferred from the nonsensitive ones.

Hence, a proper framework for evaluating and promoting fairness in ML becomes indispensable and relevant.
Depending on when the fairness criteria are imposed, there are three categories of approaches proposed to handle fairness, namely pre-processing, in-training, and post-processing.
Pre-processing approaches~\cite{FCalmon_etal_2017, RZemel_etal_2013} modify the input data representation so that the prediction outcomes from any standard learning process become fair, while post-processing~\cite{MHardt_EPrice_NSrebro_2016, GPleiss_etal_2017} tries to adjust the results of a pre-trained predictor to increase fairness while maintaining the prediction accuracy as much as possible.
Assuming that the sensitive attributes information are accessible in the training samples,
most of in-training methods~\cite{SBarocas_ADSelbst_2016, TCalders_FKamiran_MPechenizkiy_2009, TKamishima_SAkaho_JSakuma_2011, MBZafar_etal_2017b, BWoodworth_etal_2017, MBZafar_etal_2017, ANavon_etal_2021} enforce fairness during the training process either by directly imposing fairness constraints and solving constrained optimization problems or by adding penalization terms to the learning objective.

The approach proposed in our paper falls into the in-training category. We will however explicitly recognize the presence of at least two conflicting objectives in fair machine learning: (1) maximizing prediction accuracy; (2) maximizing fairness (w.r.t. certain sensitive attributes).  

Multi-objective optimization methodology has been applied to various machine learning models to tackle conflicting objectives. In neural network training, a trade-off between accuracy and complexity were naturally formulated as a bi-objective problem and solved to get the Pareto front using evolutionary optimization methods~\cite{JPTYusiong_PCNaval_2006, SKaoutar_EMohamed_2017, Senhaji_etal_2017, KSenhaji_HRamchoun_MEttaouil_2020} or scalarization-based optimization methods~\cite{IKokshenev_APBraga_2008, MReiners_KKlamroth_MStiglmayr_2020}. Similarly, two competing goals were constructed to minimize the training loss and minimize the model capacity in support vector machines~\cite{JBi_2003, CIgel_2005} and decision trees~\cite{DKim_2004}. The obtained Pareto fronts supported down-streaming tasks like feature selection and model selection. Beyond that, conflicting objectives are found in
multi-task learning~\cite{OSener_VKoltun_2018, NVVarghese_QHMahmoud_2020, YZhang_QYang_2021} and data clustering~\cite{JHandl_JKnowles_2004, Law_etal_2014}. More multi-objective training frameworks were reviewed by~\cite{YJin_2006, ABraga_etal_2006, YJin_BSendhoff_2008, Alexandropoulos_etal_2019}. The conflict between accuracy and fairness is a new research topic in multi-objective machine learning. To the best of our knowledge, only a limited number of frameworks~\cite{MBZafar_etal_2017, MBZafar_etal_2017b, ANavon_etal_2021} were designed towards computing a substantial part of the entire trade-off curve.

\subsection{Existing Fairness Criteria in Machine Learning}
\label{related_work}
Fairness in machine learning basically requires that prediction outcomes do not disproportionally benefit people from majority and minority or historically advantageous and disadvantageous groups. In the literature of fair machine learning, several prevailing criteria for fairness include~\textit{disparate impact}~\cite{SBarocas_ADSelbst_2016} (also called~\textit{demographic parity}~\cite{TCalders_FKamiran_MPechenizkiy_2009}), \textit{equalized odds}~\cite{MHardt_EPrice_NSrebro_2016}, and its special case of \textit{equal opportunity}~\cite{MHardt_EPrice_NSrebro_2016}, corresponding to different aspects of fairness requirements.

In this paper, we will focus on \textit{binary classification} to present the formula for fairness criteria and the proposed accuracy and fairness trade-off framework, although they can all be easily generalized to other ML problems (such as regression or clustering). We point out that
many real decision-making problems such as college admission, bank loan application, hiring decisions, etc. can be formulated into binary classification models.

Let $Z \in \mathbb{R}^n, A \in \{0, 1\}, Y \in \{-1, + 1\}$ denote feature vector, binary-valued sensitive attribute (for simplicity we focus on the case of a single binary sensitive attribute), and target label respectively. Consider a general predictor $\hat{Y} \in \{-1, + 1\}$ which could be a function of both $Z$ and $A$ or only $Z$.  The predictor is free of disparate impact~\cite{SBarocas_ADSelbst_2016} if the prediction outcome is statistically independent of the sensitive attribute, i.e., for $\hat{y} \in \{-1, +1\}$,
\begin{equation}
\label{def_disparate_impact}
\mathbb{P}\{\hat{Y} = \hat{y} | A = 0\} \;=\; \mathbb{P}\{\hat{Y} = \hat{y} | A = 1\}.
\end{equation}
However, disparate impact could be unrealistic when one group is more likely to be classified as a positive class than others, an example being that women are more dominating in education and healthcare services than men~\cite{JKelly_forbes_2020}. As a result, disparate impact may never be aligned with a perfect predictor~$\hat{Y} = Y$.

In terms of equalized odds~\cite{MHardt_EPrice_NSrebro_2016}, the predictor is defined to be fair if it is independent of the sensitive attribute but conditioning on the true outcome $Y$, namely for $y, \hat{y} \in \{-1, +1\}$,
\begin{equation}
\label{def_equal_odd}
\small
    \mathbb{P}\{\hat{Y} = \hat{y} | A = 0, Y = y\} \;=\; \mathbb{P}\{\hat{Y} = \hat{y} | A = 1, Y = y\}.
\end{equation}
Under this definition, a perfectly accurate predictor can be possibly defined as a fair one, as the probabilities in~(\ref{def_equal_odd}) will always coincide when $\hat{Y} = Y$. Equal opportunity~\cite{MHardt_EPrice_NSrebro_2016}, a relaxed version of equalized odds, requires that condition~\eqref{def_equal_odd} holds for only positive outcome instances ($Y = +1$), for example, students admitted to a college and candidates hired by a company.

\subsection{Our Contribution}
From the perspective of multi-objective optimization (MOO), most of the in-training methods in the literature~\cite{SBarocas_ADSelbst_2016, TCalders_FKamiran_MPechenizkiy_2009, TKamishima_SAkaho_JSakuma_2011, BWoodworth_etal_2017, MBZafar_etal_2017, MBZafar_etal_2017b} are based on the so-called \textit{a priori} methodology, where the decision-making preference regarding an objective (the level of fairness) must be specified before optimizing the other (the accuracy). For instance, the constrained optimization problems proposed in~\cite{MBZafar_etal_2017, MBZafar_etal_2017b} are to some extent nothing else than the $\epsilon$--constraint method~\cite{YVHaimes_1971} in MOO. Such procedures highly rely on the decision-maker's advanced \linebreak knowledge of the magnitude of fairness, which may vary from criterion to criterion and from dataset to dataset.

In order to better frame our discussion of accuracy vs fairness, let us introduce
the general form of a multi-objective optimization problem
\begin{equation}
    \begin{array}{rl}
     \min & F(x) \;=\; (f_1(x), \ldots, f_m(x)), \\ [0.5ex]
\end{array}
\label{smoo_form}
\end{equation}
with $m$ objectives, and
where $F: \mathbb{R}^n \rightarrow \mathbb{R}^m$. Usually, there is no single point optimizing all the objectives simultaneously. The notion of \textit{dominance} is used to define optimality in MOO. A point~$y$ (weakly) dominates~$x$ if 
$F(y) \leq F(x)$ holds element-wise and additionally $f_i(y) < f_i(x)$ holds for at least one objective $i = 1, \ldots, m$. A point~$x$ is said to be (strictly) nondominated if it is not (weakly) dominated by any other point~$y$. An unambiguous way of considering the trade-offs among multiple objectives is given by the so-called \textit{Pareto front}, which lies in the criteria space $\mathbb{R}^m$ and is defined as the set of points of the form $F(x)$ for all nondominated points~$x$. The concept of dominance is carried out from the decision space to the image one, and one can also say that $F(y)$ (strictly) dominates $F(x)$ when $F(y) \leq F(x)$ holds element-wise and $f_i(y) < f_i(y)$ holds for at least one $i = 1, \ldots, m$.  Finally,  the terminology {\it trade-off} refers to compromising and balancing among the set of equally good nondominated solutions.

In this paper, instead of looking for a single predictor that satisfies certain fairness constraints, our goal is to directly construct a complete Pareto front between prediction accuracy and fairness, and thus to identify a set of predictors associated with different levels of fairness. We propose a stochastic multi-objective optimization framework, and aim at obtaining good approximations of true Pareto fronts. We summarize below the three main advantages of the proposed framework.
\begin{itemize}
    \item By applying an algorithm for stochastic multi-objective optimization
    (such as the Pareto front stochastic multi-gradient (PF-SMG) algorithm developed in~\cite{SLiu_LNVicente_2019}), we are able to obtain well-spread and accurate Pareto fronts in a flexible and efficient way. The approach works for a variety of scenarios, including binary and categorical multi-valued sensitive attributes. It also handles multiple objectives simultaneously, such as multiple sensitive attributes and multiple fairness measures. Compared to the constrained optimization approaches, e.g.,~\cite{MBZafar_etal_2017, MBZafar_etal_2017b}, our framework is proved to be computational efficient in constructing the whole Pareto fronts.
    \item The proposed framework is quite general in the sense that it has no restriction on the type of predictors and works for any convex or nonconvex smooth objective functions. In fact, it can not only handle the fairness criteria mentioned in Section~\ref{related_work} based on covariance approximation, but also tackle other formula proposed in the literature, e.g., mutual information~\cite{TKamishima_etal_2012} and fairness as a risk measure~\cite{RCWilliamson_AKMenon_2019}.
    \item The PF-SMG algorithm falls into a Stochastic Approximation (SA) algorithmic approach, and thus it enables us to deal with the case where the training data is arriving on a streaming mode. By using such an SA framework, there is no need to reconstruct the Pareto front from scratch each time new data arrives. Instead, a Pareto front constructed based on consecutive arriving samples will eventually converge to the one corresponding to the overall true population.
\end{itemize}

By combining existing optimization methods and fairness concepts, this paper offers solutions to open ML fairness problems (non-convex measures, streaming data, multiple attributes, and multiple fairness measures). Our work provides a systematic and thorough trade-off approach from the viewpoint of stochastic multi-objective optimization for ML fairness.

The remainder of this paper is organized as follows. Our stochastic bi-objective formulation using disparate impact is suggested in Section~\ref{biobjective_approach}.
The PF-SMG algorithm, used to solve the multi-objective problems, is briefly introduced in Section~\ref{pfsmg_review} (more details in Appendix~\ref{appendix:PF_SMG}). A number of numerical results for both synthetic and real data are presented in Section~\ref{num_results} to support our claims.
Further exploring our line of thought, we introduce another stochastic bi-objective formulation, this time for trading-off accuracy vs equal opportunity (see Section~\ref{equal_opp}), also reporting numerical results. In Section~\ref{multi_attributes_measures}, we show how to handle multiple sensitive attributes and multiple fairness measures. For the purpose of getting more insight on the various trade-offs, two tri-objective problems are formulated and solved. Finally, a preliminary numerical experiment described in Section~\ref{stream_data} will illustrate the applicability of our approach to streaming data. The paper is ended with some conclusions and prospects of future work in Section~\ref{sec:conclusions}.

\section{The Stochastic Bi-Objective Formulation Using Disparate Impact}
\label{biobjective_approach}
Given that disparate impact is the most commonly used fairness criterion in the literature, we will first consider disparate impact in this section to present a stochastic bi-objective fairness and accuracy trade-off framework.

In our setting, the training samples consist of nonsensitive feature vectors $Z$, a binary sensitive attribute $A$, and binary labels $Y$. Assume that we have access to $N$ samples $\{z_j, a_j, y_j\}_{j = 1}^N$ from a given database. Let the binary predictor $\hat{Y} = \hat{Y}(Z; x) \in \{-1, +1\}$ be a function of the parameters $x$, and only learned from the nonsensitive feature~$Z$.

Recall that the predictor $\hat{Y}$ is free of disparate impact if it satisfies equation~\eqref{def_disparate_impact}.
A general measurement of disparate impact, the so-called CV score~\cite{TCalders_SVerwer_2010}, is defined by the maximum gap between the probabilities of getting positive outcomes in different sensitive groups, i.e., 
\begin{equation} \label{CVscore}
    \CV(\hat{Y}) \;=\; |\mathbb{P}\{\hat{Y} = 1 | A = 0\} - \mathbb{P}\{\hat{Y} = 1| A = 1\}|.
\end{equation}
The trade-offs between prediction accuracy and fairness can then be formulated as a general stochastic bi-objective optimization problem as follows
\begin{align}
     \min~f_1(x) \;=\; & \mathbb{E}[\ell(\hat{Y}(Z; x), Y)], \label{obj1_loss} \\
    \min~f_2(x) \;=\; & \CV(\hat{Y}(Z; x)),
    \label{obj2_fairness}
\end{align}
where the first objective~\eqref{obj1_loss} is a composition function of a loss function $\ell(\cdot,\cdot)$ and the prediction function $\hat{Y}(Z; x)$, and the expectation is taken over the joint distribution of $Z$ and $Y$.

The logistic regression model is one of the classical prediction models for binary classification problems. 
For purposes of binary classification, one aims at determining a linear classifier $\phi(z; x) = \phi(z; c, b) = c^\top z + b$ (noting $x=(c,b)^\top$)
in order to minimize a certain prediction loss.
The data would be perfectly separated if $c^\top z_j + b \geq 0$ when $y_j = +1$, and $c^\top z_j + b  < 0$ when $y_j = - 1$ for all $(z_j,y_j)$ pairs.
The classical $0$--$1$ loss function is given by
$\mathbbm{1}( \hat{Y}(z; c, b) \neq y)$,
where
$\hat{Y}(z; c, b) = 2\times\mathbbm{1}(c^\top z + b \geq 0) - 1$. The logistic loss function is of the form $\ell(z, y; c, b) = \text{log}(1+\text{exp}(-y(c^\top z + b)))$, and is a smooth and convex version of the $0$--$1$ loss. The first objective can then be approximated by the empirical logistic regression loss, i.e.,
\begin{equation}
\textstyle
    f_1(c, b) \;=\; \tfrac{1}{N}\sum_{j=1}^N \text{log}(1+\text{exp}(-y_j(c^\top z_j + b))),
    \label{obj1_empirical_loss}
\end{equation}
based on $N$ training samples. A regularization term $\frac{\lambda}{2}\|c\|^2$ can be added to avoid over-fitting.

Dealing with the second objective~\eqref{obj2_fairness} is challenging since it is nonsmooth and nonconvex.  Hence, we make use of the \textit{decision boundary covariance} proposed by~\cite{MBZafar_etal_2017b} as a convex approximate measurement of disparate impact. Specifically, the CV score~(\ref{CVscore}) can be approximated by the empirical covariance between the sensitive attributes $A$ and the hyperplane $\phi(Z; c, b)$, i.e.,
\begin{equation*}
\small
\begin{array}{ll}
    & \mbox{Cov}(A, \phi(Z; c, b))  \\
    & \;=\;   \mathbb{E}[(A - \bar{A})(\phi(Z; c, b) - \overline{\phi(Z; c, b)})] \\
     &  \;= \;  \mathbb{E}[(A - \bar{A})\phi(Z; c, b)] - \mathbb{E}[A - \bar{A}]\overline{\phi (Z; c, b)}  \\
     & \;\simeq\; \tfrac{1}{N}\textstyle\sum_{j = 1}^N (a_j - \bar{a})\phi(z_j; c, b),
\end{array}
\label{Approx_def_disparate_impact}
\end{equation*}
where $\bar{A}$ is the expected value of the sensitive attribute, and $\bar{a}$ is an approximated value of~$\bar{A}$ using $N$ samples. The intuition behind this approximation is that the disparate impact~\eqref{def_disparate_impact} basically requires the predictor completely independent from the sensitive attribute.

Given that zero covariance is a necessary condition for independence, the second objective can be approximated as:
\begin{equation}
\textstyle
f_2^\DI(c, b) \;=\; \left(\frac{1}{N}\sum_{j=1}^N (a_j - \bar{a})(c^\top z_j + b)\right)^2, \label{obj2_Approx_fairness}
\end{equation}
which, as we will see later in the paper, is monotonically increasing with disparate impact. We were thus able to construct a finite-sum bi-objective problem
\begin{equation}
    \min \; \left(f_1(c, b), f_2^\DI(c, b)\right),
    \label{bi_obj_DI_binary}
\end{equation}
where both functions are now convex and smooth.

\section{The Stochastic Multi-Gradient Method and Its Pareto Front Version}
\label{pfsmg_review}
Consider again a stochastic MOO of the same form as in~\eqref{smoo_form}, where some or all of the objectives involve uncertainty. Denote by $g_i(x, w)$ a stochastic gradient of the $i$-th objective function, where $w$ indicates the batch of samples used in the estimation. The stochastic multi-gradient (SMG) algorithm is described in Algorithm~\ref{alg:SMG} (see Appendix~\ref{appendix:SMG}). It essentially takes a
step along the stochastic multi-gradient $g(x_k, w_k)$ which is a convex linear combination of
$g_i(x, w)$, $i=1,\ldots,m$. 
The SMG method is a generalization of stochastic gradient (SG) to multiple objectives. It was first proposed by~\cite{MQuentin_PFabrice_JADesideri_2018} and further analyzed by~\cite{SLiu_LNVicente_2019}. In the latter paper it was proved that the SMG algorithm has the same convergence rates as SG (although now to a nondominated point), for both convex and strongly convex objectives. As we said before, when $m=1$ SMG reduces to SG. When~$m>1$ and the~$f$'s are deterministic, $-g(x_k)=-g(x_k, w_k)$ is the direction that is the most descent among all the~$m$
functions~\cite{JFliege_BFSvaiter_2000, JFliege_AVaz_LNVicente_2019}.

Note that the two smooth objective functions~\eqref{obj1_empirical_loss} and~\eqref{obj2_Approx_fairness} are both given in a finite-sum form, for which one can efficiently compute stochastic gradients using batches of samples.

To compute good approximations of the entire Pareto front in a single run, we use the Pareto Front SMG algorithm (PF-SMG) developed by~\cite{SLiu_LNVicente_2019}
(see Appendix~\ref{appendix:PF_SMG} for an algorithm description). PF-SMG essentially maintains a list of nondominated points using SMG updates.  It solves stochastic multi-objective problems in an \textit{a posteriori} way, by determining Pareto fronts without predefining weights or adjusting levels of preferences.
One starts with an initial list of randomly generated points ($5$ in our experiments).  

At each iteration of PF-SMG, we apply SMG multiple times at each point in the current list, and by doing so one obtains different final points due to stochasticity.
At the end of each iteration, all the dominated points are removed to get a new list for the next iteration (see also Appendix~\ref{appendix:PF_SMG} for an illustration of PF-SMG). The process can be stopped when either the number of nondominated points is greater than a certain budget (1,500 in our experiments) or when the total number of SMG iterates applied in any trajectory exceeds a certain budget (1,000 in our experiments). We refer to the paper~\cite{SLiu_LNVicente_2019} for more details.

\section{Numerical Results for Disparate Impact}
\label{num_results}
\subsection{Experiment setup}
To numerically illustrate our approach based on the bi-objective formulation~\eqref{bi_obj_DI_binary}, we have used both the \textit{Adult Income} dataset~\cite{RKohavi_1996}, which is available in the UCI Machine Learning Repository~\cite{DDua_and_CGraff_2017} and
synthetic data (see Appendix~\ref{appendix_datasets} for the data generation).

There are several parameters to be tuned in PF-SMG for a better performance: (1) $p_1$: number of times SMG is applied at each point in the current list; (2) $p_2$: number of SMG iterations each time SMG is called; (3) $\{\alpha_k\}_1^{T}$: step size sequence; (4) $\{b_{1, k}\}_1^{T}, \{b_{2, k}\}_1^{T}$: batch size sequences used in computing stochastic gradients for the two objectives. To control the rate of generated nondominated points, we remove nondominated points from regions where such points tend to grow too densely\footnote{Our implementation code is available at~\url{https://github.com/sul217/MOO_Fairness}.}.  In our experiments, the parameters are selected using grid search. Specifically, for each combination of parameters, we run the PF-SMG using training data to get a set of nondominated solutions. The best parameter combination is chosen as the one with the most accurate predictor on the Pareto fronts which are evaluated using a set of validation samples.

Our approach is compared to the $\epsilon$-constrained optimization model proposed in~\citet[Equation~(4)]{MBZafar_etal_2017b}. From now on, we note their $\epsilon$-constrained method as~EPS-fair. It basically minimizes prediction loss subject to disparate impact being bounded above by a constant $\epsilon$, i.e.,
\begin{equation*}
\textstyle \min~\eqref{obj1_empirical_loss} \quad \mbox{s.t.}\quad |\tfrac{1}{N}\textstyle\sum_{j = 1}^N (a_j - \bar{a})\phi(z_j; c, b)| \leq \epsilon.
\end{equation*}
Since the bi-objective problem~\eqref{bi_obj_DI_binary} under investigation is convex, EPS-fair is able to compute a set of nondominated points by varying the value of~$\epsilon$. The implementation details of EPS-fair method can be found in~\cite{MBZafar_etal_2017b}. First, by solely minimizing prediction loss, a reasonable upper bound is obtained for disparate impact. Then, to obtain the Pareto front, a sequence of thresholds $\epsilon$ is evenly chosen from $0$ to such an upper bound, leading to a set of convex constrained optimization problems. The Sequential Least SQuares Programming (SLSQP) solver~\cite{DKraft_1988} based on Quasi-Newton methods is then used for solving those problems. We found that 70-80\% of the final points produced by this process were actually dominated ones, and we removed them for the purpose of analyzing results.

The cleaned up version of \textit{Adult Income} dataset contains 45,222 samples. Each instance is characterized by 12 nonsensitive attributes (including age, education, marital status, and occupation), a binary sensitive attribute (gender), and a multi-valued sensitive attribute (race). The prediction target is to determine whether a person makes over 50K per year. Tables~\ref{Adult_dataset_gender} and \ref{Adult_dataset_race} in Appendix~\ref{data_tables} show the detailed demographic composition of the dataset with respect to gender and race.

In the following experiments, we have randomly sampled 60\% of the whole dataset for training, using 10\% for validating and the remaining 30\% instances as the testing dataset. The PF-SMG algorithm is applied using the training dataset, but all the Pareto fronts and the corresponding trade-off information will be presented using the testing dataset.

\subsection{Numerical results} \label{subsec:real_data}

\begin{figure*}[ht]
  \centering
  \vskip 0cm
  \subfloat[Pareto front.]{\includegraphics[width = 0.23\textwidth]{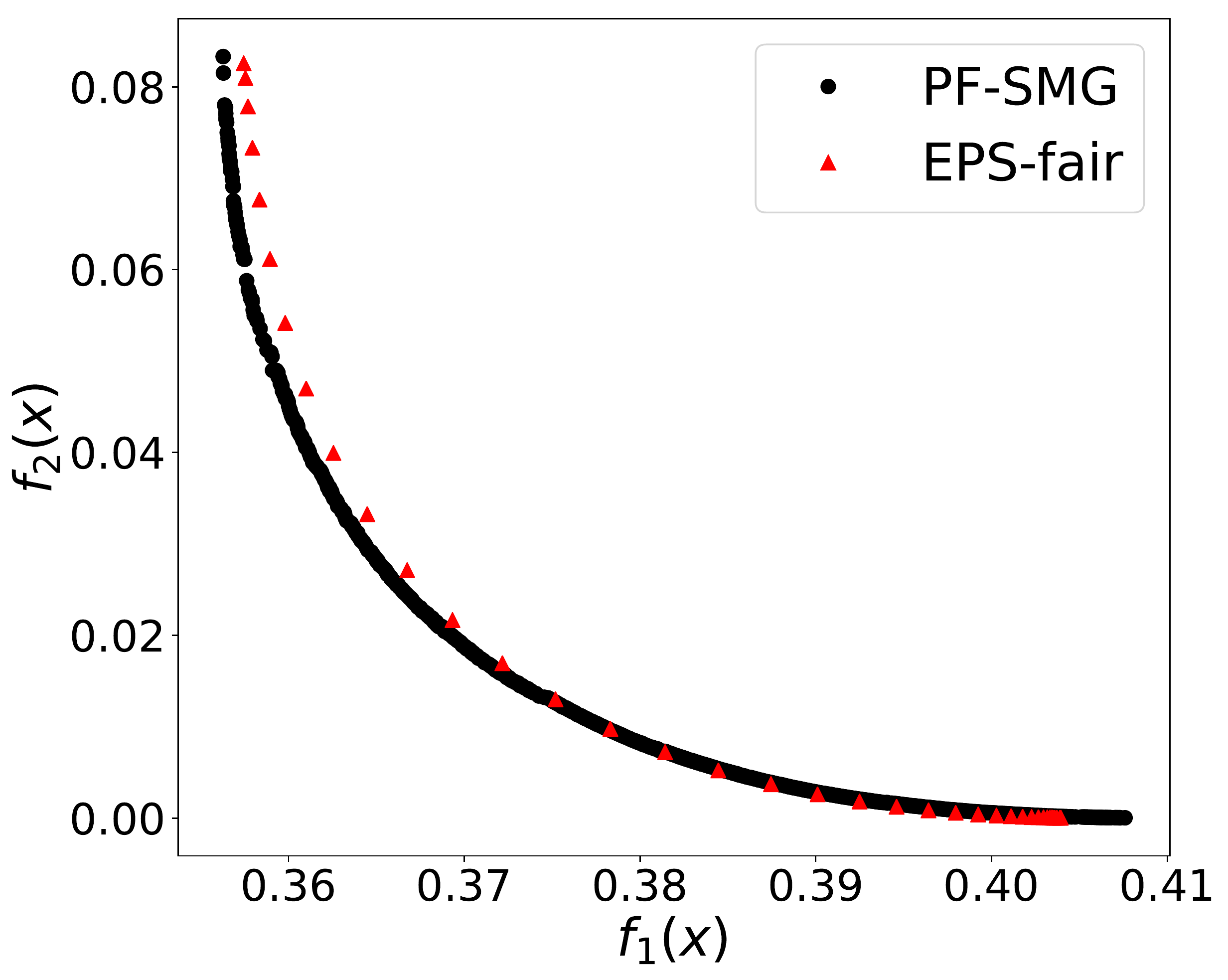}}
  \subfloat[$f_2^{\DI}(x)$ vs \%pos. outcomes.]{\includegraphics[width = 0.23\textwidth]{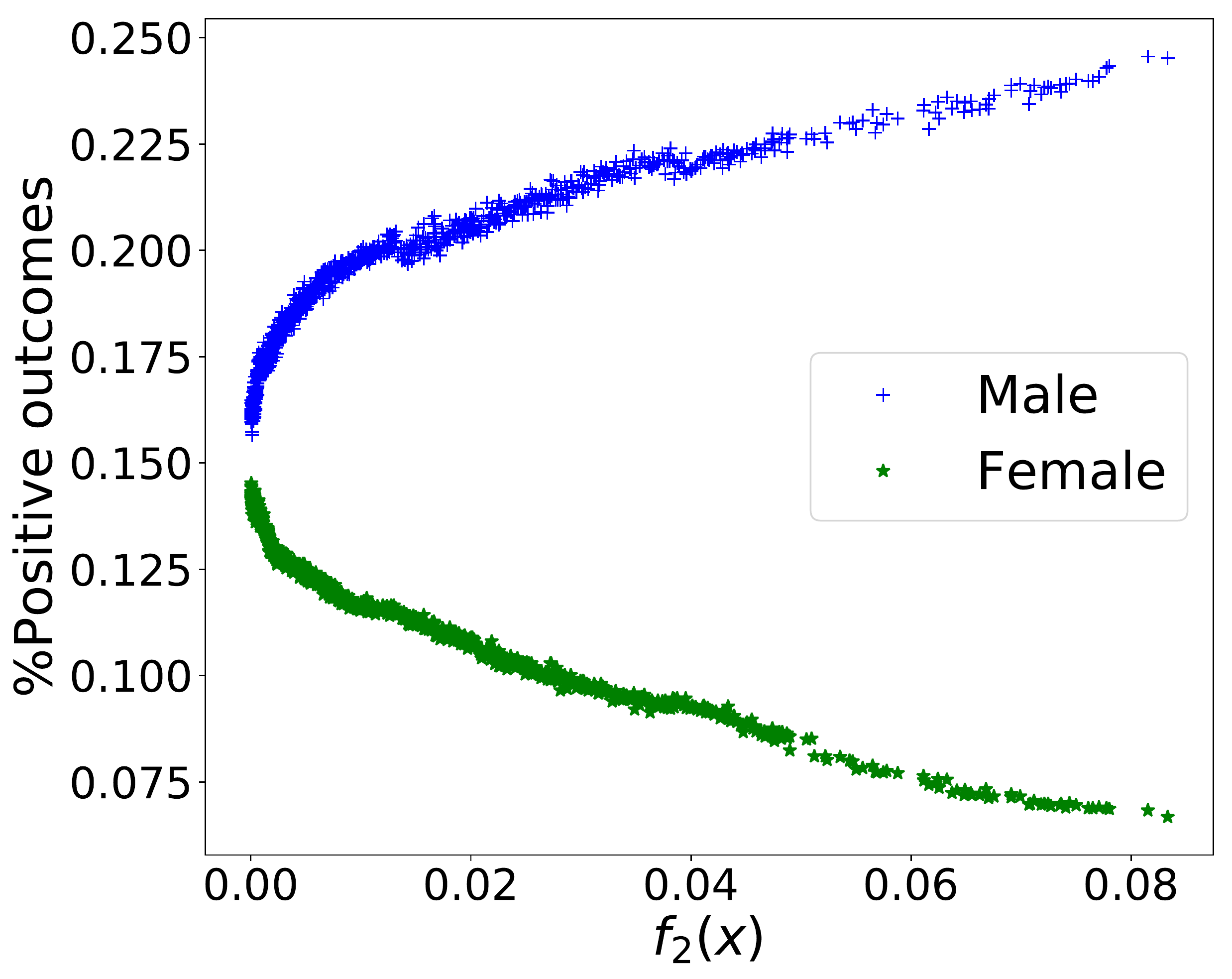}}
  \subfloat[$f_2^{\DI}(x)$ vs CV score.]{\includegraphics[width = 0.23\textwidth]{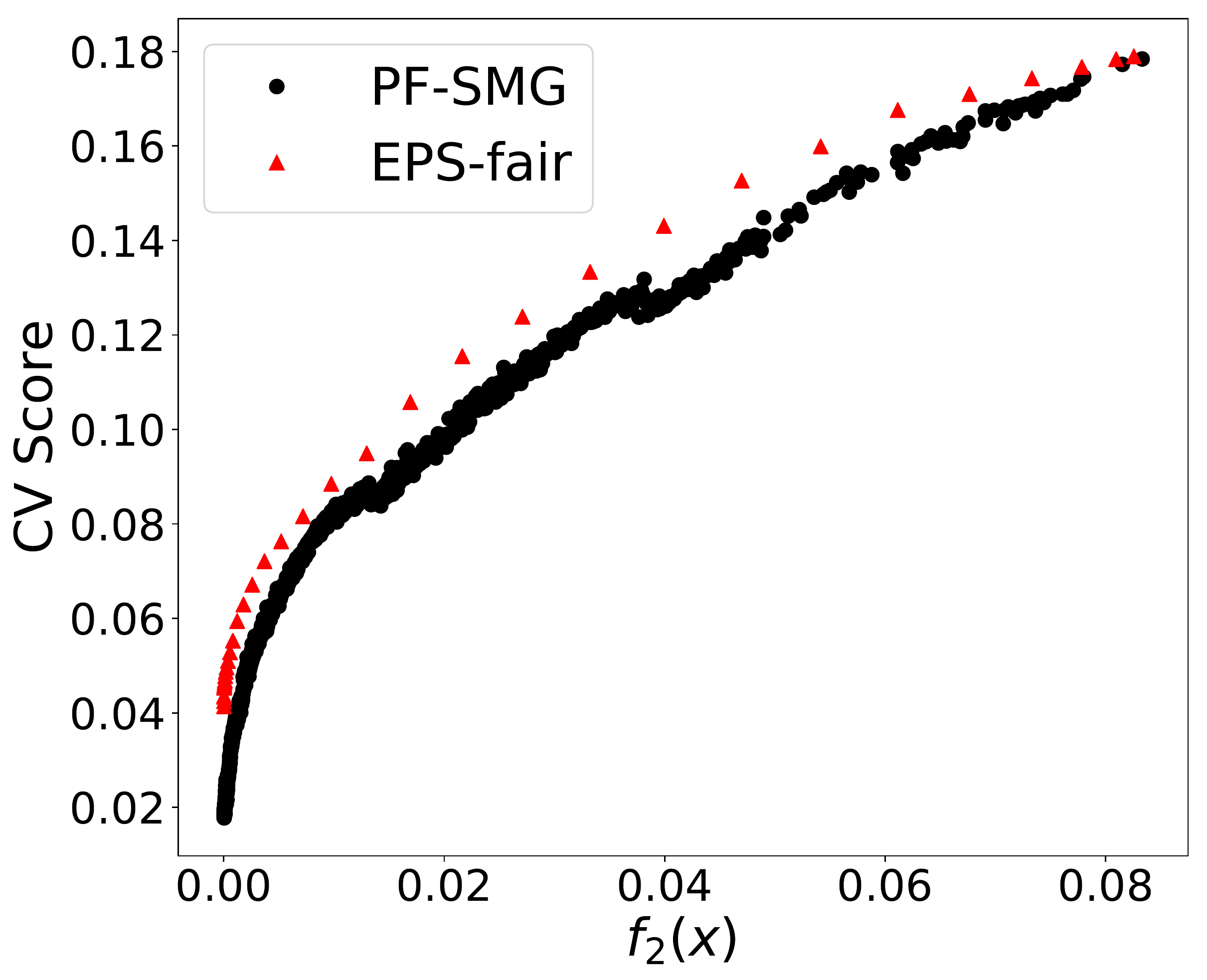}}
  \subfloat[Accuracy vs CV score.]{\includegraphics[width = 0.23\textwidth]{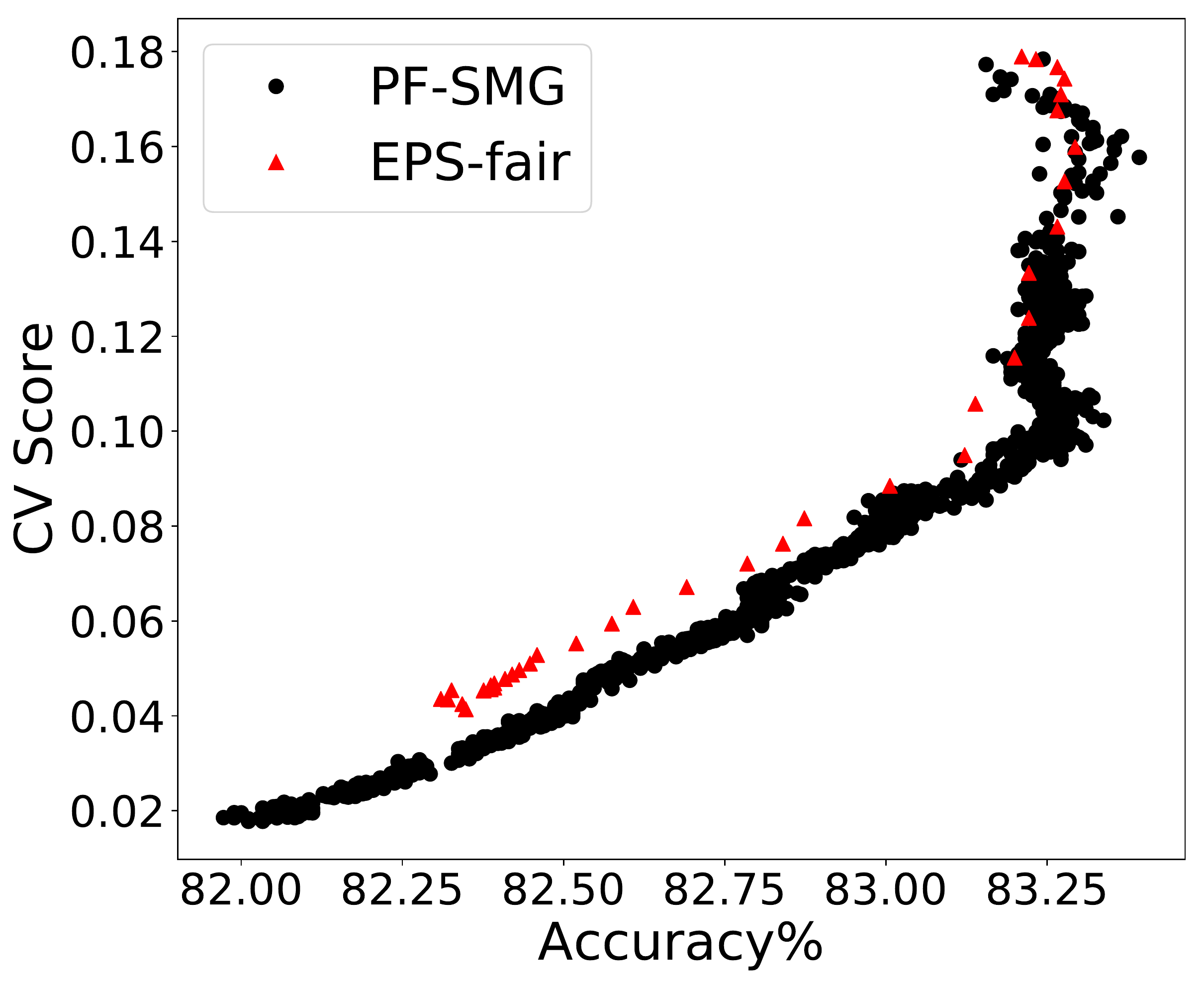}}
  \caption{Trade-off results for Adult Income dataset w.r.t. gender. Parameters used in PF-SMG: $p_1=2, p_2 = 3, \alpha_0 = 2.1$ and then multiplied by $1/3$ every $500$ iterates of SMG, $b_{1, k} = 80\times 1.018^k$, and $b_{2, k} = 50\times 1.018^k$.\label{res:Adult_gender}}
  \vskip-.2cm
\end{figure*}

{\bf Performance in terms of trade-offs.}
Considering gender as the sensitive attribute, the obtained Pareto front by PF-SMG is plotted in Figure~\ref{res:Adult_gender}~(a), reconfirming the conflicting nature of the two objectives. Given a nondominated solution $x=(c,b)$ from Figure~\ref{res:Adult_gender}~(a), the probability of getting positive prediction for each sensitive group is approximated by the percentage of positive outcomes for the data samples, i.e.,
$\mathbb{P}\{\hat{Y}(Z; x) = 1 | A = a\} \simeq \frac{N(\hat{Y} = 1, A = a)}{N(A = a)}$,
where $N(\hat{Y} = 1, A = a)$ denotes the number of instances predicted as positive in group $a$ and $N(A = a)$ is the number of instances in group $a$. For conciseness, we will only compute the proportion of positive outcomes for analysis. 

From Figure~\ref{res:Adult_gender}~(a), predictors of high accuracy are those of high disparate impact.
It is then observed from Figure~\ref{res:Adult_gender}~(b) that as $f_2^\DI$ increases, the proportion of high income adults in females decreases, which means that the predictors of high accuracy are actually unfair for females. 
Figure~\ref{res:Adult_gender}~(c) we can conclude that the value of $f_2^\DI$ has positive correlation with CV score for this dataset. Figure~\ref{res:Adult_gender}~(d) implies that almost zero disparate impact can be achieved by reducing~1.5\% of accuracy (the range of the x-axis is nearly~1.5\%).
To eliminate the impact of the fact that female is a minority in the dataset, we ran the algorithms for several sets of training samples with $50\%$ females and $50\%$ males. It turns out that the conflict is not alleviated at all. For comparison, the trade-off for the same training-testing split obtained by EPS-fair is also shown in Figures~\ref{res:Adult_gender}~(a),~(c), and (d). More comparisons using different training-testing splits are given in Appendix~\ref{more_res:DI_binary}. It is clear that PF-SMG is more robust in capturing more complete and well-spread Pareto fronts.

\begin{figure*}[ht]
  \centering
  \vskip 0cm
  \subfloat[Pareto front.]{\includegraphics[width = 0.23\textwidth]{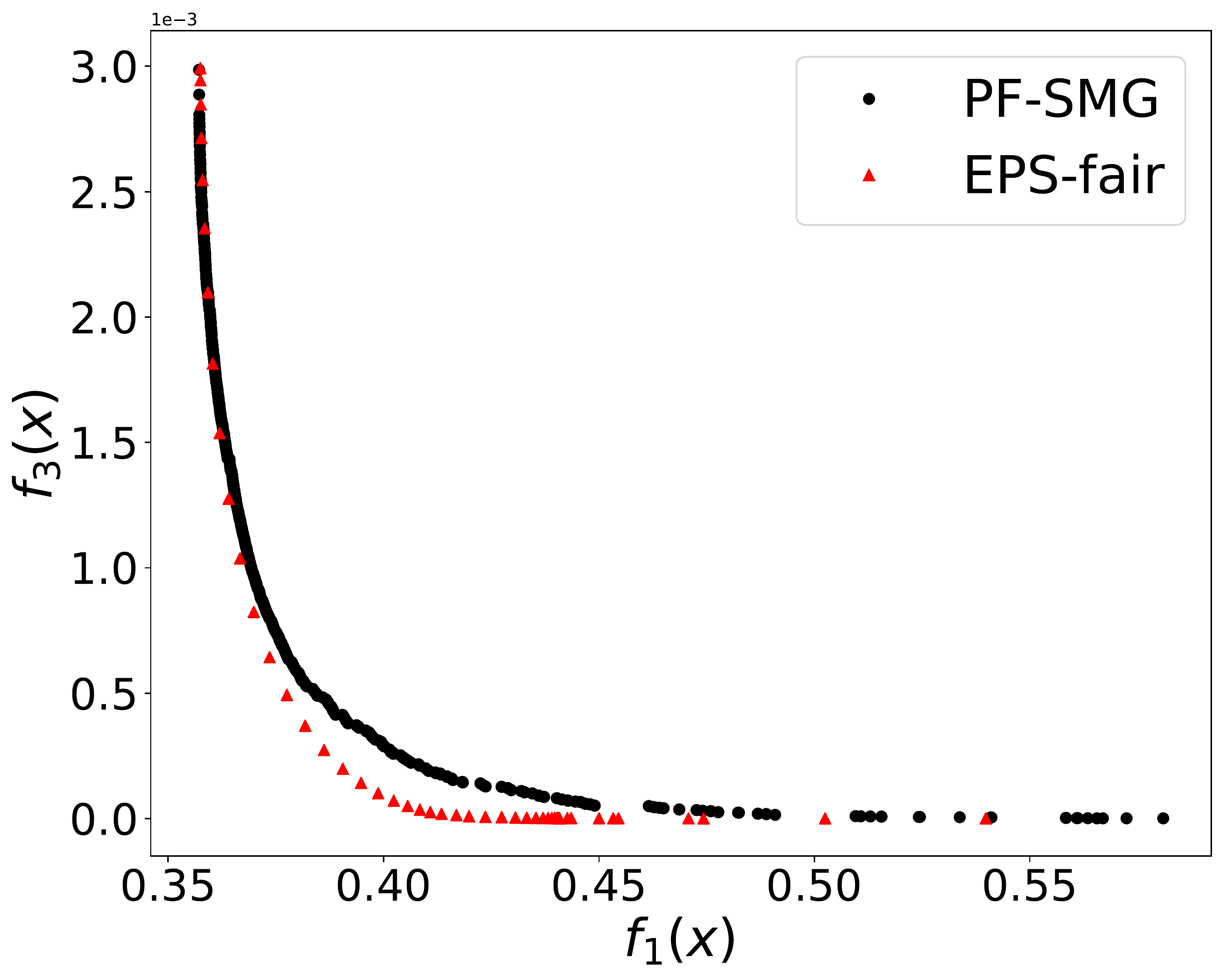}}
  \subfloat[$f_3^{\DI}(x)$ vs \%pos. outcomes.]{\includegraphics[width = 0.23\textwidth]{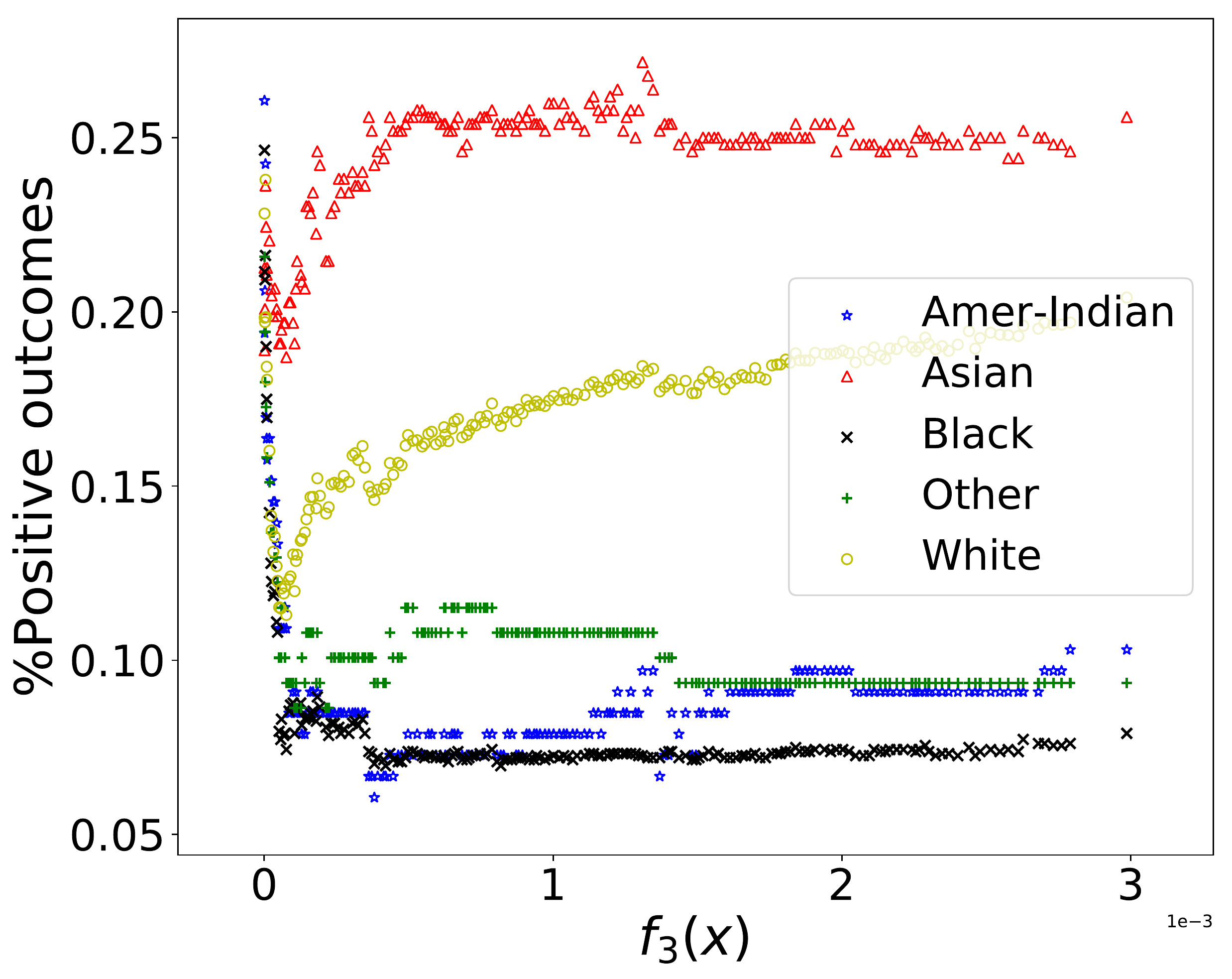}}
  \subfloat[$f_3^{\DI}(x)$ vs CV score.]{\includegraphics[width = 0.23\textwidth]{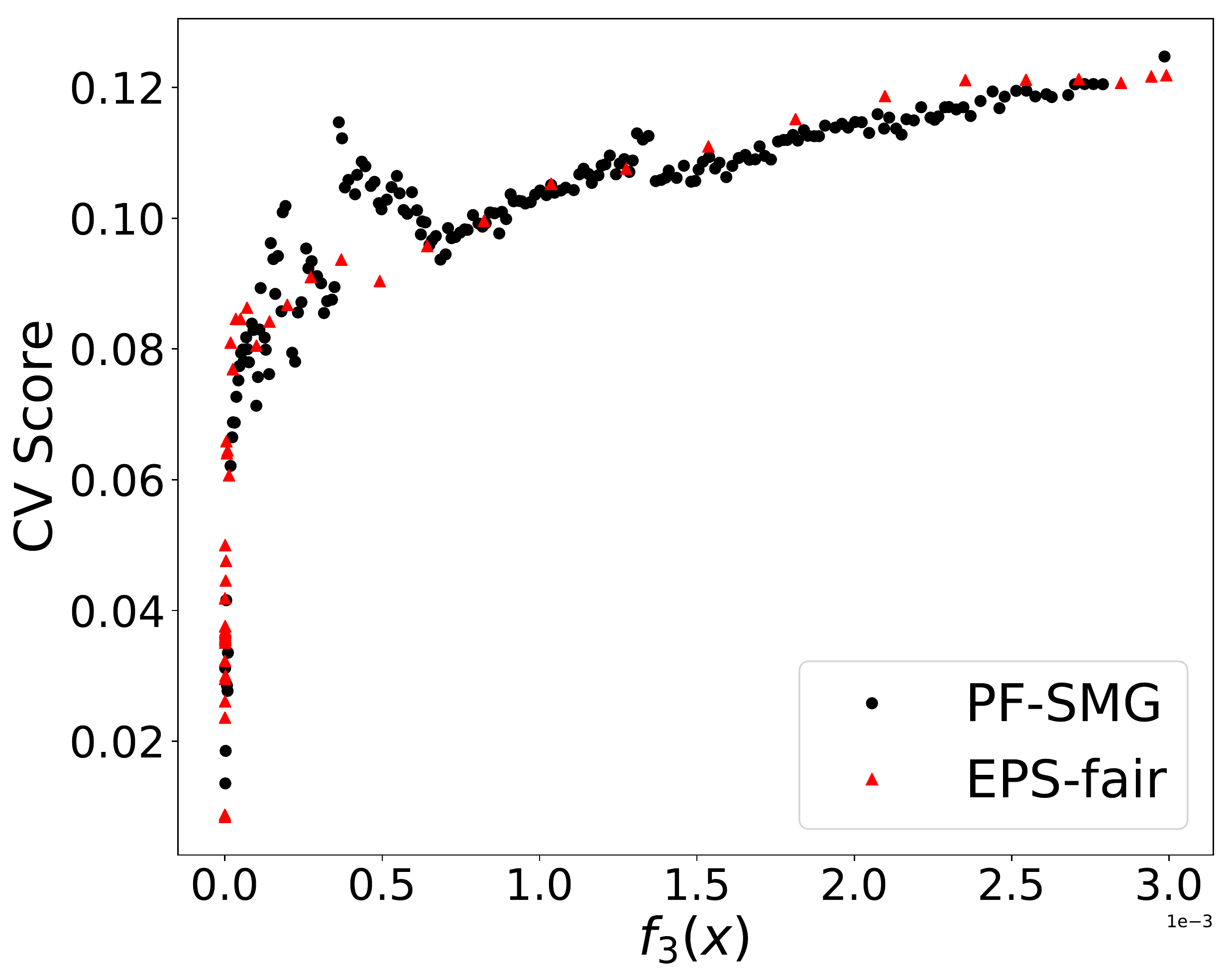}}
  \subfloat[Accuracy vs CV score.]{\includegraphics[width = 0.23\textwidth]{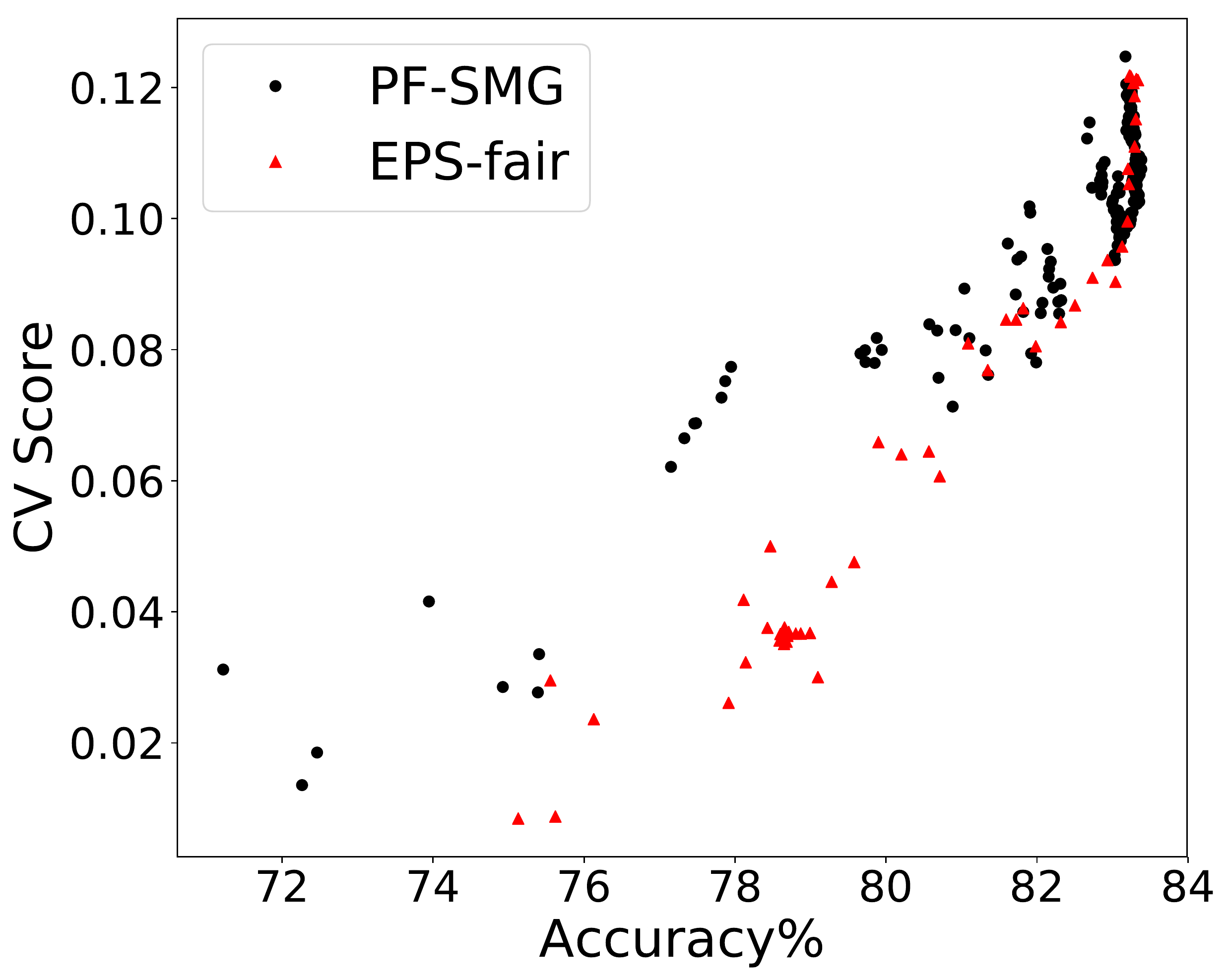}}
  \caption{Trade-off results for Adult dataset w.r.t. race. Parameters used in PF-SMG: $p_1 = 3, p_2 = 2, \alpha_0 = 2.6$ and multiplied by $1/3$ every $100$ iterates of SMG, $b_{1, k} = 50\times 1.005^k$, and $b_{2, k} = 80\times 1.005^k$. \label{res:Adult_race}}
  \vskip-.4cm
\end{figure*}

Dealing with multi-valued sensitive attribute race is more complicated. In general, if a multi-valued sensitive attribute has $K$ categorical values, we convert it to $K$ binary attributes denoted by $A^1, \ldots, A^K \in \{0, 1\}$.  Note that the binary attribute $A^i$ indicates whether the original sensitive attribute has $i$-th categorical value or not. The second objective is then modified as follows
\begin{equation}
\small
f_3^\DI(c, b) = \max_{i = 1, \ldots, K} \left(\frac{1}{N}\sum_{j=1}^N(a^i_j - \bar{a}^i)(c^\top z_j + b)\right)^2,
\label{obj2_Approx_fairness_race}
\end{equation}
which is still a convex function. We have observed that the non-smoothness introduced by the max operator in~(\ref{obj2_Approx_fairness_race}) led to more discontinuity in the true trade-off curves, and besides stochastic gradient type methods are designed for smooth objective functions. We have thus approximated the max operator in~(\ref{obj2_Approx_fairness_race}) using $S_\beta(\max(x^1, \ldots, x^\ell)) = \sum_{i = 1}^{\ell} x^i e^{\beta x^i}/\sum_{i=1}^{\ell} e^{\beta x^i}$. In our experiments, we set $\beta = 8$. Figure~\ref{res:Adult_race}~(a) plots the obtained Pareto front of the bi-objective problem of~$\min (f_1(c,b), f_3^\DI(c, b))$.
Figure~\ref{res:Adult_race}~(b) implies that solely optimizing over prediction accuracy might result in unfair predictors for American-Indian, Black, and Other. Even though the individual percentage is not monotonically decreasing, the maximum gap of percentage is increasing while~$f_3^{\DI}$ increases (Asian group excluded). Regardless of the noise, it is observed that the value of $f_3^\DI$ is increasing with CV score (Figure~\ref{res:Adult_race}~(c)) and that the prediction accuracy and CV score have positive correlation~(Figure~\ref{res:Adult_race}~(d)). Note that CV score in this case was computed as the absolute difference between maximum and minimum proportions of positive outcomes among $K$ groups. 
The dominance effect of the Pareto fronts produced by PF-SMG over the ones produced by EPS-fair, seen in Figures~\ref{res:Adult_gender}~(a),~(c), and (d), is now mixed in the ones of Figures~\ref{res:Adult_race}~(a),~(c), and~(d). 
However, the spread and overall coverage of the fronts is better for PF-SMG in all the three plots.
More results can be found in Appendix~\ref{appendix_DI_multi}.

{\bf Performance in terms of efficiency and robustness.}
We systematically evaluated the performance of the two approaches by comparing CPU time per nondominated solution, number of gradient evaluations per nondominated solution, and the quality of Pareto fronts. Such a quality is measured by a formula called \textit{purity} (which tries to evaluate how the fronts under analysis dominate each other) and two formulas for the spread of the fronts ($\Gamma$ and $\Delta$, measuring how well the nondominated points on a Pareto front are distributed). Higher purity corresponds to higher accuracy, while smaller $\Gamma$ and $\Delta$ indicate better spread. In addition, we also used hypervolume~\cite{EZitzler_LThiele_1999}, a performance measure combining purity and spread. The detailed formulas of the four measures are given in Appendix~\ref{appendix:metrics_comp}.

\begin{figure*}[ht]
  \centering
  \subfloat[][Purity.]{\includegraphics[width=0.25\textwidth]{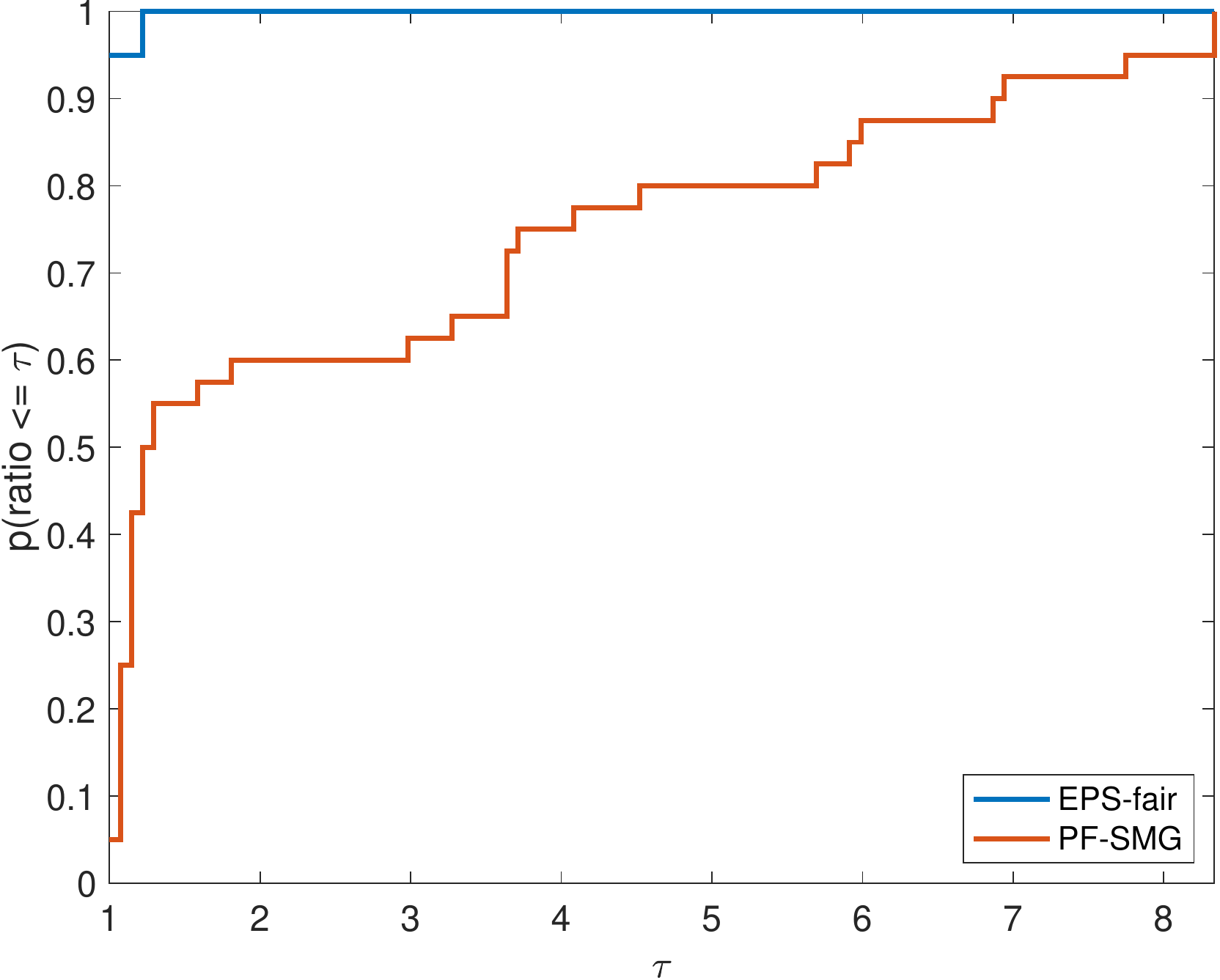}} 
  \subfloat[][Spread: $\Gamma$.]{\includegraphics[width=0.25\textwidth]{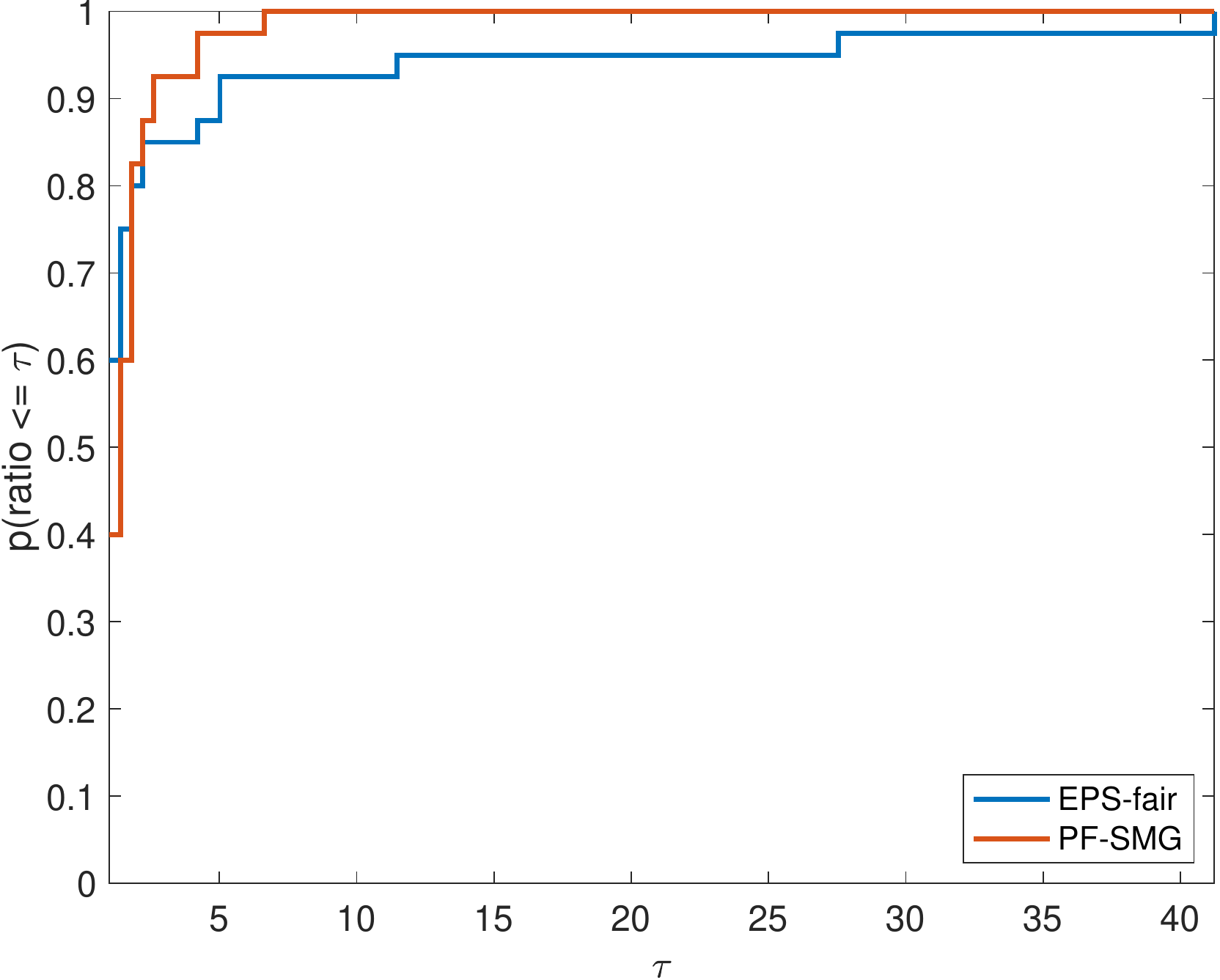}}
  \subfloat[][Spread: $\Delta$.]{\includegraphics[width= 0.25\textwidth]{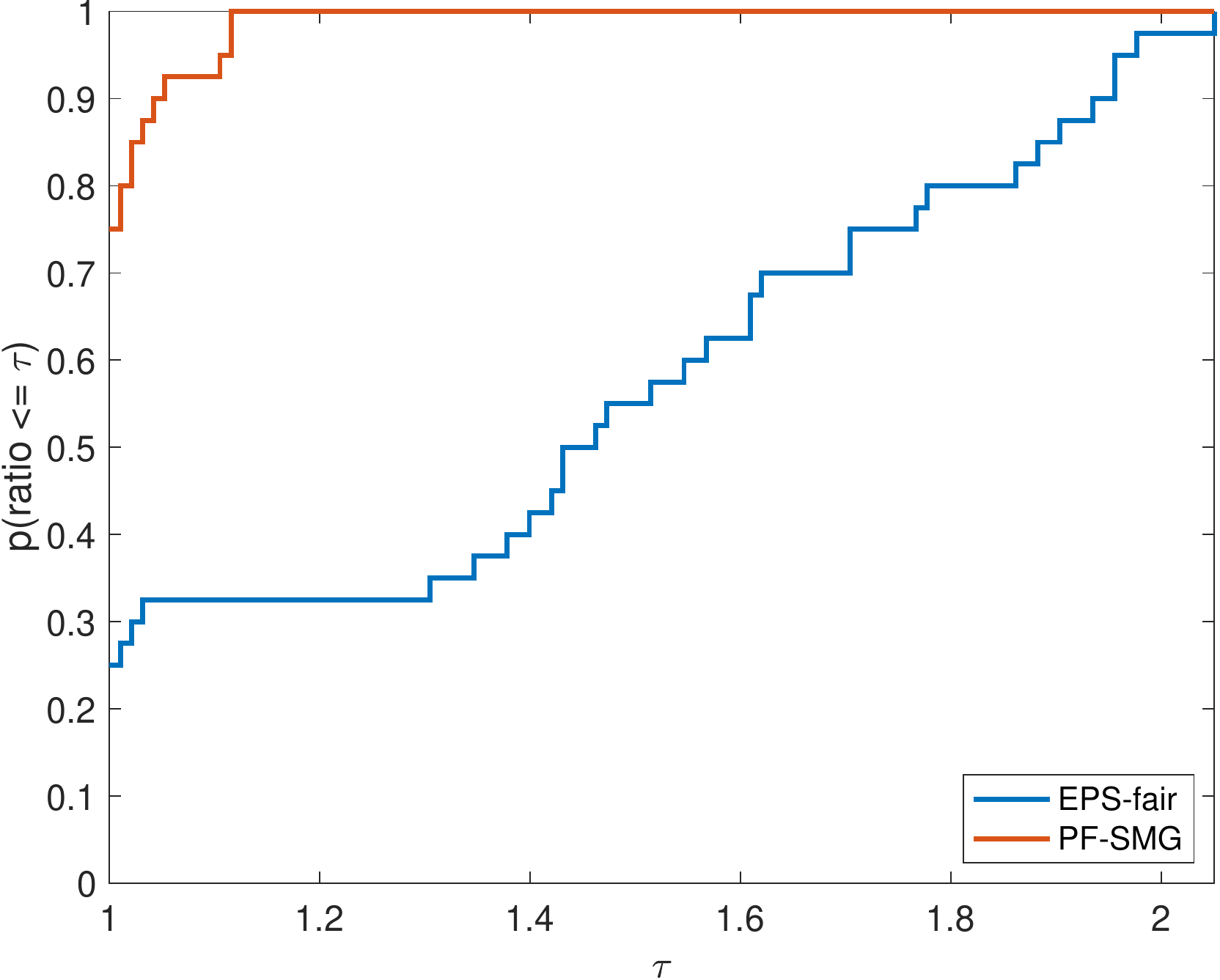}} \\
   \subfloat[][Hypervolume.]{\includegraphics[width= 0.25\textwidth]{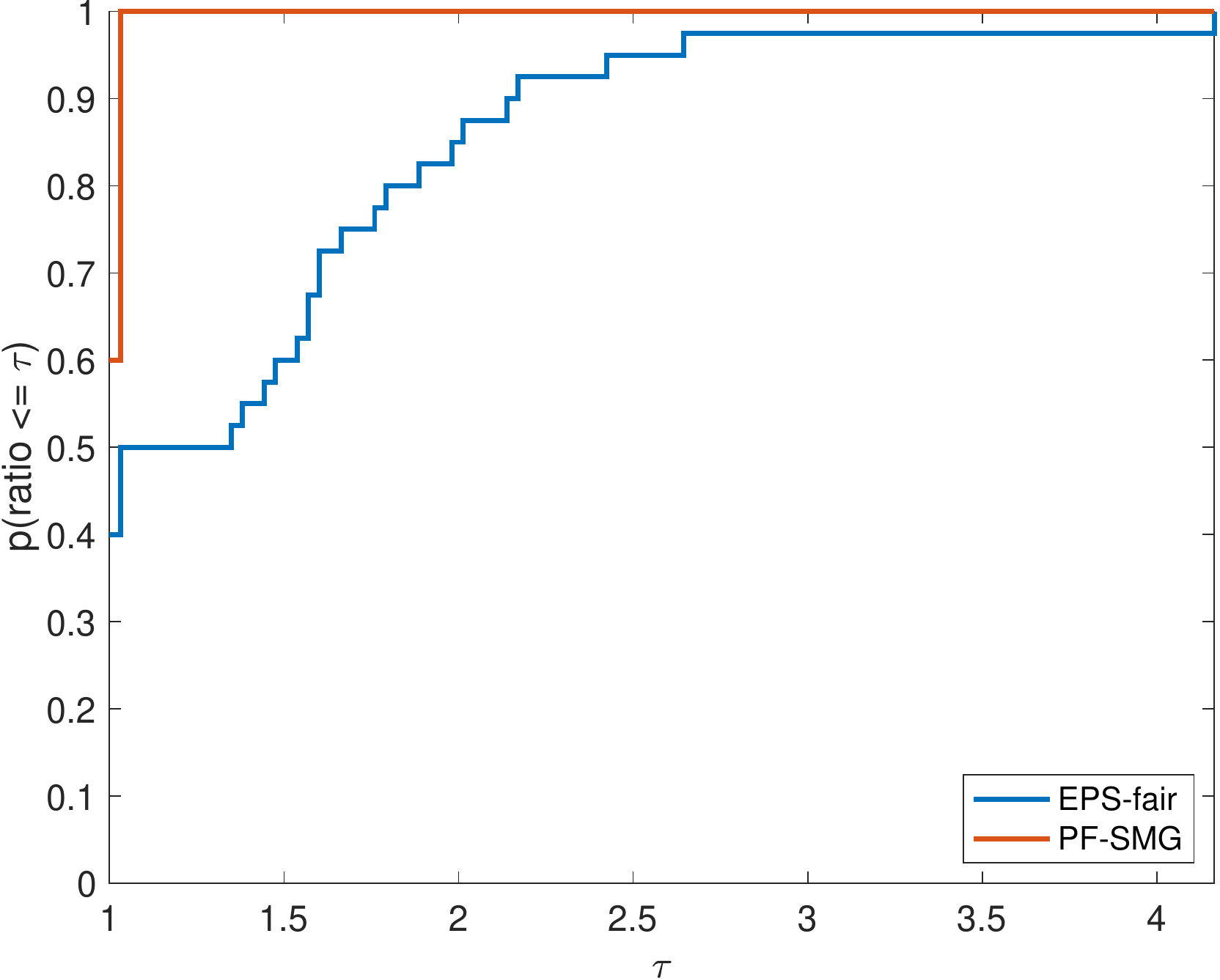}}
  \subfloat[][CPU time.]{\includegraphics[width=0.25\textwidth]{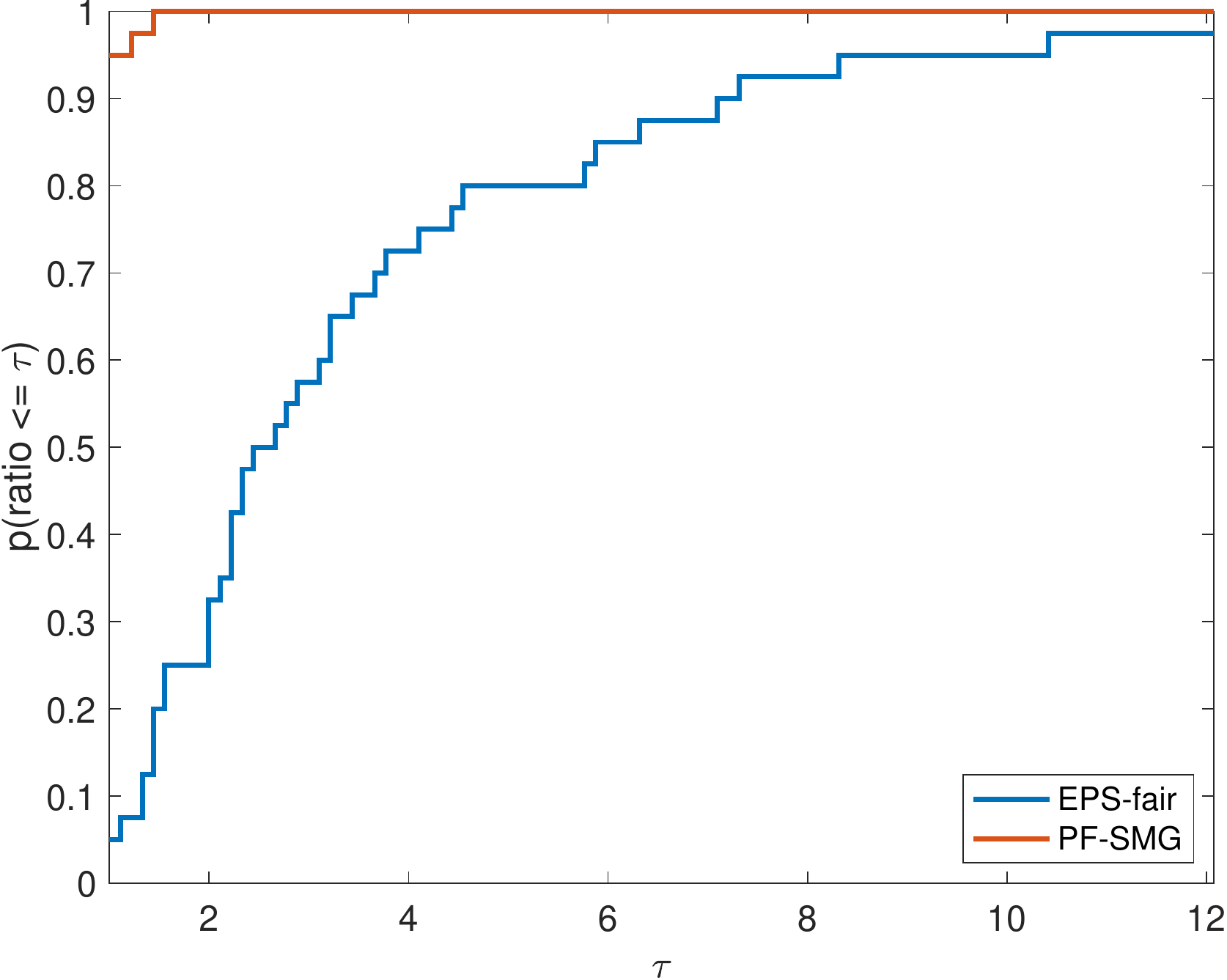}}
   \subfloat[][\#Gradient evaluations.]{\includegraphics[width=0.25\textwidth]{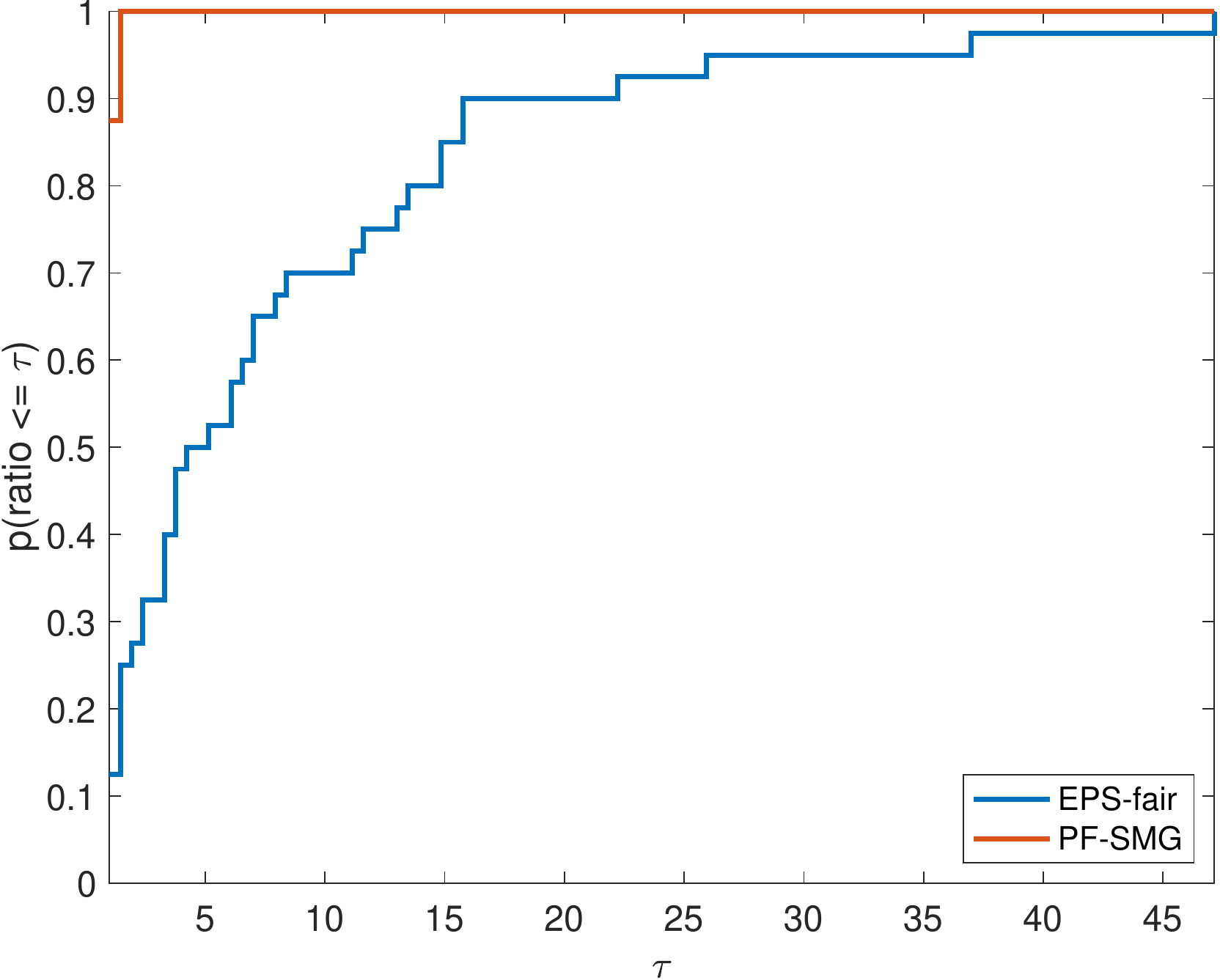}}
  \caption{Performance profiles for $40$ training datasets: PF-SMG versus EPS-fair.\label{fig:syn_profile}}
  \vskip-.4cm
\end{figure*}

We formed a set of 40 datasets with different random seeds:  $20$ from synthetic data and $20$ from \textit{Adult income} data ($10$ binary and $10$ multi-valued problems). For EPS-fair, 1,500 and 100 different values of thresholds $\epsilon$ are evenly selected to form Pareto fronts for synthetic and real datasets respectively. To guarantee a fair comparison in terms of purity, spread, and hypervolume, the Pareto front generated by PF-SMG is down-sampled to matches the size of the corresponding Pareto front obtained by EPS-fair.
The five performance profiles (see~\cite{EDDolan_JJMore_2002}) are shown in Figure~\ref{fig:syn_profile}. The purity (see Figure~\ref{fig:syn_profile}~(a)) of the Pareto fronts produced by the EPS-fair method is slightly better than the one of those determined by PF-SMG. 
However, notice that PF-SMG produced better spread fronts than EPS-fair without compromising accuracy too much (see Figures~\ref{fig:syn_profile}~(b)--(c)). In fact, one can observe from Appendix~\ref{appendix_more_res} that the resulting Pareto fronts from PF-SMG are quite close to that from EPS-fair in the cases where the latter ones dominate the former ones. Further, Figure~\ref{fig:syn_profile}~(d) shows that PF-SMG wins against EPS-fair in terms of the combination of individual solution quality and objective space coverage. In addition, PF-SMG outperforms EPS-fair in terms of computational cost quantified by CPU time and gradient evaluations (see Figures~\ref{fig:syn_profile}~(e)--(f)). 

\section{Equal Opportunity}
\label{equal_opp}
Recall that equal opportunity focuses on positive outcomes $Y = + 1$ and requires the following for $\hat{y} \in \{-1, +1\}$
\begin{equation*}
\mathbb{P}\{\hat{Y} = \hat{y} | A = 0, Y = +1\} \;=\; \mathbb{P}\{\hat{Y} = \hat{y} | A = 1, Y = +1\}.
\end{equation*}
When $\hat{y} = -1$ in the above equation, this condition essentially suggests equalized \textit{false negative rate} (FNR) across different groups. Similarly, the case of $\hat{y} = +1$ corresponds to equalized \textit{true positive rate} (TPR). Given that $\mbox{FNR} + \mbox{TPR} = 1$ always holds, we will focus on the $\hat{y} = -1$ case where qualified candidates are falsely classified in a negative class by the predictor $\hat{Y}$.

For simplicity, let $\FNR_a(\hat{Y}) = \mathbb{P}\{\hat{Y} = -1 | A = a, Y = +1\}$, $a \in \{0, 1\}$. The CV score associated with equal opportunity is now defined as follows
\begin{equation}
    \CV_{\FNR}(\hat{Y}) = |\FNR_0(\hat{Y}) - \FNR_1(\hat{Y})|.
    \label{CV_FNR}
\end{equation}
Since equalized FNR indicates statistical independence between sensitive attributes and instances that have positive targets but falsely predicted as negative, $\CV_{\FNR}(\hat{Y})$ could thus be approximated~\cite{MBZafar_etal_2017} by
\[\mbox{Cov}(A, \psi(Z,Y; c, b)) \simeq\; \tfrac{1}{N} \sum_{j = 1}^N (a_j - \bar{a})\psi(z_j,y_j; c, b),\]
where $\psi(z,y; c, b) = \min\{0, \frac{(1 + y)}{2}y\phi(z; c, b)\}$. Here, $(1 + y)/2$ excludes truly negative instances $y = -1$ and $y\phi(z; c, b) < 0$ implies wrong prediction.
Similar to~\eqref{obj2_Approx_fairness}, the objective function for equalized FNR is given by
\begin{equation*}
\textstyle f_4^{\FNR}(c, b) = \left(\tfrac{1}{N} \sum_{j = 1}^N (a_j - \bar{a})\psi(z_j,y_j; c, b)\right)^2,
\end{equation*}
which is a nonconvex finite-sum function. (Note that as in~(\ref{obj2_Approx_fairness_race}) we have also smoothed here the min operator in $\psi(z,y; c, b)$.)
Now, the finite-sum bi-objective problem becomes
\begin{equation}
    \label{bi_obj_FNR_race}
    \min \; \left(f_1(c, b), f_4^{\FNR}(c, b)\right).
\end{equation}
Due to nonconvexity of the above problem, EPS-fair was not able to produce any reasonable trade-offs, and thus a comparison to our approach is not even applicable. The ProPublica COMPAS dataset~\cite{ProPublica_COMPAS} contains features that are used by COMPAS algorithms~\cite{JLarson_etal_2016} for scoring defendants together with binary labels indicating whether or not a defendant recidivated within 2 years after the screening. For analysis, we take blacks and whites from the \textit{two-years-violent} dataset~\cite{ProPublica_COMPAS} and consider features including gender, age, number of prior offenses, and charge for which the person was arrested. For consistency with the word ``opportunity'', we marked the case where a defendant is non-recidivist as the positive outcome. The demographic composition of the dataset is given in Table~\ref{COMPAS_dataset_race} in Appendix~\ref{data_tables}. Due to shortage of data, we use the whole dataset for training and testing. 

\begin{figure*}[ht]
  \centering
  \vskip 0cm
  \subfloat[Pareto front.]{\includegraphics[width = 0.25\textwidth]{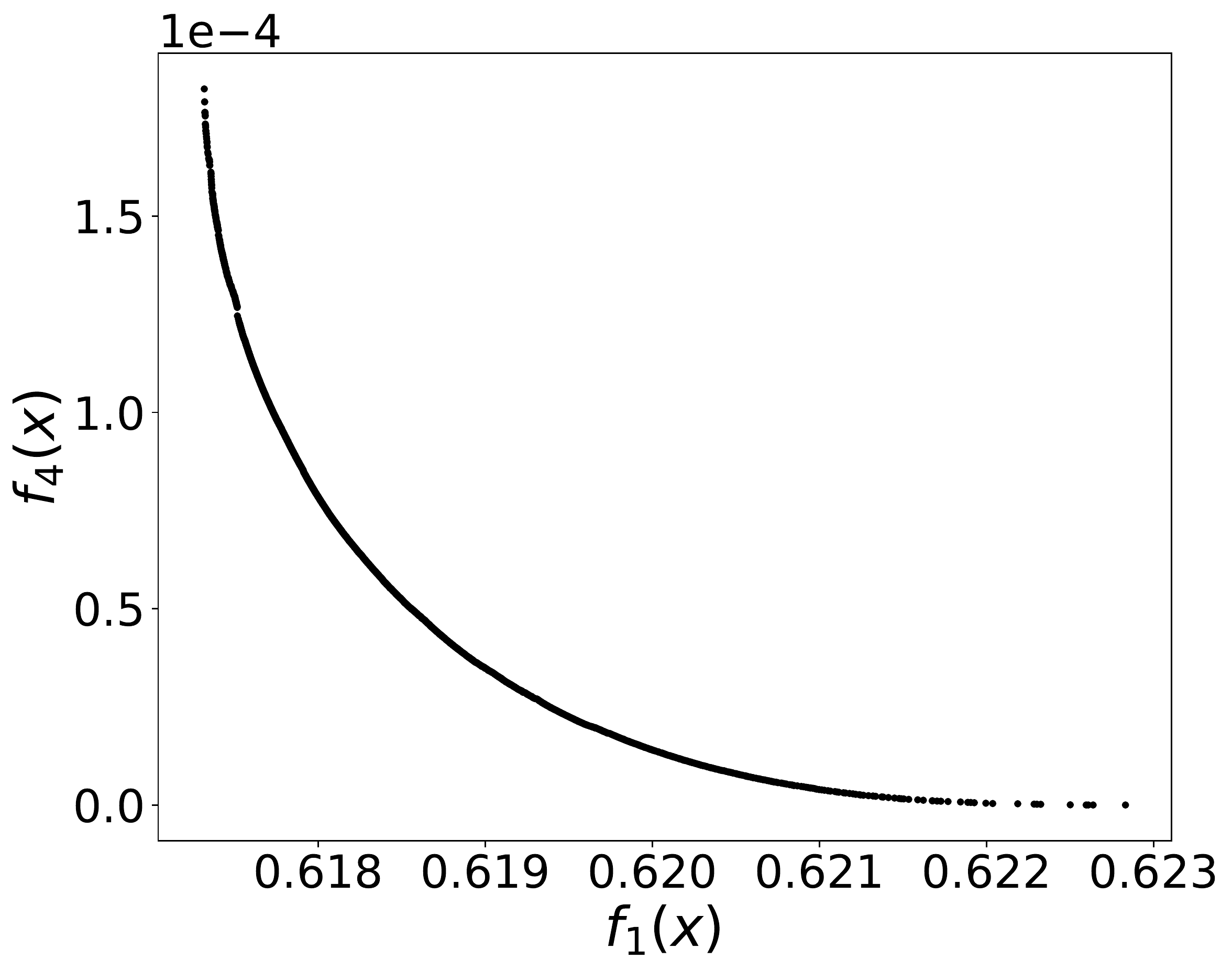}}
  \subfloat[$f^{\FNR}_4(x)$ vs FNR.]{\includegraphics[width = 0.25\textwidth]{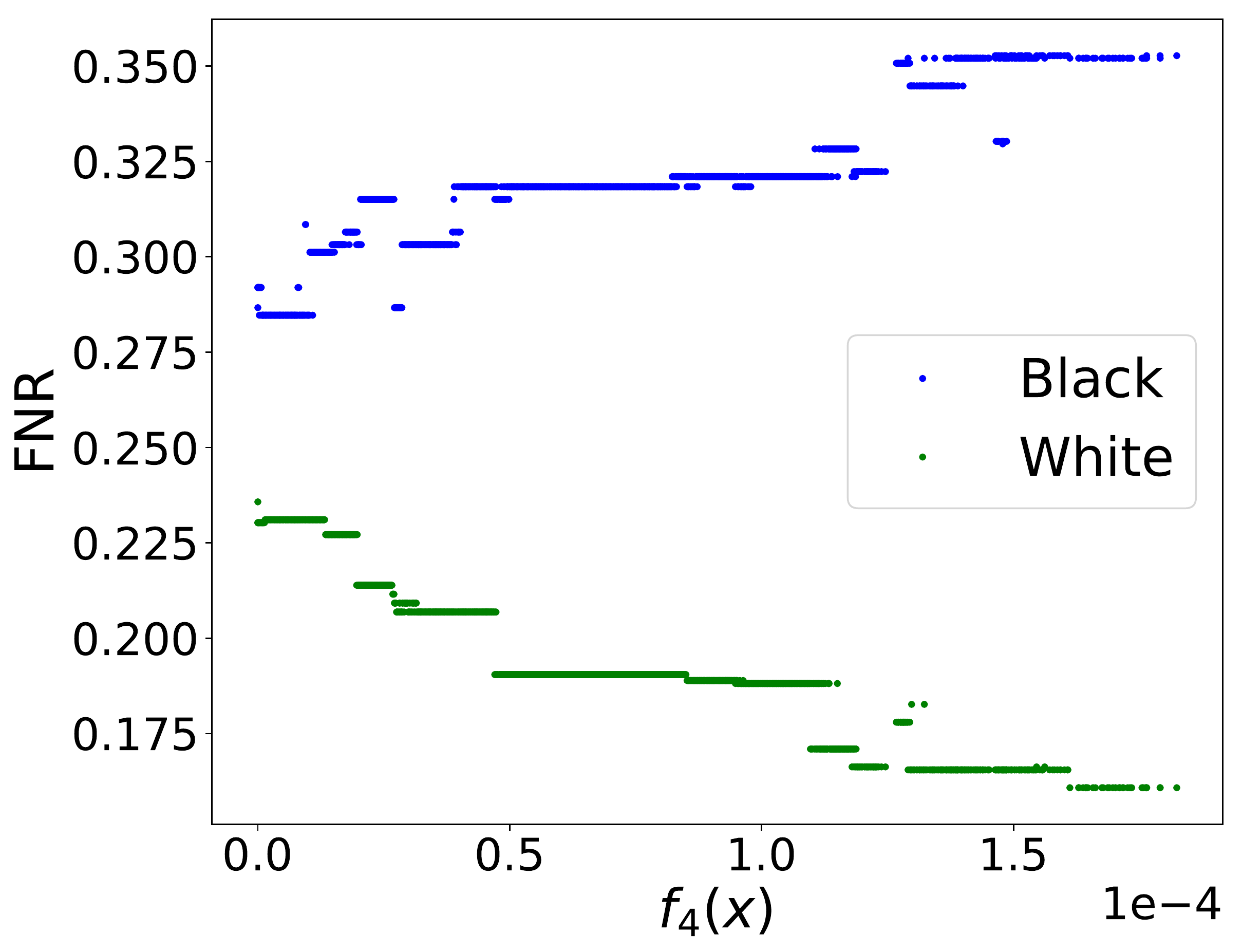}}
  \subfloat[$f^{\FNR}_4(x)$ vs CV score.]{\includegraphics[width = 0.25\textwidth]{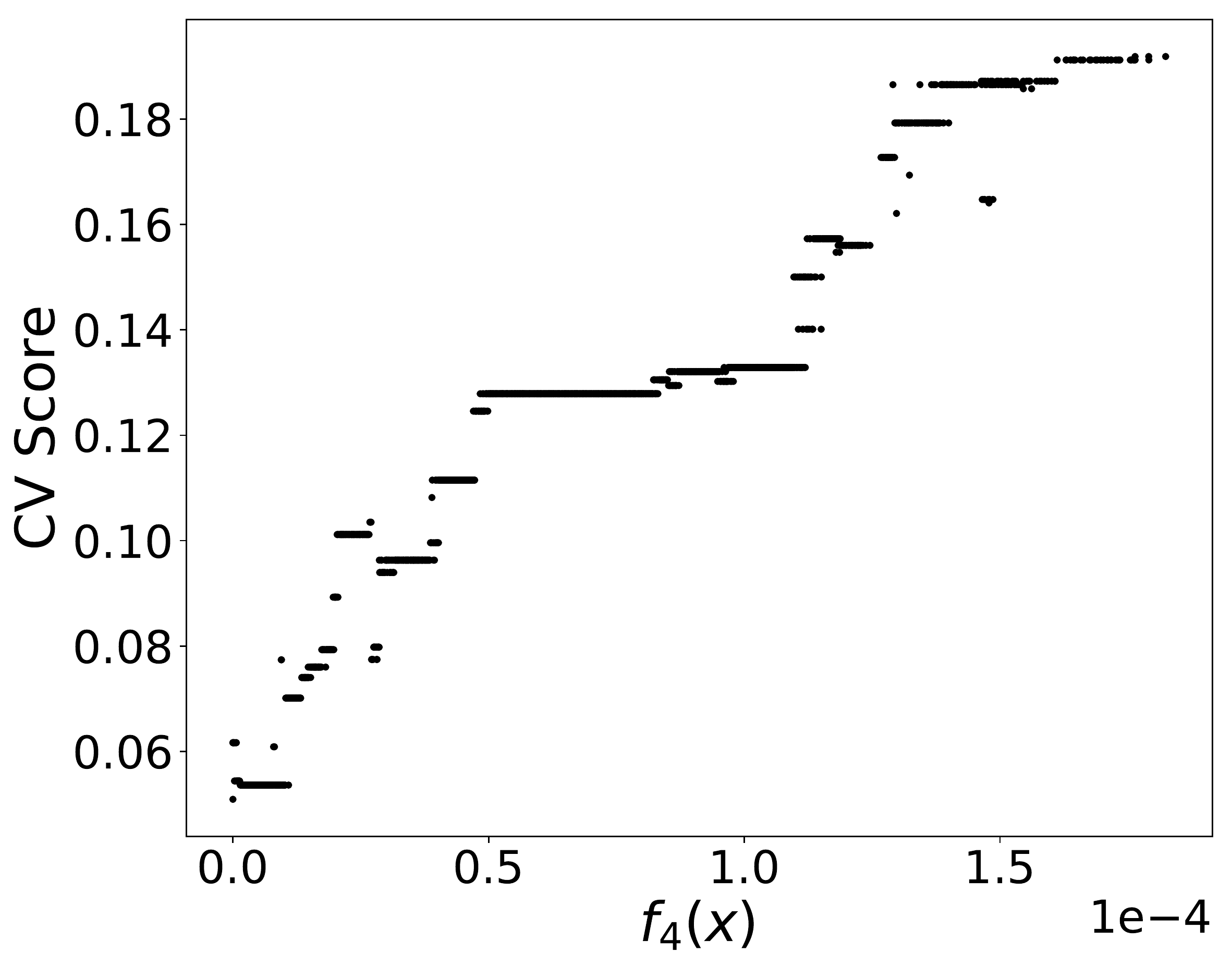}}
  \subfloat[Accuracy vs CV score.]{\includegraphics[width = 0.25\textwidth]{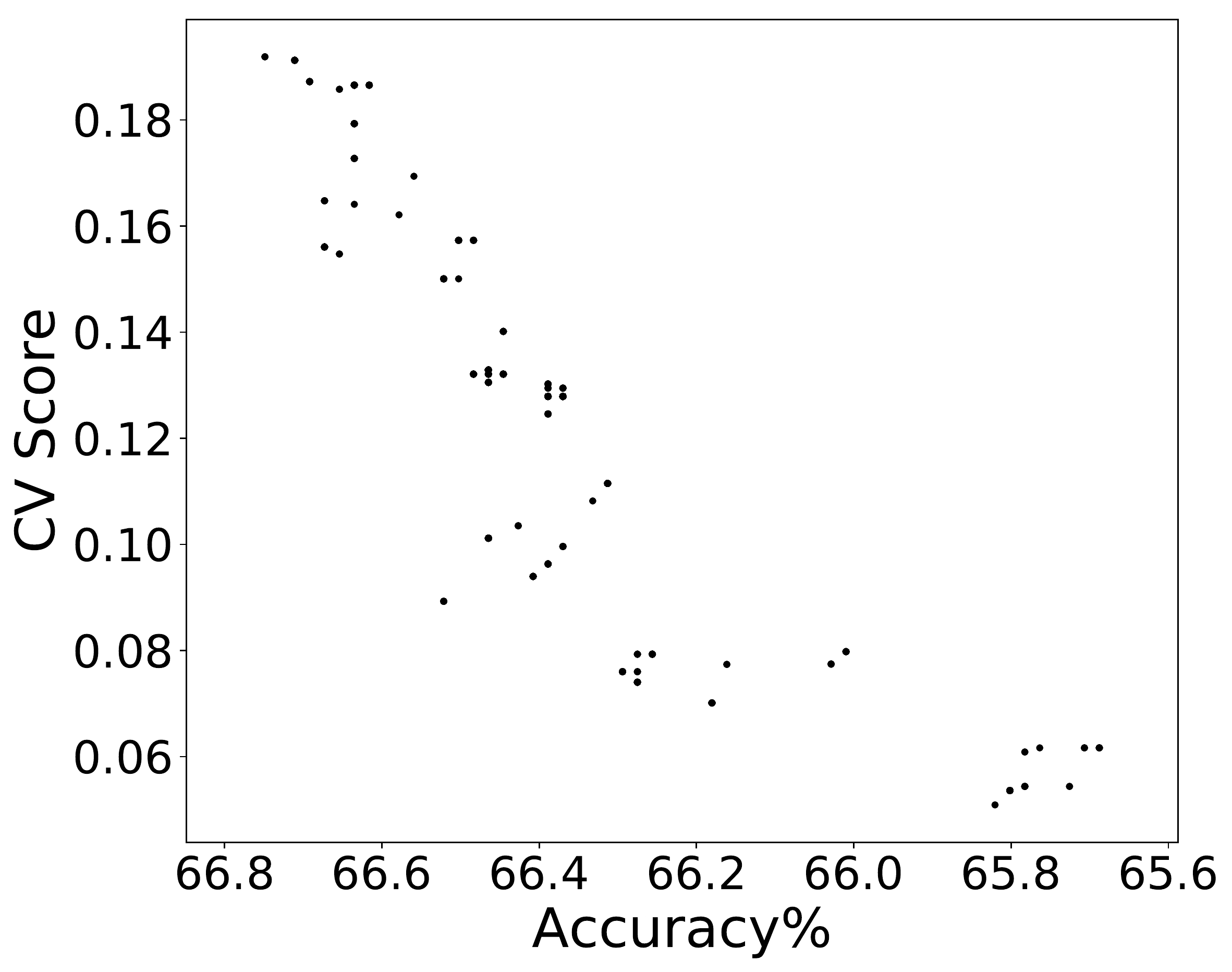}}
  \caption{Trade-off results for COMPAS dataset w.r.t. race. Parameters used in PF-SMG: $p_1 = 3, p_2 = 3, \alpha_0 = 4$ and multiplied by $1/3$ every $100$ iterates of SMG, and $b_{1, k} = b_{2, k} = 80\times 1.005^k$\label{res:COMPAS_race}.}
  \vskip-.4cm
\end{figure*}

By applying PF-SMG to the bi-objective problem~\eqref{bi_obj_FNR_race}, we obtained the trade-off results in Figure~\ref{res:COMPAS_race}. The conflicting nature of prediction loss and equalized FNR is confirmed by the Pareto front in Figure~\ref{res:COMPAS_race}~(a). For each nondominated solution $x$, we approximated FNR using samples by~$\FNR_a(\hat{Y}(Z; x)) \simeq {N(\hat{Y}(Z; x) = -1, A = a, Y = +1})/{N(A = a, Y = +1})$ where $N(\cdot)$ is the number of instances satisfying all the conditions.

From the rightmost part of Figure~\ref{res:COMPAS_race}~(b), we can draw a similar conclusion as in~\cite{JLarson_etal_2016} that black defendants (blue curve) who did not reoffend are accidentally predicted as recidivists twice as often as white defendants (green curve) when using the most accurate predictor obtained (i.e., $0.35$ versus $0.175$). However, the predictor associated with zero covariance (see the leftmost part) mitigates the situation to $0.28$ versus $0.23$, although by definition the two rates should converge to the same point. This is potentially due to the fact that the covariance is not well approximated using a limited number of samples. In fact, the leftmost part of Figure~\ref{res:COMPAS_race}~(c) shows that zero covariance does not correspond to zero $\CV_{\FNR}$.  Finally, Figure~\ref{res:COMPAS_race}~(d) provides a rough confirmation of positive correlation between CV score and prediction accuracy.

The results for equal opportunity presented in this section show the applicability of our multi-objective optimization framework when dealing with nonconvex fairness measures.

\section{Handling Multiple Sensitive Attributes and Multiple Fairness Measures}
\label{multi_attributes_measures}
A main advantage of handling fairness in machine learning through multi-objective optimization is
the possibility of considering any number of criteria. In this section, we explore two possibilities, multiple sensitive attributes and multiple fairness measures.

\subsection{Multiple Sensitive Attributes}
\begin{figure*}[ht]
  \centering
  \subfloat[Pareto front.]{\includegraphics[width = 0.26\textwidth]{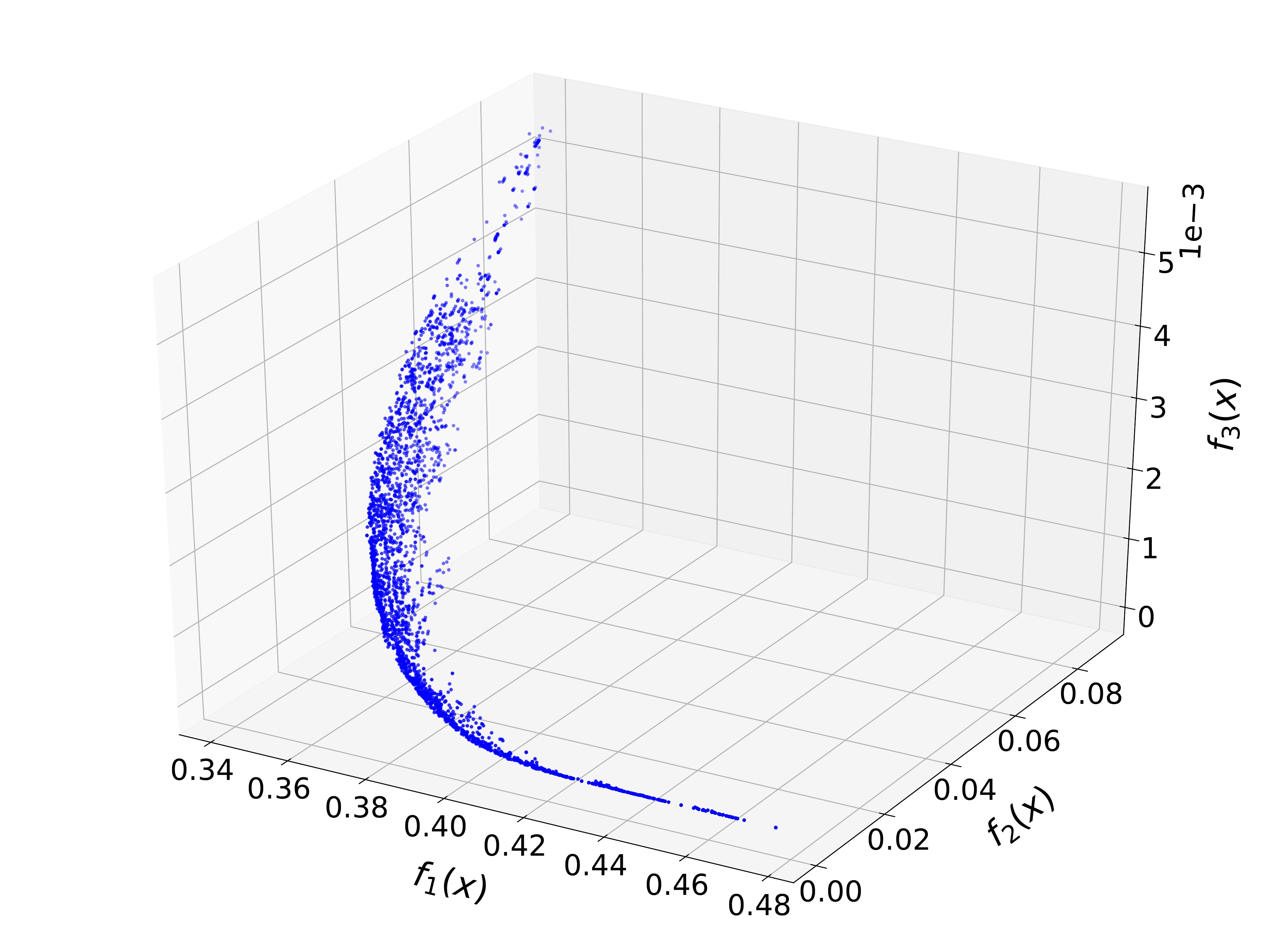}}
  \subfloat[Projection to $f_2$-$f_3$ objective space.]{\includegraphics[width = 0.23\textwidth]{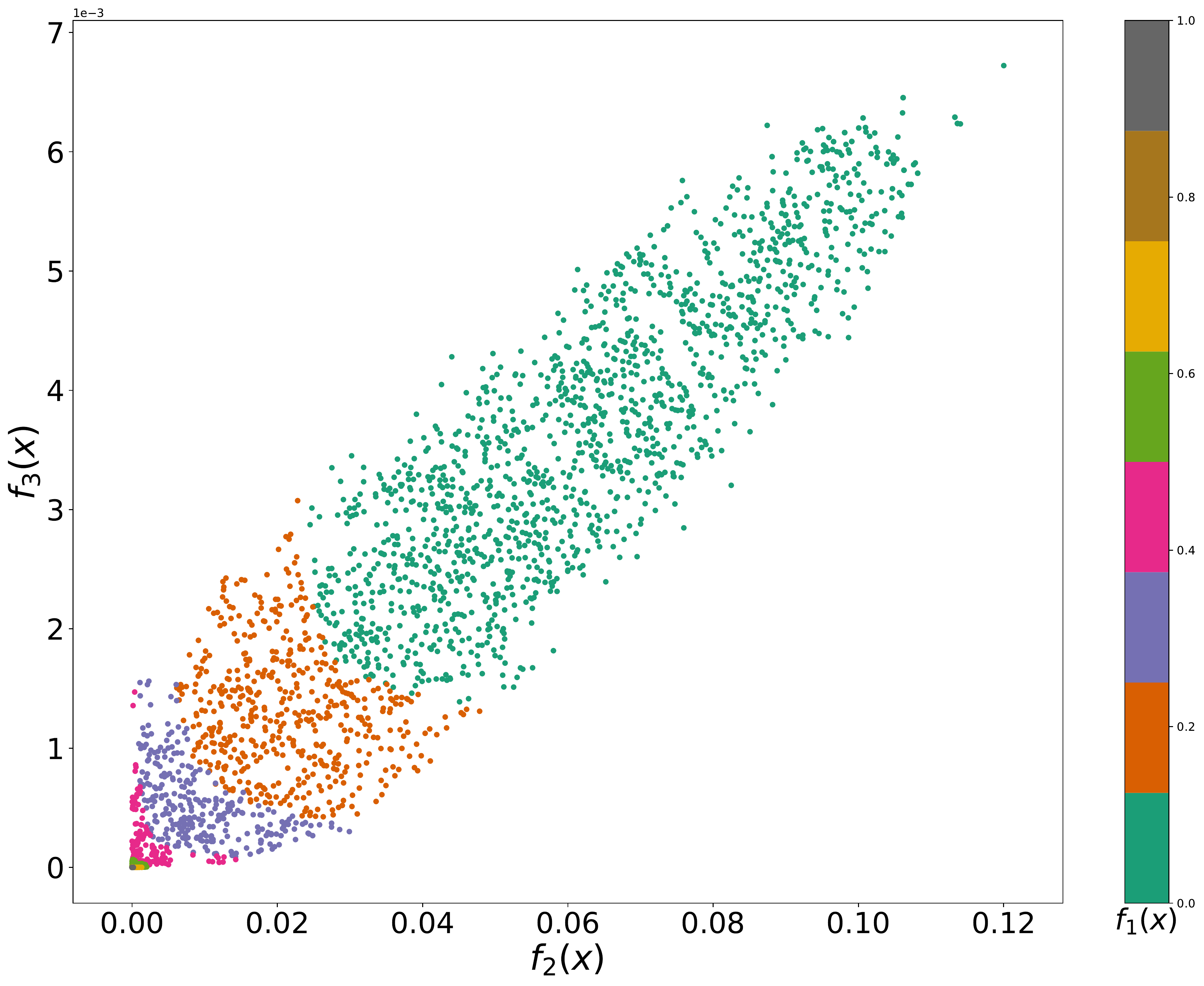}}
  \subfloat[Pareto front.]{\includegraphics[width = 0.26\textwidth]{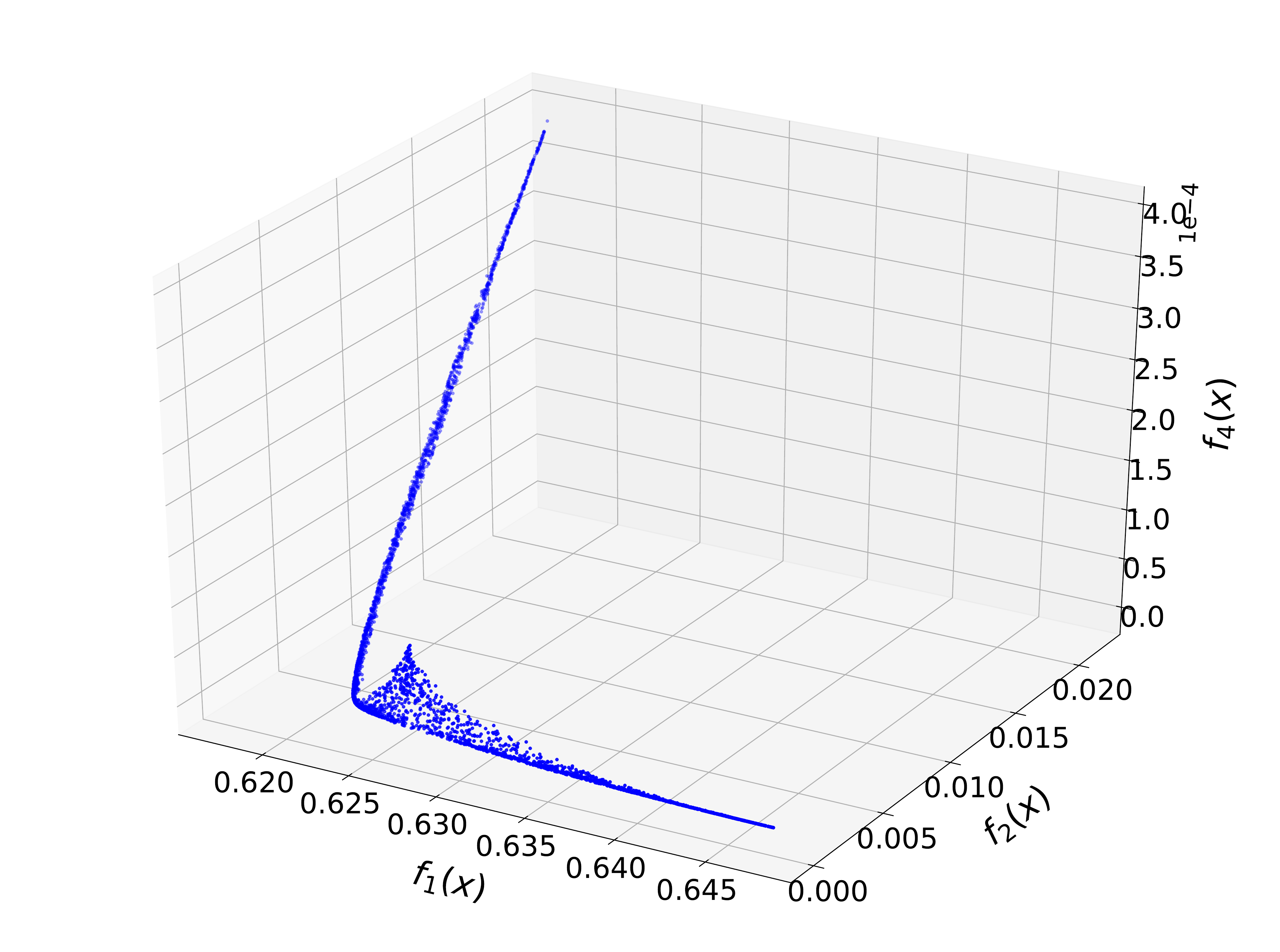}} 
  \subfloat[Projection to $f_2$-$f_4$ objective space.]{\includegraphics[width = 0.23\textwidth]{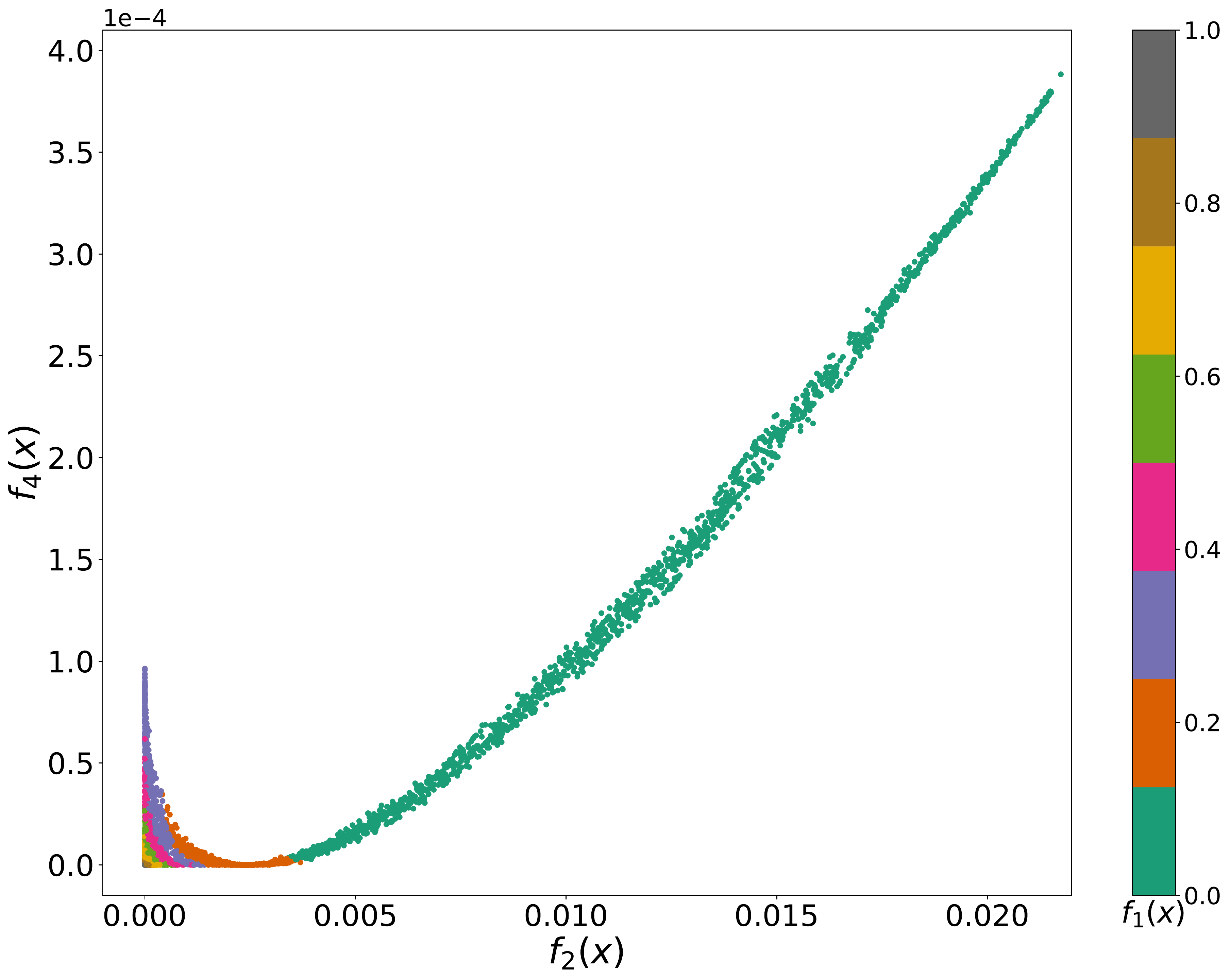}}
  \caption{(a)-(b):trade-off results for problem~\eqref{3_obj_multi_attributes} using Adult Income dataset. Parameters used in PF-SMG: same as in Fig. \ref{res:Adult_race} except for $b_{1, k} = b_{2, k} = b_{3, k} = 80\times 1.005^k$; (c)-(d): trade-off results for problem~\eqref{3_obj_multi_fairness} using COMPAS dataset. Parameters used in PF-SMG: same as in Fig.~\ref{res:COMPAS_race} except for $b_{1, k} = b_{2, k} = b_{3, k} = 80\times 1.005^k$. \label{res:threeObjs}}
  \vskip-.4cm
\end{figure*}

Let us see first how we can handle more than one sensitive attribute.
One can consider a binary sensitive attribute (e.g. gender)
and a multi-valued sensitive attribute (e.g. race),
and formulate the following tri-objective problem
\begin{equation}
\textstyle
    \min~(f_1(c, b), f_2^{\DI}(c, b), f_3^{\DI}(c, b)). \label{3_obj_multi_attributes}
\end{equation}
In our experiments, we use the \textit{Adult Income} dataset and the splitting of training and testing samples of Subsection~\ref{subsec:real_data}.
A 3D Pareto front is plotted in Figure~\ref{res:threeObjs}~(a) resulting from the application of PF-SMG to~(\ref{3_obj_multi_attributes}), with gender ($f_2^{\DI}$) and race ($f_3^{\DI}$) as the two sensitive attributes.

Figure~\ref{res:threeObjs} (b) depicts all the nondominated points projected onto the $f_2$--$f_3$ objective space, where the points are colored according to the range of prediction loss. It is observed that there is no conflict between $f_2^{\DI}$ and  $f_3^{\DI}$. Although it could happen for other datasets, 
eliminating disparate impact with respect to gender does not hinder that with respect to race for this dataset.
Intuitively, one could come up with a predictor where the proportions of positive predictions for female and male are equalized and the proportions of positive predictions for different races are equalized within the female and male groups separately, which would lead to zero disparate impact in terms of gender and race simultaneously.

\subsection{Multiple Fairness Measures}
Now we see how to handle more than one fairness measure.
As an example, we consider handling two fairness measures (disparate impact and equal opportunity) in the case of a binary sensitive attribute, and formulate the following tri-objective problem
\begin{equation}
\textstyle
    \min~(f_1(c, b), f_2^{\DI}(c, b), f_4^{\FNR}(c, b)).     \label{3_obj_multi_fairness}
\end{equation}
In our experiments, we use the whole ProPublica COMPAS \textit{two-years-violent} dataset (see~Section \ref{equal_opp}) for both training and testing. Figure~\ref{res:threeObjs}~(c) shows an approximated 3D Pareto front
(resulting from the application of PF-SMG to~(\ref{3_obj_multi_fairness})).
By projecting all the obtained nondominated points onto the 2D $f_2$--$f_4$ objective space, we have subplot~(d), where the different colors indicate the different levels of prediction loss.  From Figure~\ref{res:threeObjs}~(d), one can easily find that an unique minimizer (in the gray area with lower prediction accuracy) exists for both $f_2^{\DI}$ and $f_4^{\FNR}$, and thus conclude that there is indeed no conflict between disparate impact and equal opportunity. In fact, by definition, the CV score~\eqref{CV_FNR} generalized to equal opportunity is a component of the CV score~\eqref{CVscore} measuring disparate impact. Therefore, in the darkgreen area where the accuracy is high enough, the values of the two fairness measures are aligned and increasing as the prediction accuracy increases. Interestingly, we have discovered a little Pareto front between $f_2$ and $f_4$ when the accuracy is limited in a certain medium level in between.


The proposed multi-objective approach works well in handling more than one sensitive attribute or multiple fairness measures. We point out that looking at Pareto fronts for three objectives helps us identifying the existence of conflicts among any subset of two objectives (compared to looking at Pareto fronts obtained just by solving the corresponding bi-objective problems). In the above experiments, by including $f_1$, we were able to obtain additional helpful information in terms of decision-making reasoning.

\section{Streaming Data}
\label{stream_data}
As we claimed in the Abstract and Introduction, another advantage of an SA-based approach like ours is its ability to handle streaming training data. We conducted a preliminary test using the Adult Income dataset and gender as the binary sensitive attribute. To simulate the streaming scenario, the whole dataset is split into batches of 2,000. The initial Pareto front is constructed by applying PF-SMG to one batch of 2,000 samples. Each time a new batch of samples is given, the Pareto front is then updated by selecting a number of nondominated points from the current Pareto front as the starting list for PF-SMG. Figures~\ref{res:Adult_gender_stramingData} given in Appendix~\ref{appendix:streaming_data} shows how the successive Pareto fronts approach the final one computed for the whole dataset.

\section{Concluding Remarks}
\label{sec:conclusions}
We have proposed a novel stochastic multi-objective optimization framework to evaluate trade-offs between prediction accuracy and fairness for binary classification.
A Stochastic Approximation (SA) algorithm like PF-SMG was proved to be computationally efficient to produce well-spread and sufficiently accurate Pareto fronts, when compared to the existing literature approach based on constraining the level of fairness.
One can also handle both binary and categorical multi-valued sensitive attributes, but we improve over the literature since our approach can handle more than one sensitive attribute or different fairness measures simultaneously as well as
streaming data.

The proposed framework can be generalized to accommodate different types of predictors and loss functions, including multi-class classification and regression. Moreover, our approach allows us to handle nonconvex approximations of disparate impact, equalized odds, or equal opportunity, two potential ones being mutual information~\cite{TKamishima_etal_2012} and fairness risk measures~\cite{RCWilliamson_AKMenon_2019}.


\clearpage
\bibliography{fairness}
\bibliographystyle{icml2020}

\onecolumn
\appendix
\section{The Stochastic Multi-Gradient (SMG) Algorithm}
\label{appendix:SMG} 

\begin{algorithm}[h]
   \caption{Stochastic Multi-Gradient (SMG) Algorithm}
   \label{alg:SMG}
\begin{algorithmic}
   \STATE {\bfseries Input:} an initial point $x_1 \in \mathbb{R}^n$, a step size sequence $\{\alpha_k\}_{k \in \mathbb{N}} > 0$, and maximum iterates $T$.
\FOR{$k = 1, \ldots, T$}
\STATE Compute the stochastic gradients $g_i(x_{k}, w_k)$ with batch size $b_{i, k}$ for the individual functions, $i = 1, \ldots, m$.
\STATE Solve the quadratic subproblem
\begin{equation*} \label{subproblem3}
\begin{array}{ll}
\lambda^k \;\in\;  \argmin_{\lambda \in \mathbb{R}^m} \left\| \sum^m_{i = 1} \lambda_i g_i(x_k, w_k) \right\|^2  \\[1ex]
\mbox{s.t.}  \quad \sum^m_{i = 1} \lambda_i = 1,\lambda_i \geq 0,  \forall i = 1,...,m.
\end{array}
\end{equation*}
\STATE Calculate the \textit{stochastic multi-gradient} $g(x_k, w_k) \; = \; \sum^m_{i = 1} \lambda_i^k g_i(x_k, w_k)$.
\STATE Update the iterate $x_{k + 1} = x_k - \alpha_k g(x_k, w_k)$.
\ENDFOR
\end{algorithmic}
\end{algorithm}

\section{Description and illustration of the Pareto-Front Stochastic Multi-Gradient algorithm}
\label{appendix:PF_SMG}

A formal description of the PF-SMG algorithm is given in Algorithm~\ref{alg:PF_SMG}.

\begin{algorithm}[h] 
\caption{Pareto-Front Stochastic Multi-Gradient (PF-SMG) Algorithm}
\label{alg:PF_SMG}
\begin{algorithmic} 
\STATE {\bfseries Input:} A list of starting points $\mathcal{L}_0$, and parameters $r, p_1, p_2 \in \mathbb{N}$.
\STATE {\bf for} $k=0,1, \ldots$ {\bf do}
\STATE \quad\quad Set $\mathcal{L}_{k+1} = \mathcal{L}_k$.
\STATE \quad\quad {\bf for} each point $x$ in the list $\mathcal{L}_{k+1}$ {\bf do}
\STATE \quad\quad\quad\quad Add $r$ perturbed points to the list $\mathcal{L}_{k+1}$ from a neighborhood of $x$.
\STATE \quad\quad{\bf end for}
\STATE \quad\quad {\bf for} each point $x$ in the list $\mathcal{L}_{k+1}$ {\bf do}
\STATE \quad\quad\quad\quad {\bf for} $t = 1, \ldots, p_2$ {\bf do}
\STATE \quad \quad\quad\quad\quad Apply $p_1$ iterations of the SMG algorithm starting from~$x$.
 \STATE \quad \quad\quad\quad\quad Add the final output point $x_t$ to the list $\mathcal{L}_{k+1}$.
\STATE \quad \quad\quad\quad{\bf end for}
 \STATE \quad\quad {\bf end for}
\STATE \quad\quad Remove all the dominated points from $\mathcal{L}_{k+1}$:
{\bf for} each point $x$ in the list $\mathcal{L}_{k+1}$ {\bf do}
\STATE \quad\quad \quad  If $\exists~ y \in \mathcal{L}_{k+1}$ such that $F(y) < F(x)$ holds, remove $x$ from the list.
{\bf end for}
\STATE {\bf end for}
\par\vspace*{0.1cm}
\end{algorithmic}
\end{algorithm}

\begin{figure}[ht]
  \centering
  \subfloat[][Adding perturbed points.]{\includegraphics[width=5cm]{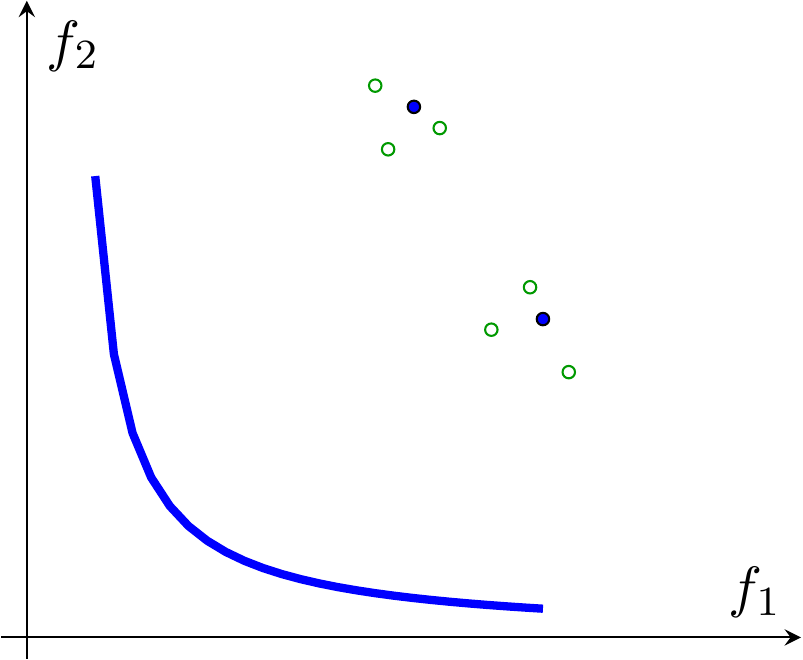}}
  \subfloat[][Applying SMG steps.]{\includegraphics[width=5cm ]{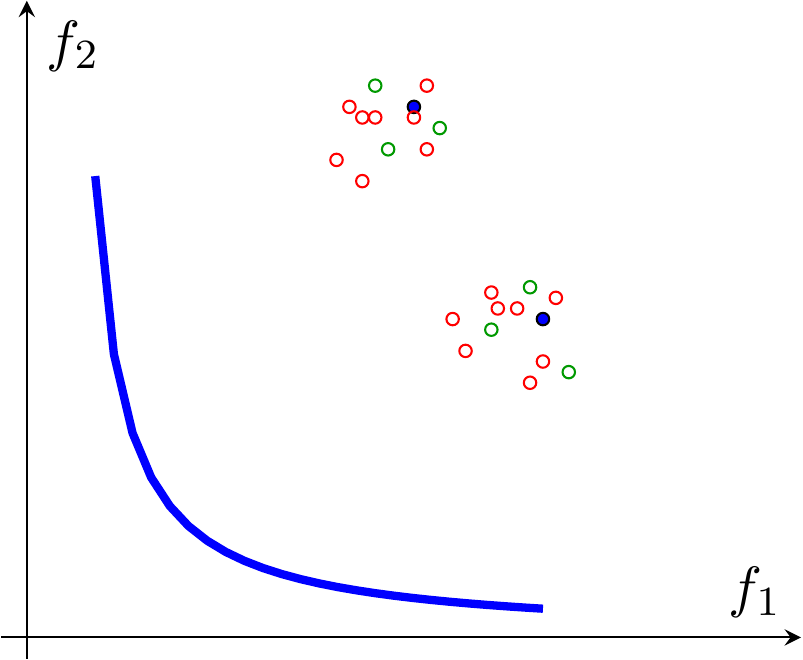}} \\ \quad
  \subfloat[][Removing dominated points.]{\includegraphics[width=5cm]{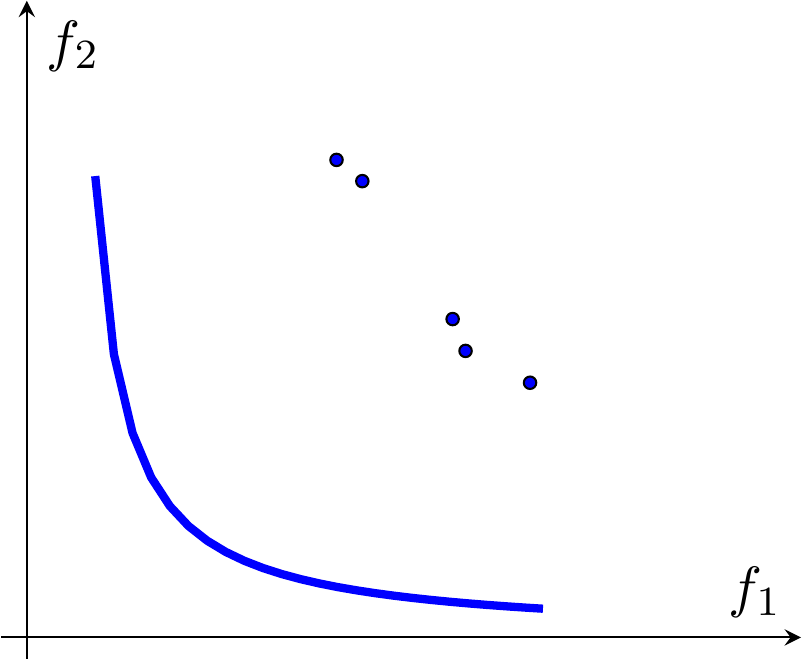}}
  \subfloat[][Moving front.]{\includegraphics[width=5cm ]{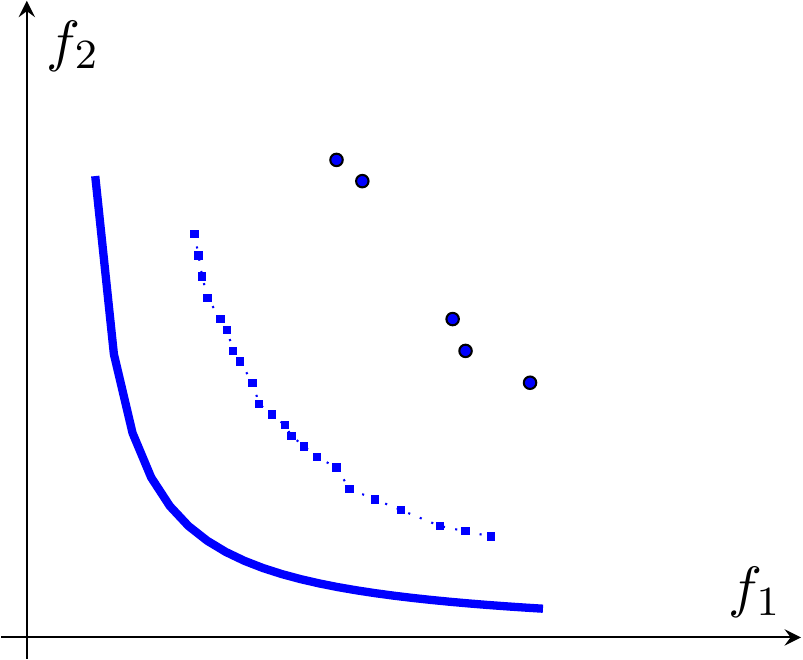}}
  \caption{Illustration of Pareto-Front stochastic multi-gradient algorithm.}
  \label{fig:pfsmg}
  \vskip-.5cm
\end{figure}

An illustration is provided in Figure~\ref{fig:pfsmg}. The blue curve represents the true Pareto front. The PF-SMG algorithm first randomly generates a list of starting feasible points (see blue points in Figure~\ref{fig:pfsmg}~(a)). For each point in the current list, a certain number of perturbed points (see green circles in Figure~\ref{fig:pfsmg}~(a)) are added to the list, after which multiple runs of the SMG algorithm are applied to each point in the current list. These newly generated points are marked by red circles in Figure~\ref{fig:pfsmg}~(b). At the end of the current iteration, a new list for the next iteration is obtained by removing all the dominated points. As the algorithm proceeds, the front will move towards the true Pareto front. 

The complexity rates to determine a point in the Pareto front using stochastic multi-gradient are reported in~\cite{SLiu_LNVicente_2019}. However, in multiobjective optimization, as far as we know, there are no convergence or complexity results to determine the whole Pareto front (under reasonable assumptions that do not reduce to evaluating the objective functions in a set that is dense in the decision space).

\section{Metrics for Pareto front comparison}
\label{appendix:metrics_comp}

Let $\mathcal{A}$ denote the set of algorithms/solvers and $\mathcal{T}$ denote the set of test problems. The Purity metric measures the accuracy of an approximated Pareto front. Let us denote $F(\mathcal{P}_{a, t})$ as an approximated Pareto front of problem $t$ computed by algorithm~$a$. We approximate the ``true'' Pareto front $F(\mathcal{P}_t)$ for problem~$t$ by all the nondominated points in $\cup_{a \in \mathcal{A}} F(\mathcal{P}_{a, t})$. Then, the Purity of a Pareto front computed by algorithm $a$ for problem $t$ is the ratio $r_{a, t} = |F(\mathcal{P}_{a, t}) \cap F(\mathcal{P}_t)|/|F(\mathcal{P}_{a, t})| \in [0, 1]$, which calculates the percentage of ``true'' nondominated solutions among all the nondominated points generated by algorithm $a$. A higher ratio value corresponds to a more accurate Pareto front.

The Spread metric is designed to measure the extent of the point spread in a computed Pareto front, which requires the computation of extreme points in the objective function space~$\mathbb{R}^m$.
Among the $m$ objective functions, we select a pair of nondominated points in $\mathcal{P}_t$ with the highest pairwise distance (measured using $f_i$) as the pair of extreme points.
More specifically, for a particular algorithm $a$, let $(x_{\min}^i, x_{\max}^i) \in \mathcal{P}_{a, t}$ denote the pair of nondominated points where $x_{\min}^i = \argmin_{x \in \mathcal{P}_{a, t}} f_i(x)$ and $x_{\max}^i = \argmax_{x \in \mathcal{P}_{a, t}} f_i(x)$. Then, the pair of extreme points is $(x_{\min}^k, x_{\max}^k)$ with $k = \argmax_{i = 1, \ldots, m} f_i(x_{\max}^i) - f_i(x_{\min}^i)$.

The first Spread formula calculates the maximum size of the holes for a Pareto front. Assume algorithm $a$ generates an approximated Pareto front with $M$ points, indexed by $1, \ldots, M$, to which the extreme points $F(x_{\min}^k)$,$F(x_{\max}^k)$ indexed by $0$ and $M+1$ are added. Denote the maximum size of the holes by $\Gamma$. We have
\begin{equation*}
\label{Spread_Gamma}
    \Gamma \;=\; \Gamma_{a, t} \;=\; \max_{i \in \{1, \ldots, m\}} \left(\max_{j \in \{1, \ldots, M\}}\{\delta_{i,j}\}\right),
\end{equation*}
where $\delta_{i,j} = f_{i,j + 1} - f_{i, j}$, and we assume each of the objective function values $f_i$ is sorted in an increasing order.

The second formula was proposed by~\cite{KDeb_etal_2002} for the case $m = 2$
(and further extended to the case $m \geq 2$ in~\cite{ALCustodio_etal_2011}) and indicates how well the points are distributed in a Pareto front. Denote the point spread by~$\Delta$. It is computed by the following formula:
\begin{equation*}
\label{Spread_Delta}
    \Delta \;=\; \Delta_{a, t} \;=\; \max_{i \in \{1, \ldots, m\}} \left(\frac{\delta_{i, 0} + \delta_{i, M} + \sum_{j = 1}^{M-1}|\delta_{i, j} - \bar{\delta}_i|}{\delta_{i, 0} + \delta_{i, M} + (M-1)\bar{\delta}_i} \right),
\end{equation*}
where $\bar{\delta}_i, i = 1, \ldots, m$ is the average of $\delta_{i, j}$ over $j = 1, \ldots, M -1$. Note that the lower $\Gamma$ and $\Delta$ are, the more well distributed the Pareto front is.

\begin{figure}[ht]
  \centering
  \includegraphics[width=5cm]{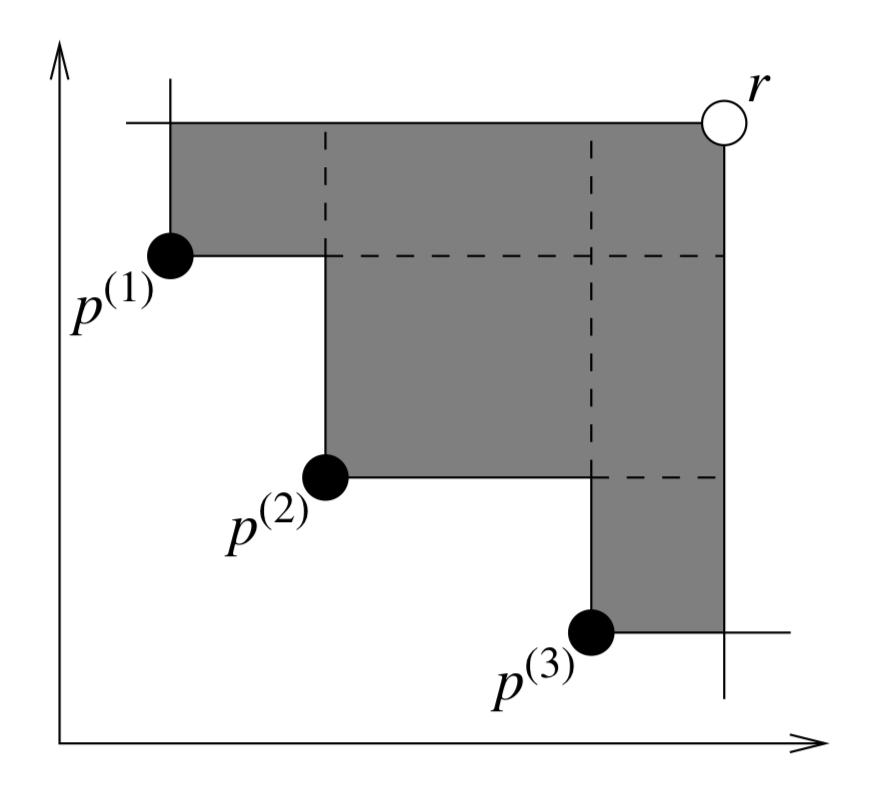}
  \caption{Illustration of hypervolume using a bi-objective example~\cite{CMFonseca_etal_2006}.}
  \label{fig:hypervolume}
  \vskip-.5cm
\end{figure}

Hypervolume~\cite{EZitzler_LThiele_1999} is another classical performance indicator taking into account both the quality of the individual Pareto points and also their overall objective space coverage. It essentially calculates the area/volume dominated by the provided set of nondominated solutions with respect to a reference point. Figure~\ref{fig:hypervolume} demonstrates a bi-objective case where the area dominated by a set of points $\{p^{(1)}, p^{(2)}, p^{(3)}\}$ with respect to the reference point $r$ is shown in grey.
In our experiments, we calculate hypervolume using the Pymoo package (see \url{https://pymoo.org/misc/indicators.html}).

\section{Datasets generation and pre-processing}
\label{appendix_datasets}
The synthetic data is formed by $20$ sets of 2,000 binary classification data instances randomly generated from the same distributions setting specified in~\citet[Section~4]{MBZafar_etal_2017b}, specifically using an uniform distribution for generating binary labels $Y$, two different Gaussian distributions for generating 2-dimensional nonsensitive features $Z$, and a Bernoulli distribution for generating the binary sensitive attribute $A$.

The data pre-processing details for the Adult Income dataset are given below.
\begin{enumerate}
    \item First, we combine all instances in \textit{adult.data} and \textit{adult.test} and remove those that values are missing for some attributes.
    \item We consider the list of features: Age, Workclass,  Education, Education number, Martial Status, Occupation, Relationship, Race, Sex, Capital gain, Capital loss, Hours per week, and Country. In the same way as the authors~\cite{MBZafar_etal_2017} did for attribute Country, we reduced its dimension by merging all non-United-Stated countries into one group. Similarly for feature Education, where ``Preschool'', ``1st-4th'', ``5th-6th'', and ``7th-8th'' are merged into one group, and ``9th'', ``10th'', ``11th'', and ``12th'' into another.
    \item Last, we did one-hot encoding for all the categorical attributes, and we normalized attributes of continuous value.
\end{enumerate}

\label{data_tables}

\begin{table}[H]
    \centering
    \begin{minipage}{.5\textwidth}
        \centering
    \caption{Adult Income dataset: Gender
    \label{Adult_dataset_gender}}
    \vskip 0.15in
    \begin{small}
    \begin{sc}
    \begin{tabular}{c|c|c|c}
    \toprule
    Gender & $\leq 50 K$ & $> 50K$ & Total  \\ \midrule
     Males & $20, 988$ & $9, 539$ & $30, 527$ \\
     Females & $13, 026$ & $1, 669$ & $14, 695$  \\
     \hline
     Total & $34, 014$ & $11, 208$ & $45, 222$ \\ \bottomrule
    \end{tabular}
    \end{sc}
    \vskip -0.1in
    \end{small}
    \end{minipage}%
    \begin{minipage}{0.5\textwidth}
        \centering
    \caption{Adult Income dataset: Race
    \label{Adult_dataset_race}}
    \vskip 0.15in
    \begin{small}
    \begin{sc}
      \begin{tabular}{c|c|c|c}
     \toprule
     Race & $\leq 50 K$ & $> 50K$ & Total  \\ \midrule
     Asian & $934$ & $369$ & $1, 303$ \\
     American-Indian & $382$ & $53$ & $435$ \\
     White & $28, 696$ & $10, 207$ & $38, 903$\\
     Black & $3, 694$ & $534$ & $4, 228$ \\
     Other & $308$ & $45$ & $353$ \\
     \hline
     Total & $34, 014$ & $11, 208$ & $45, 222$ \\
     \bottomrule
    \end{tabular}
    \end{sc}
    \vskip -0.1in
    \end{small}
    \end{minipage}
\end{table}



\begin{table}[H]
    \centering
    \caption{COMPAS dataset: Race
    \label{COMPAS_dataset_race}}
    \vskip 0.15in
\begin{small}
\begin{sc}
    \begin{tabular}{c|c|c|c}
    \toprule
    race & reoffend & Not reoffend & Total  \\ \midrule
     White & $822$ & $1, 281$ & $2, 103$  \\
     Black &  $1, 661$ & $1, 514$ & $3, 175$ \\
     \hline
     Total & $2,483$ & $2, 795$ & $5, 278$ \\ \bottomrule
    \end{tabular}
\end{sc}
\vskip -0.1in
\end{small}
\end{table}

In terms of gender, the dataset contains $67.5\%$ males ($31.3\%$ high income) and $32.5\%$ females ($11.4\%$ high income).
Similarly, the demographic compositions in terms of race are $2.88\%$ Asian ($28.3\%$), $0.96\%$ American-Indian ($12.2\%$), $86.03\%$ White ($26.2\%$), $9.35\%$ Black ($1.2\%$), and $0.78\%$ Other ($12.7\%$), where the numbers in brackets are the percentages of high-income instances.

\section{More numerical results}
\label{appendix_more_res}
\subsection{Disparate impact w.r.t. binary sensitive attribute}
\label{more_res:DI_binary}
\begin{figure}[H]
  \centering
  \subfloat[Seed $1$.]{\includegraphics[width = 0.33\textwidth]{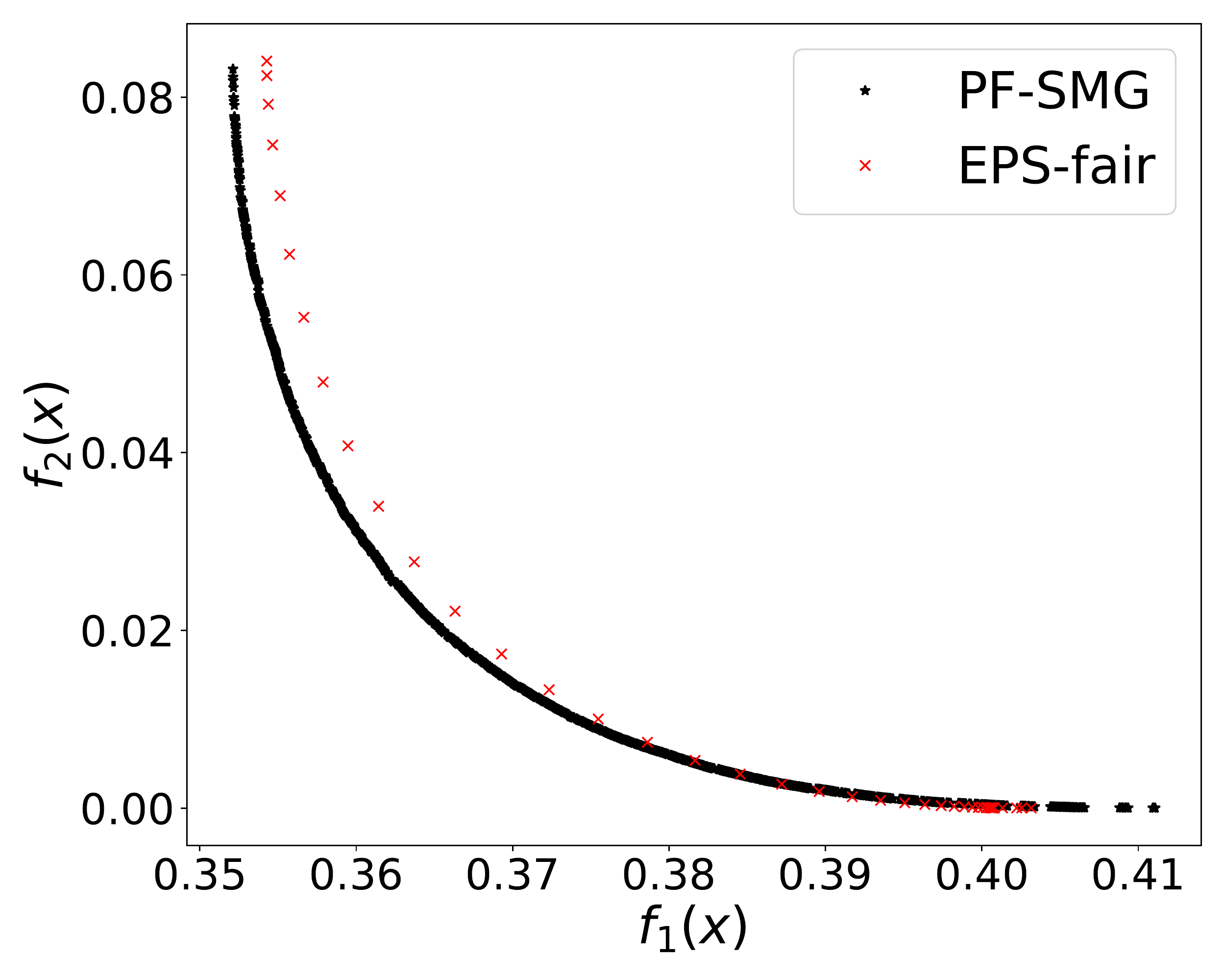}}
  \subfloat[Seed $2$.]{\includegraphics[width = 0.33\textwidth]{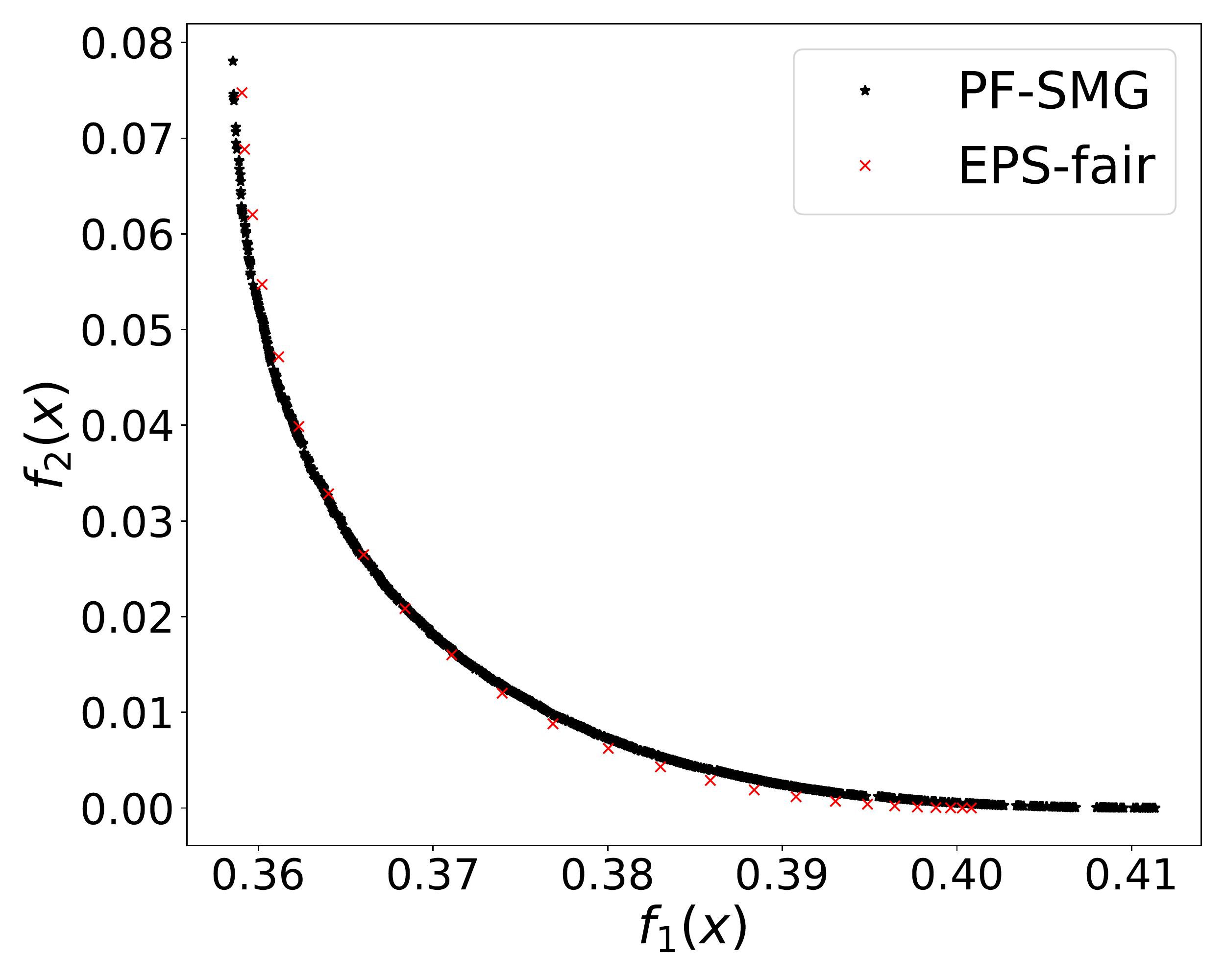}}
  \subfloat[Seed $3$.]{\includegraphics[width = 0.33\textwidth]{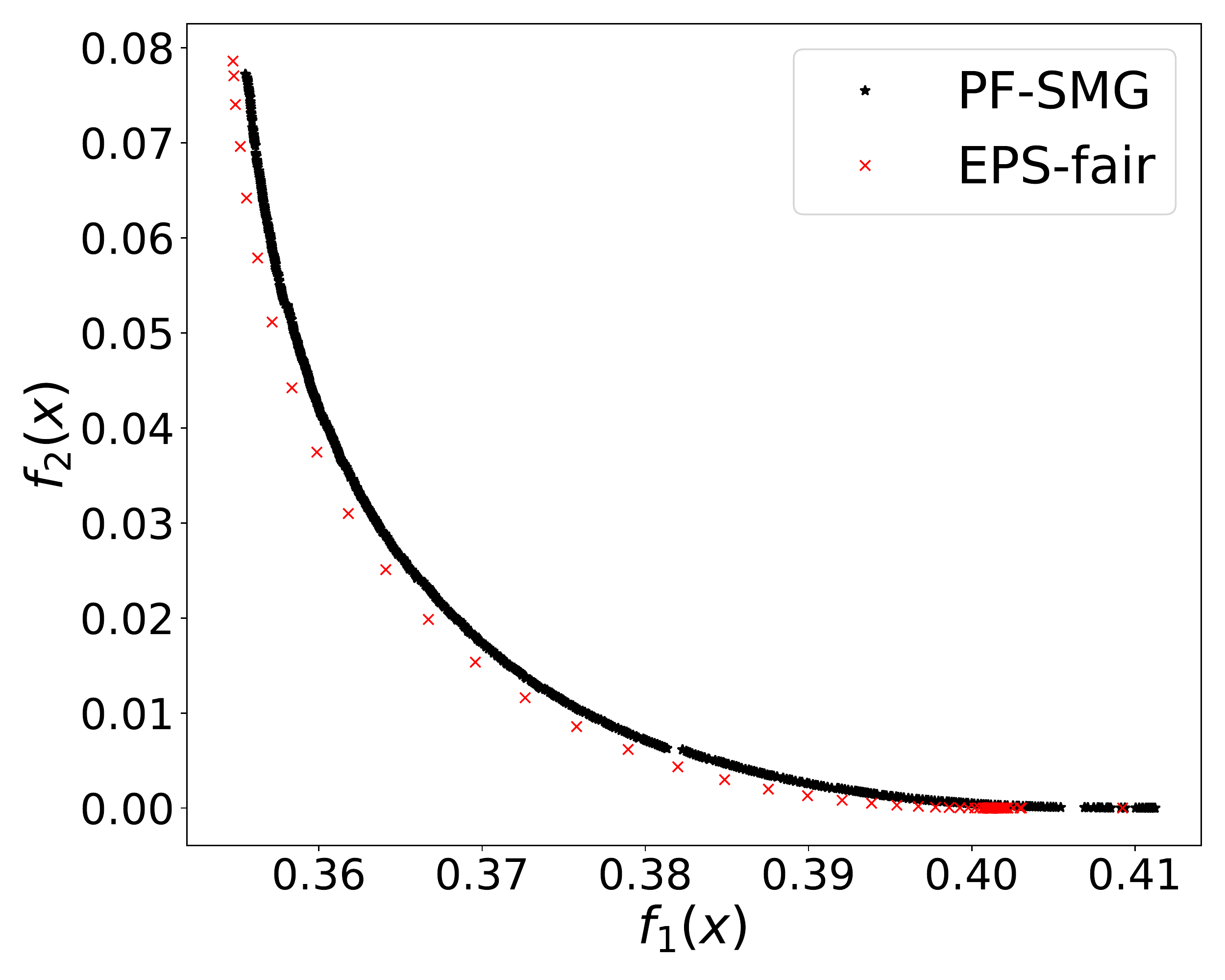}} \\
  \subfloat[Seed $4$.]{\includegraphics[width = 0.33\textwidth]{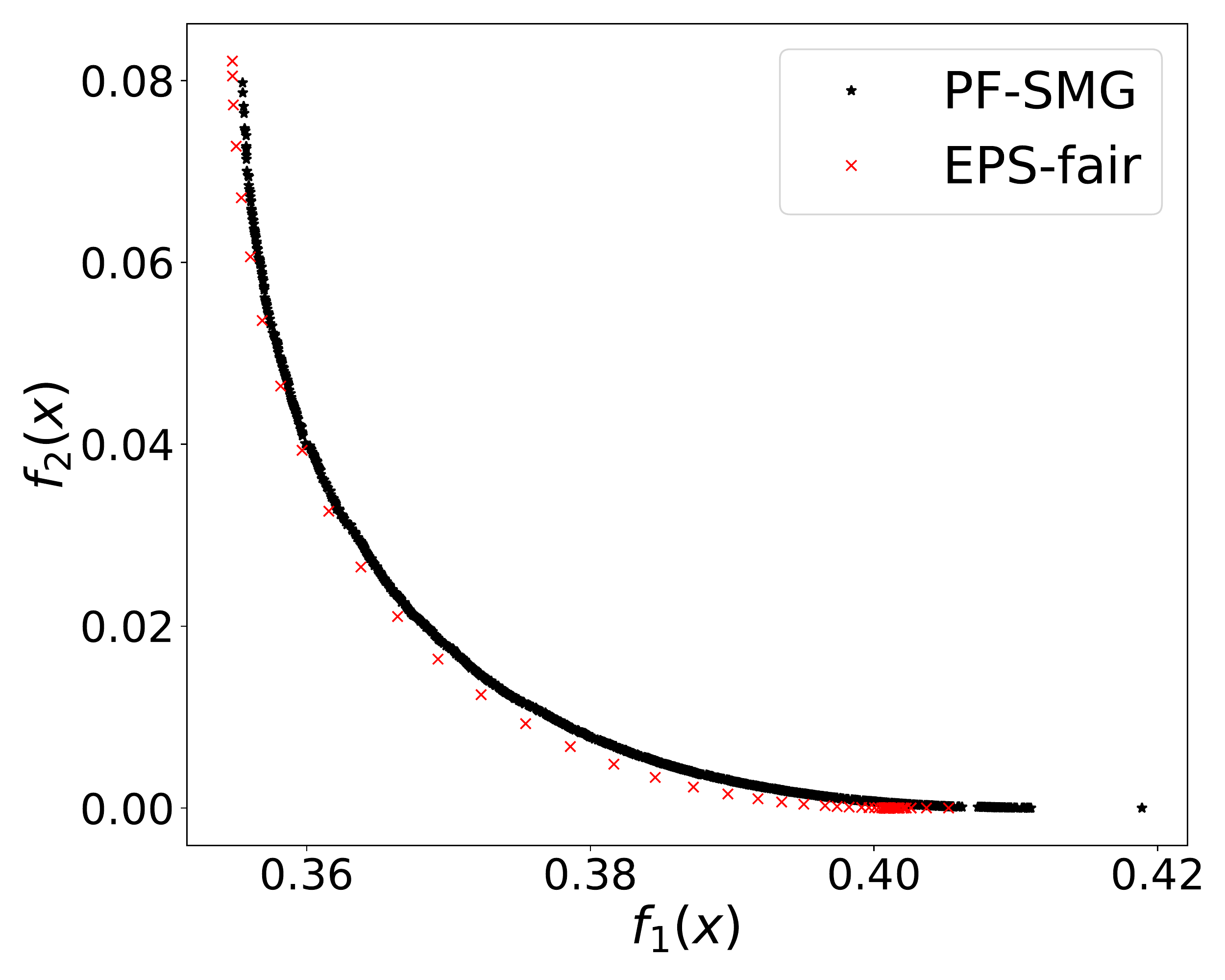}}
  \subfloat[Seed $5$.]{\includegraphics[width = 0.33\textwidth]{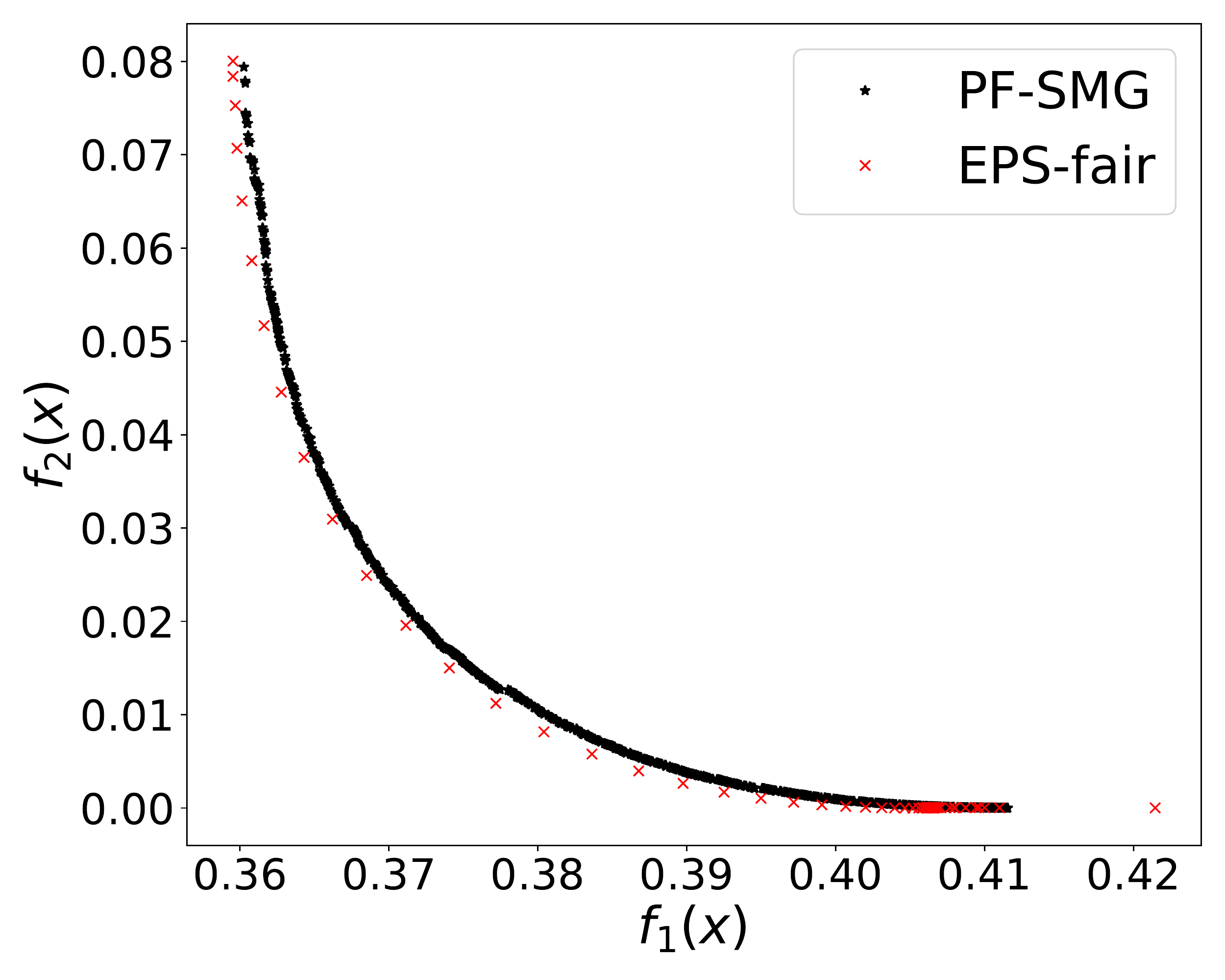}}
  \subfloat[Seed $6$.]{\includegraphics[width = 0.33\textwidth]{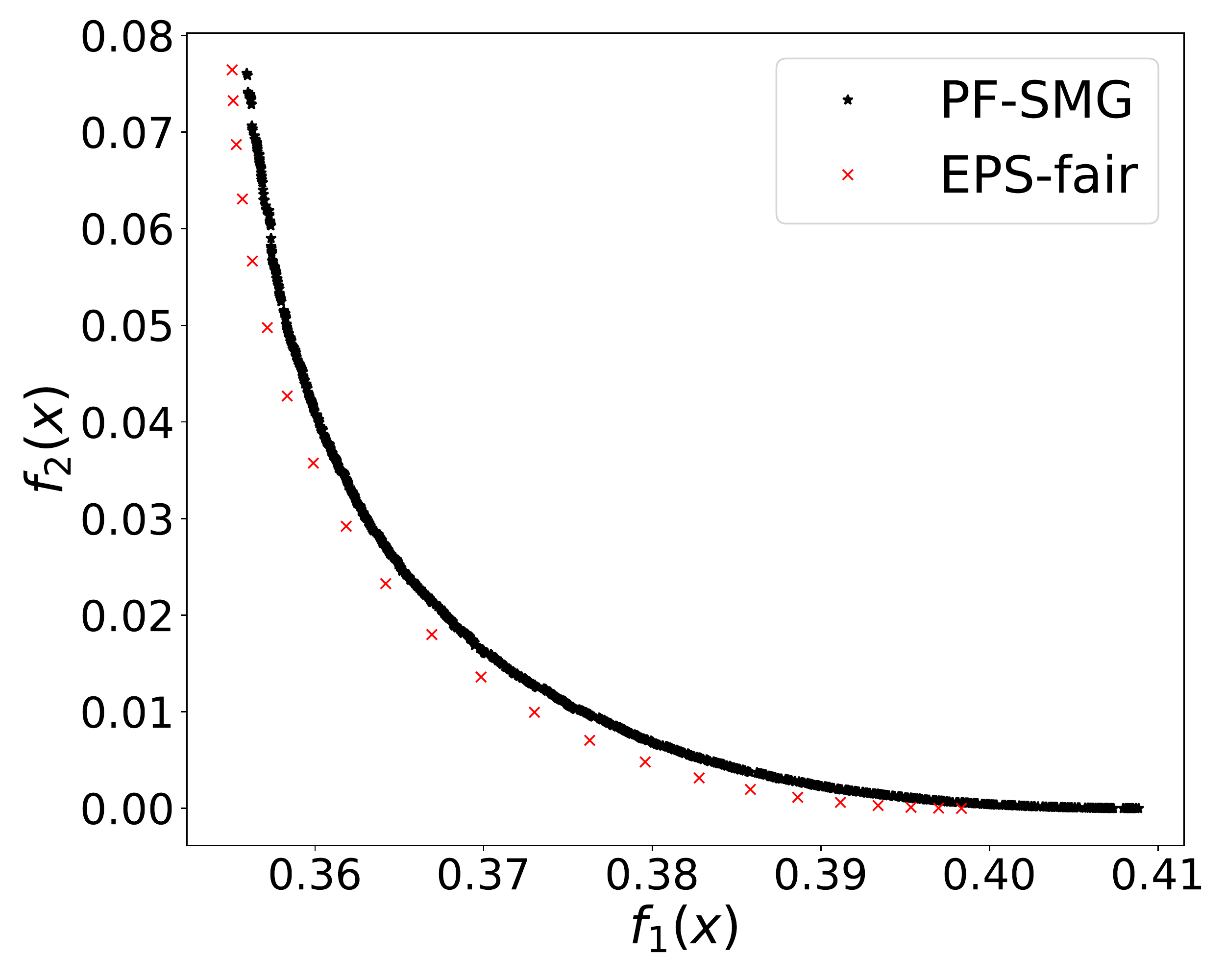}}\\
  \subfloat[Seed $7$.]{\includegraphics[width = 0.33\textwidth]{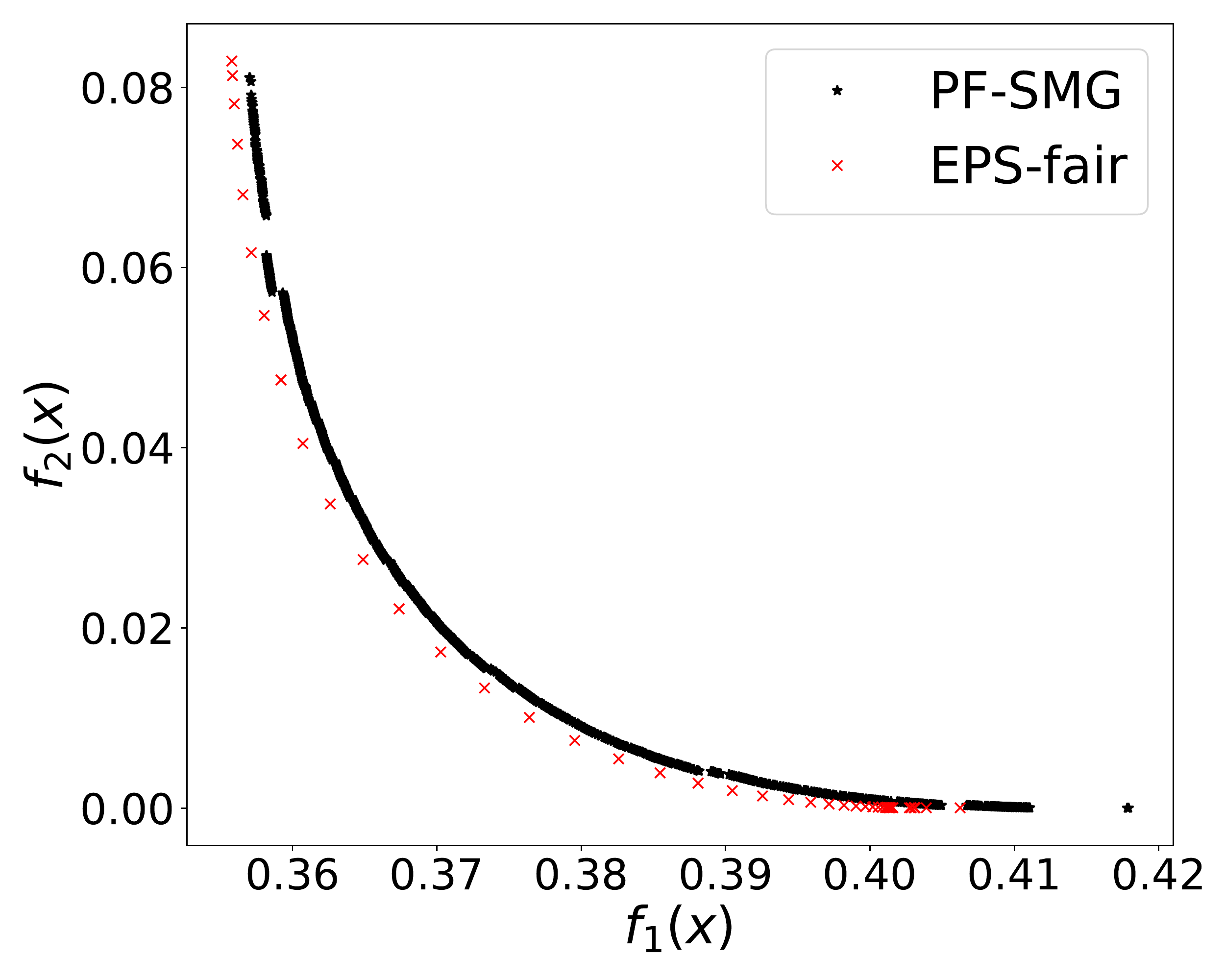}}
  \subfloat[Seed $8$.]{\includegraphics[width = 0.33\textwidth]{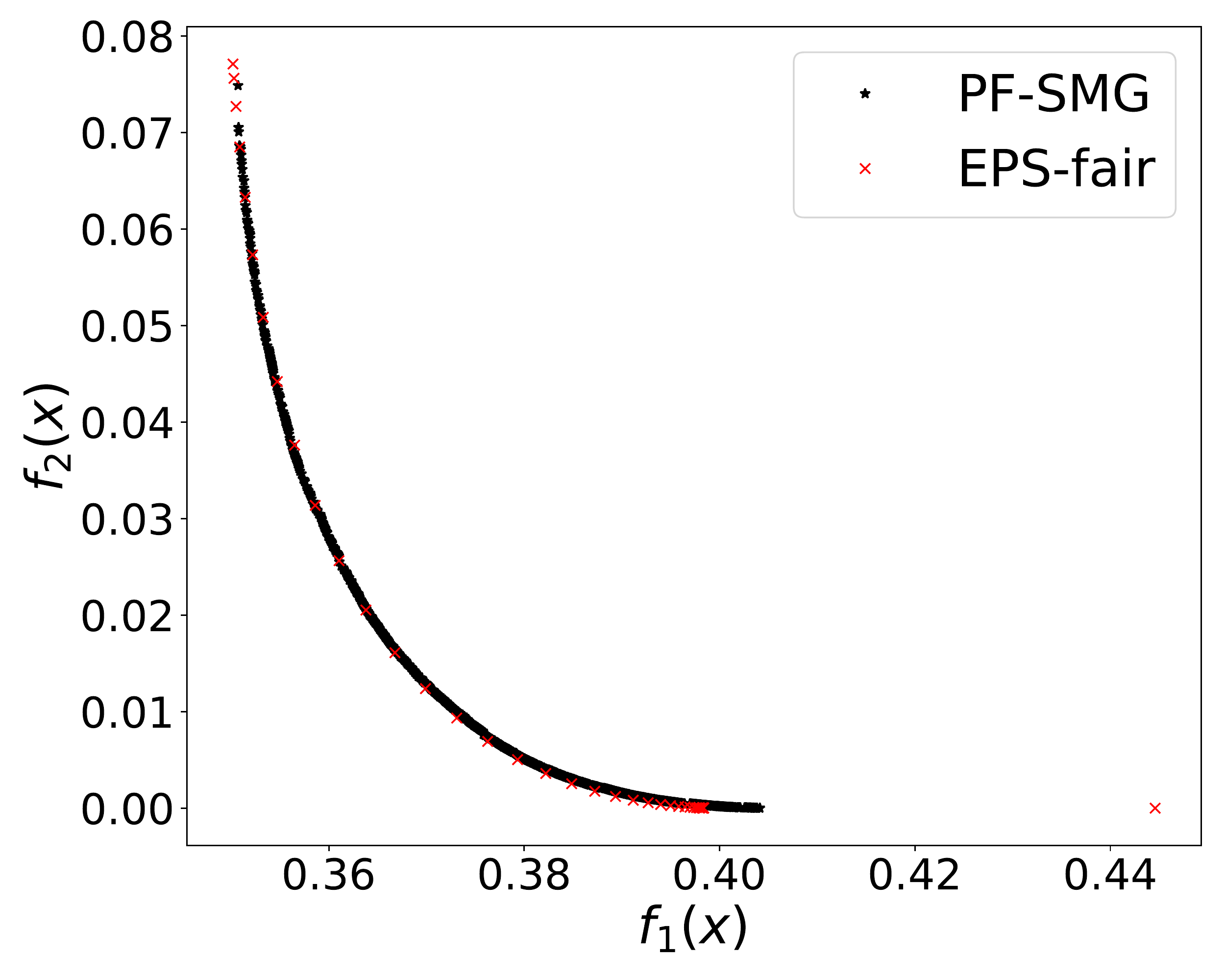}}
  \subfloat[Seed $9$.]{\includegraphics[width = 0.33\textwidth]{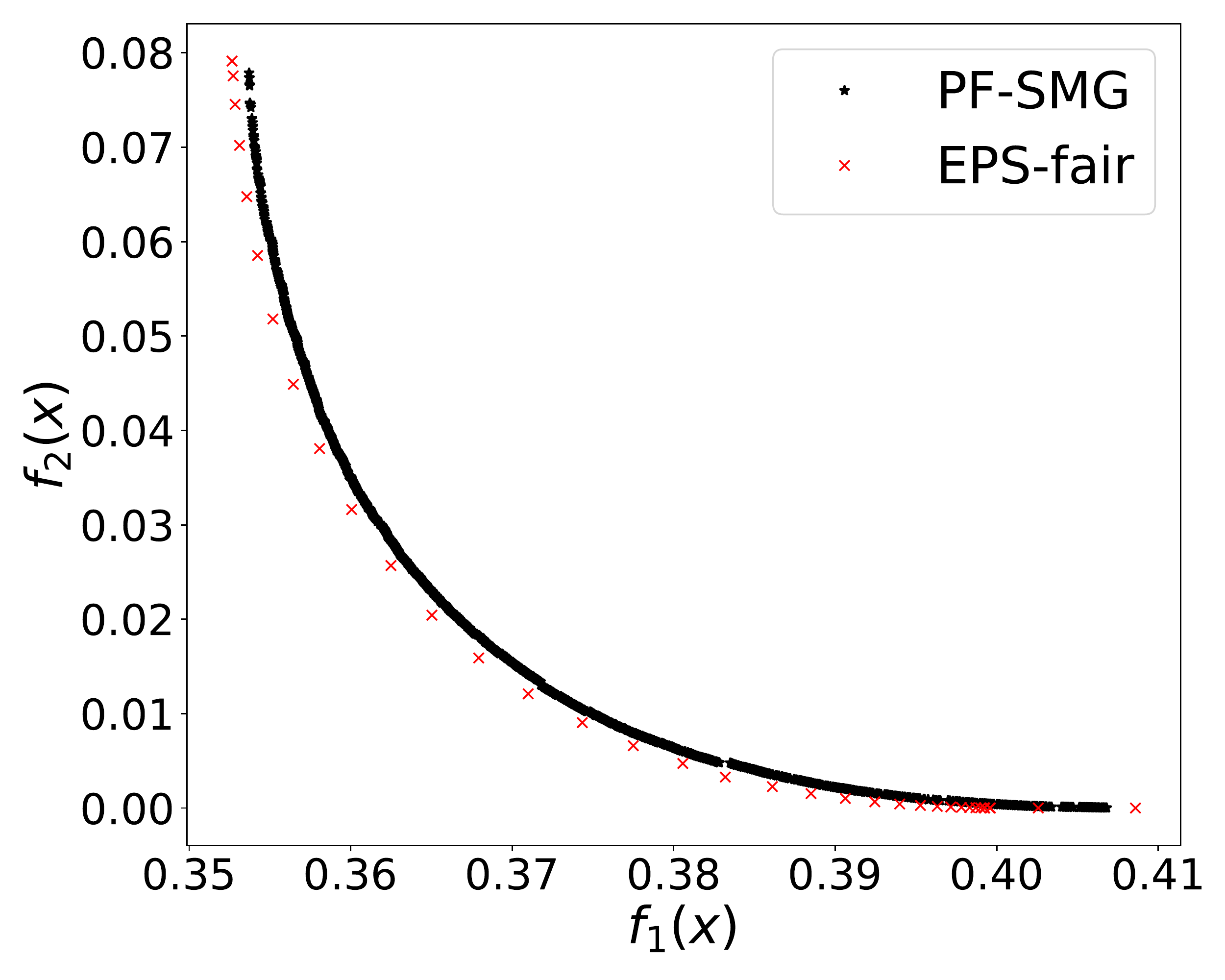}}
  \caption{Pareto front comparison for Adult Income dataset w.r.t. gender. Parameters used in PF-SMG: $p_1=2, p_2 = 3, \alpha_0 = 2.1$ and then multiplied by $1/3$ every $500$ iterates of SMG,  $b_{1, k} = 80\times 1.018^k$, and $b_{2, k} = 50\times 1.018^k$. \label{res:Adult_gender_nineSeeds}}
\end{figure}

\subsection{Disparate impact w.r.t. multi-valued sensitive attribute}
\label{appendix_DI_multi}
\begin{figure}[H]
  \centering
  \subfloat[Seed $1$.]{\includegraphics[width = 0.33\textwidth]{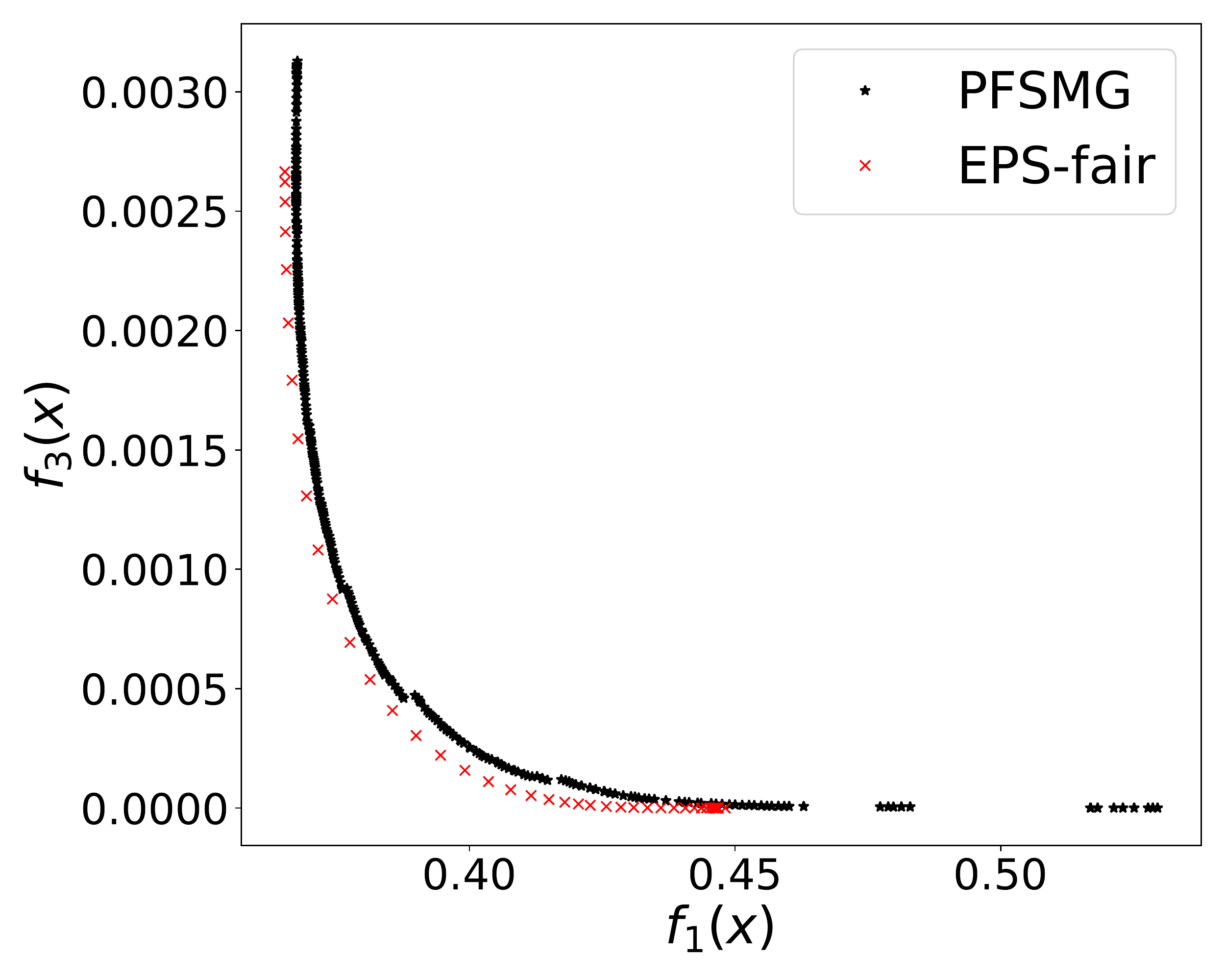}}
  \subfloat[Seed $2$.]{\includegraphics[width = 0.33\textwidth]{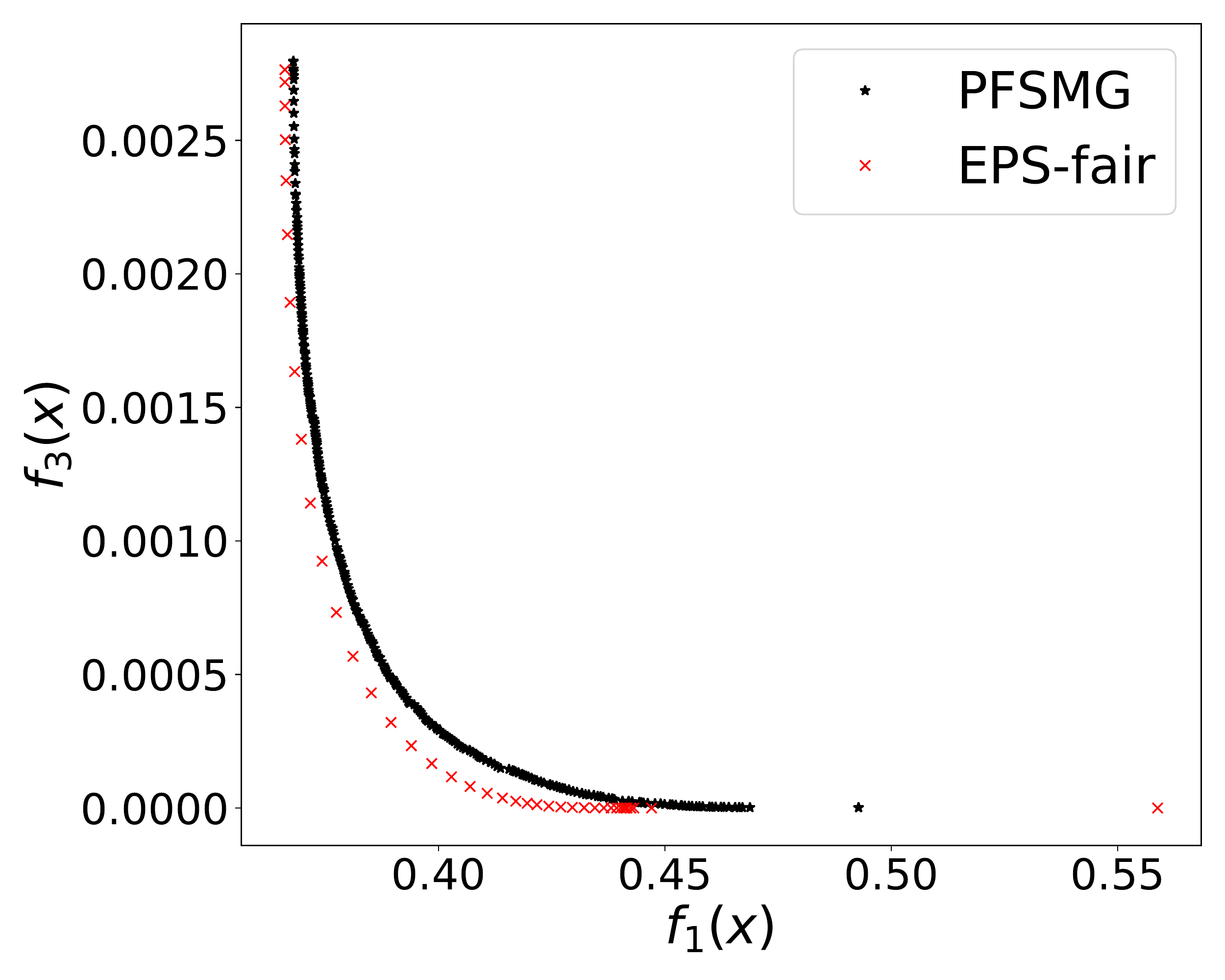}}
  \subfloat[Seed $3$.]{\includegraphics[width = 0.33\textwidth]{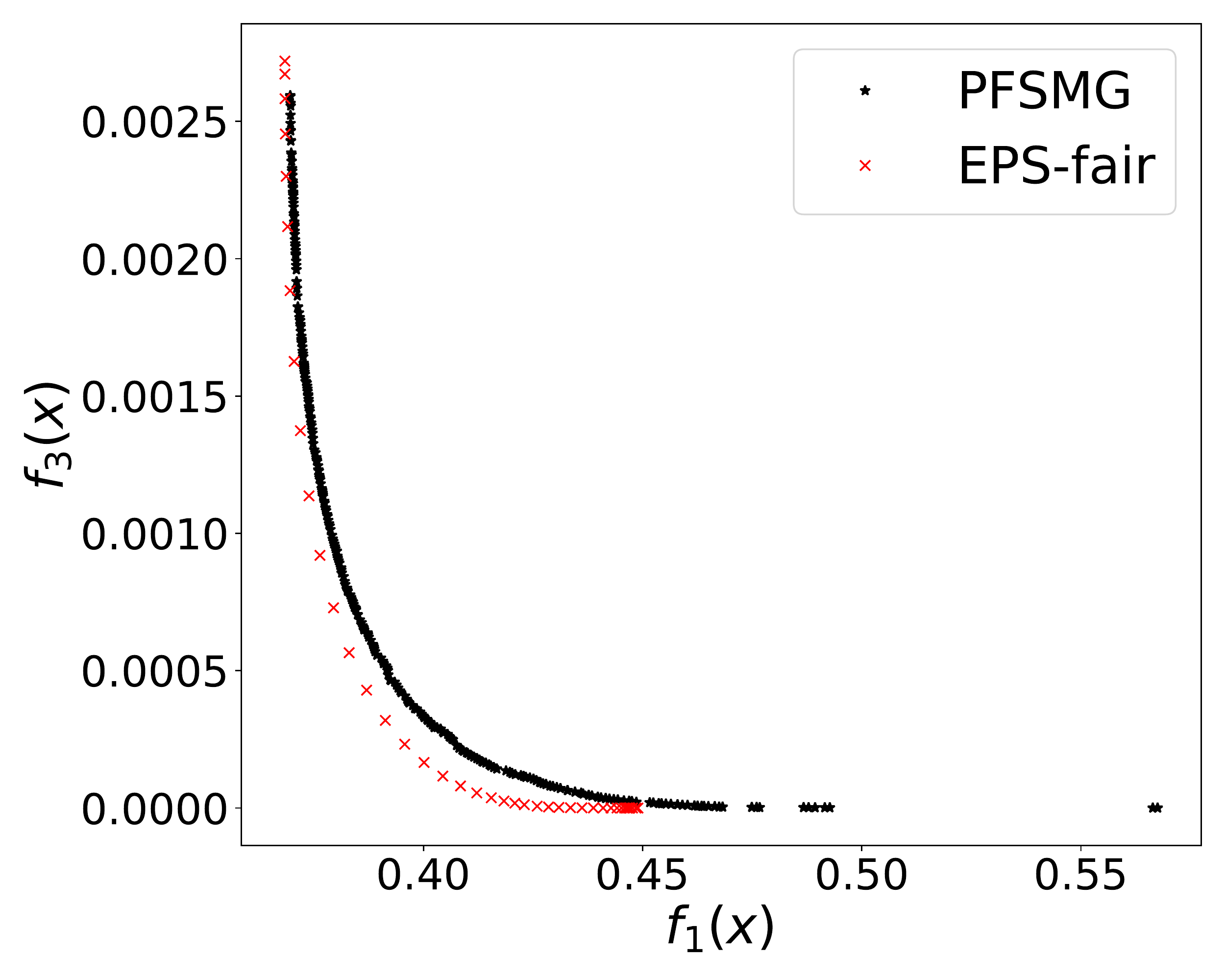}} \\
  \subfloat[Seed $4$.]{\includegraphics[width = 0.33\textwidth]{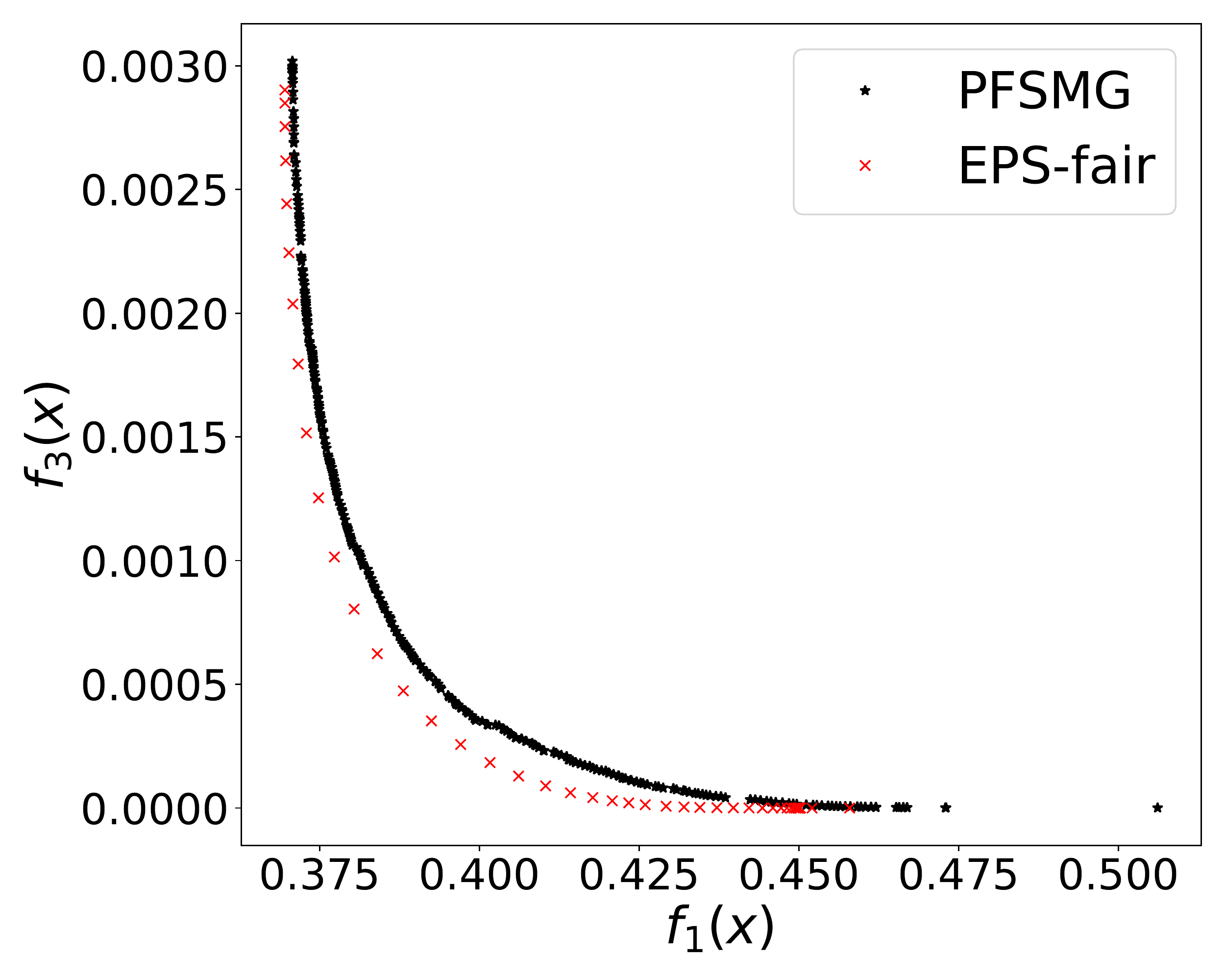}}
  \subfloat[Seed $5$.]{\includegraphics[width = 0.33\textwidth]{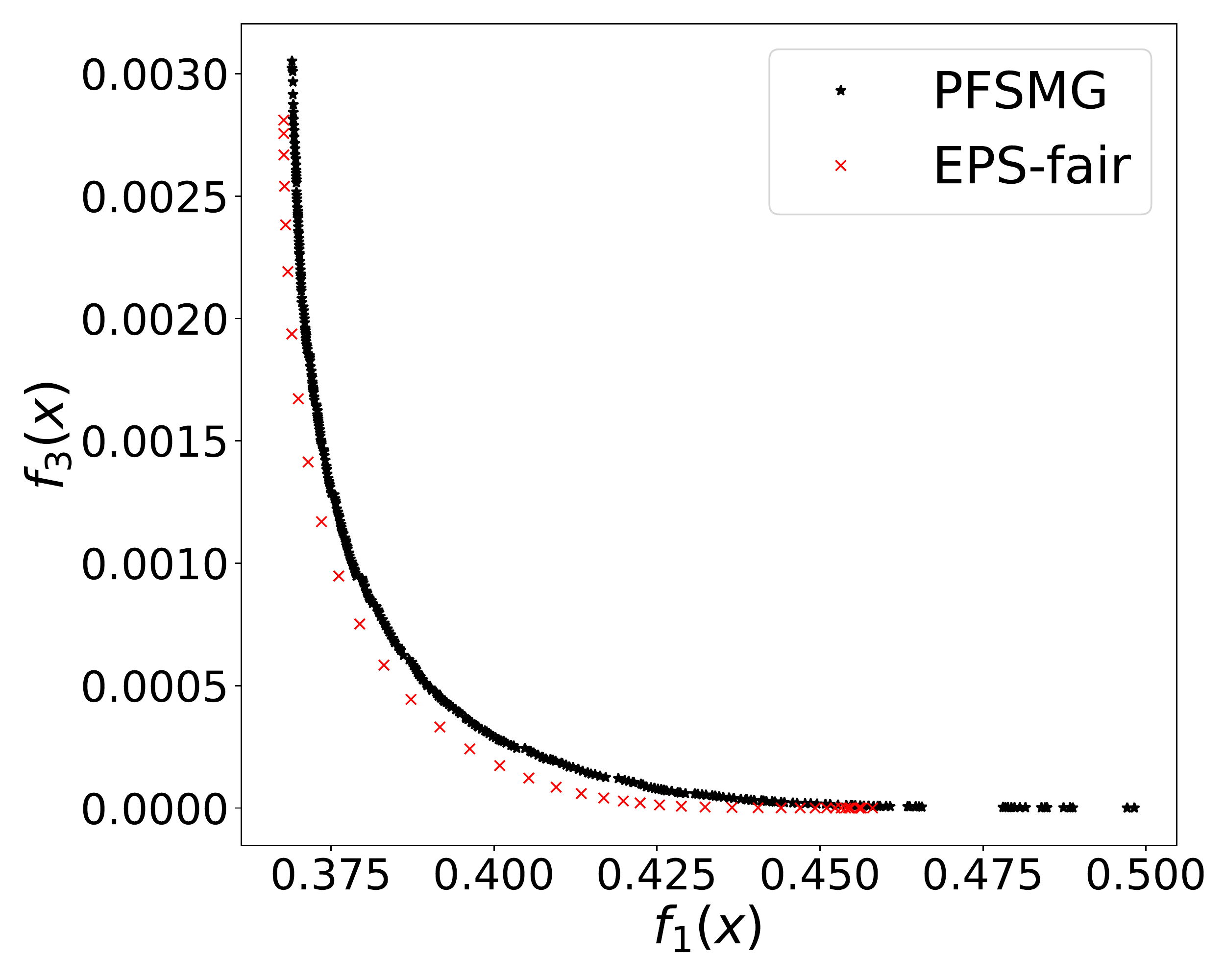}}
  \subfloat[Seed $6$.]{\includegraphics[width = 0.33\textwidth]{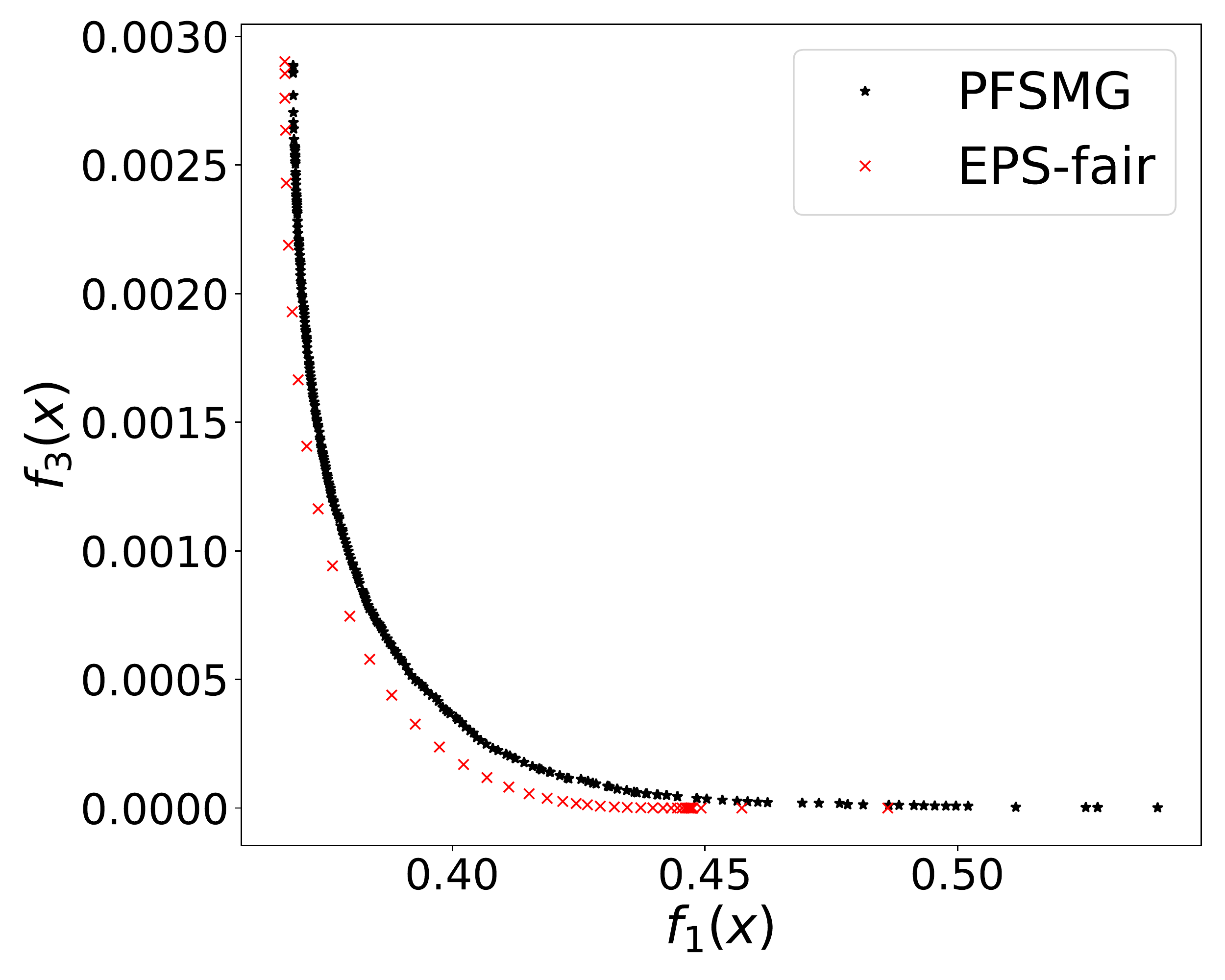}}\\
  \subfloat[Seed $7$.]{\includegraphics[width = 0.33\textwidth]{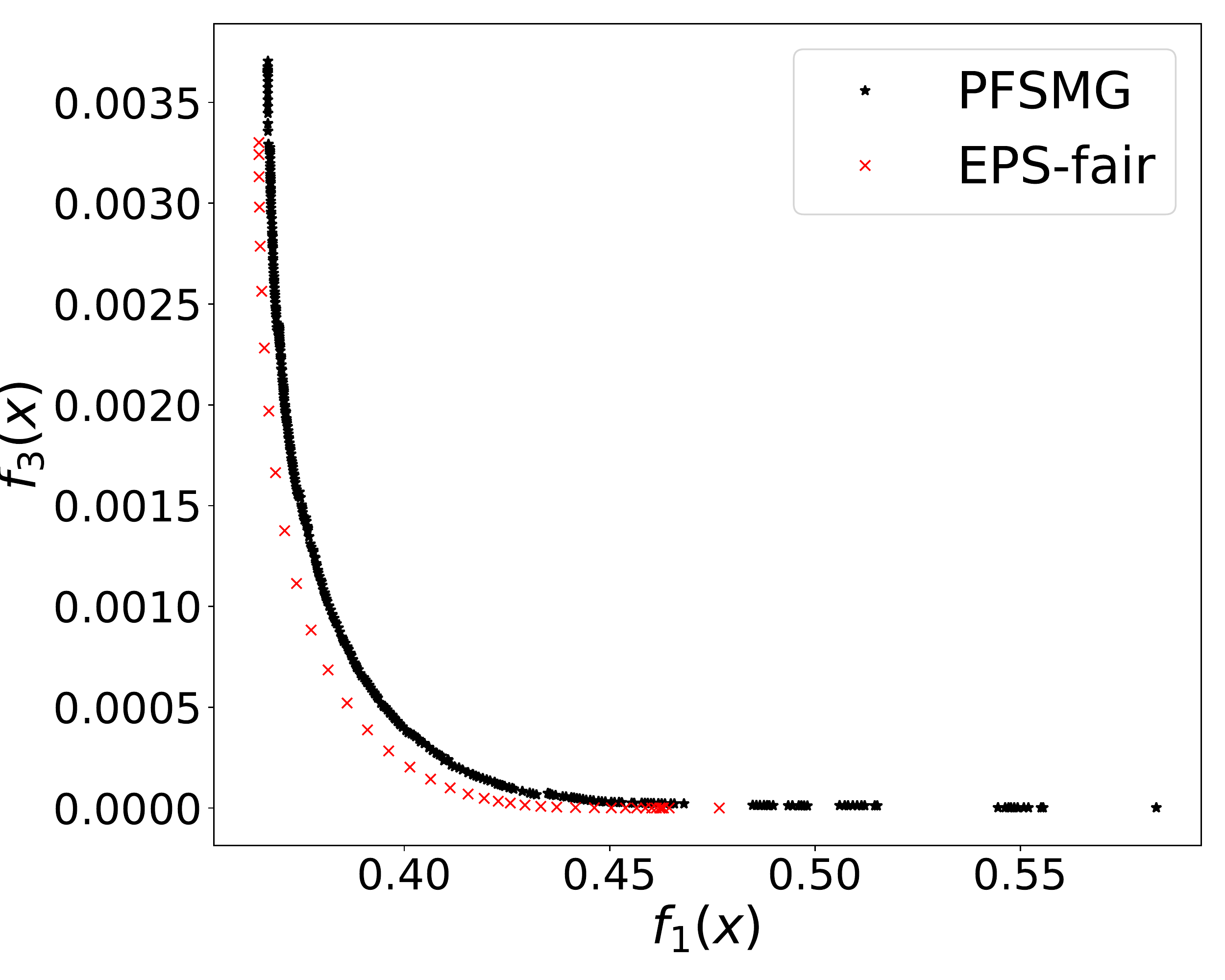}}
  \subfloat[Seed $8$.]{\includegraphics[width = 0.33\textwidth]{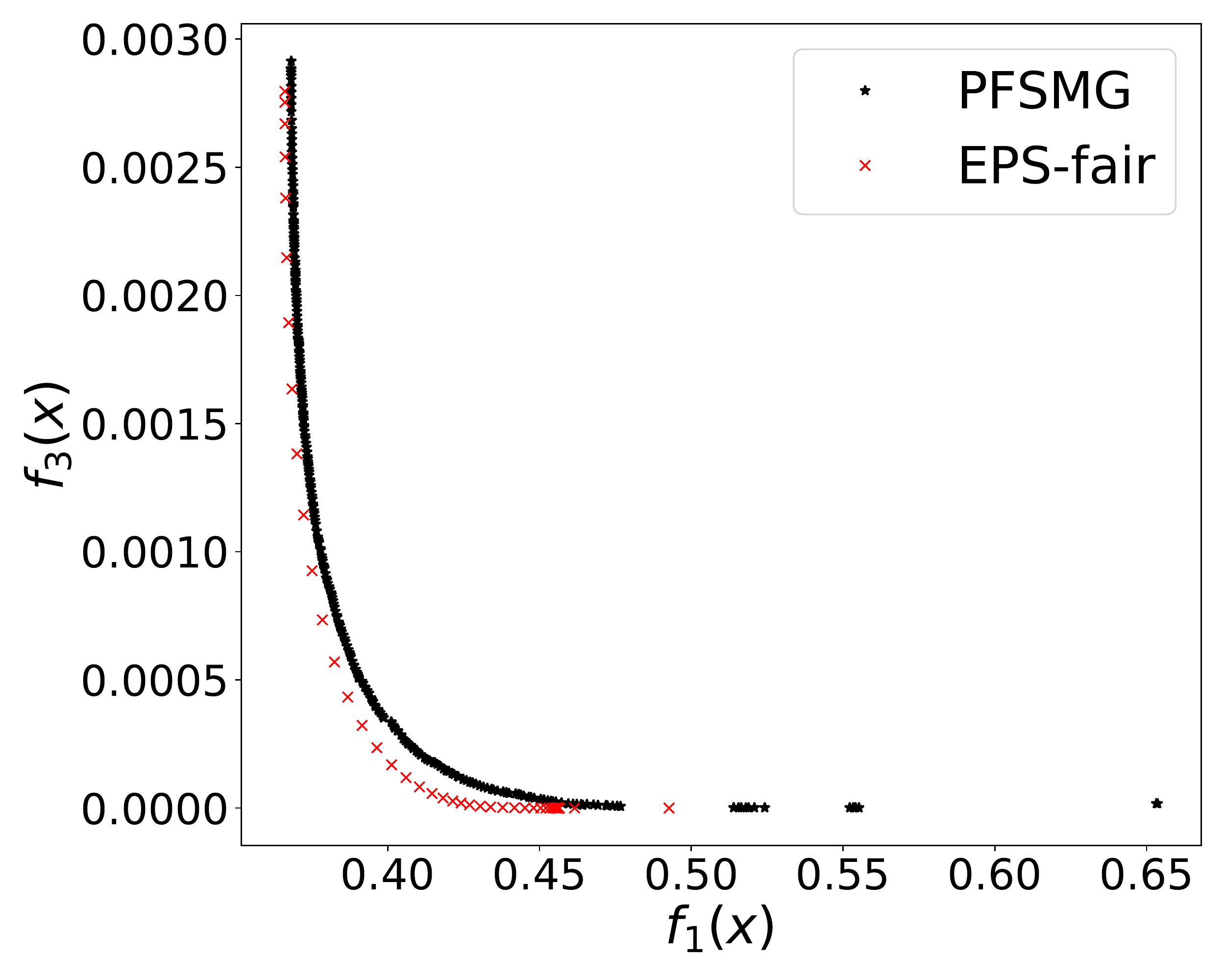}}
  \subfloat[Seed $9$.]{\includegraphics[width = 0.33\textwidth]{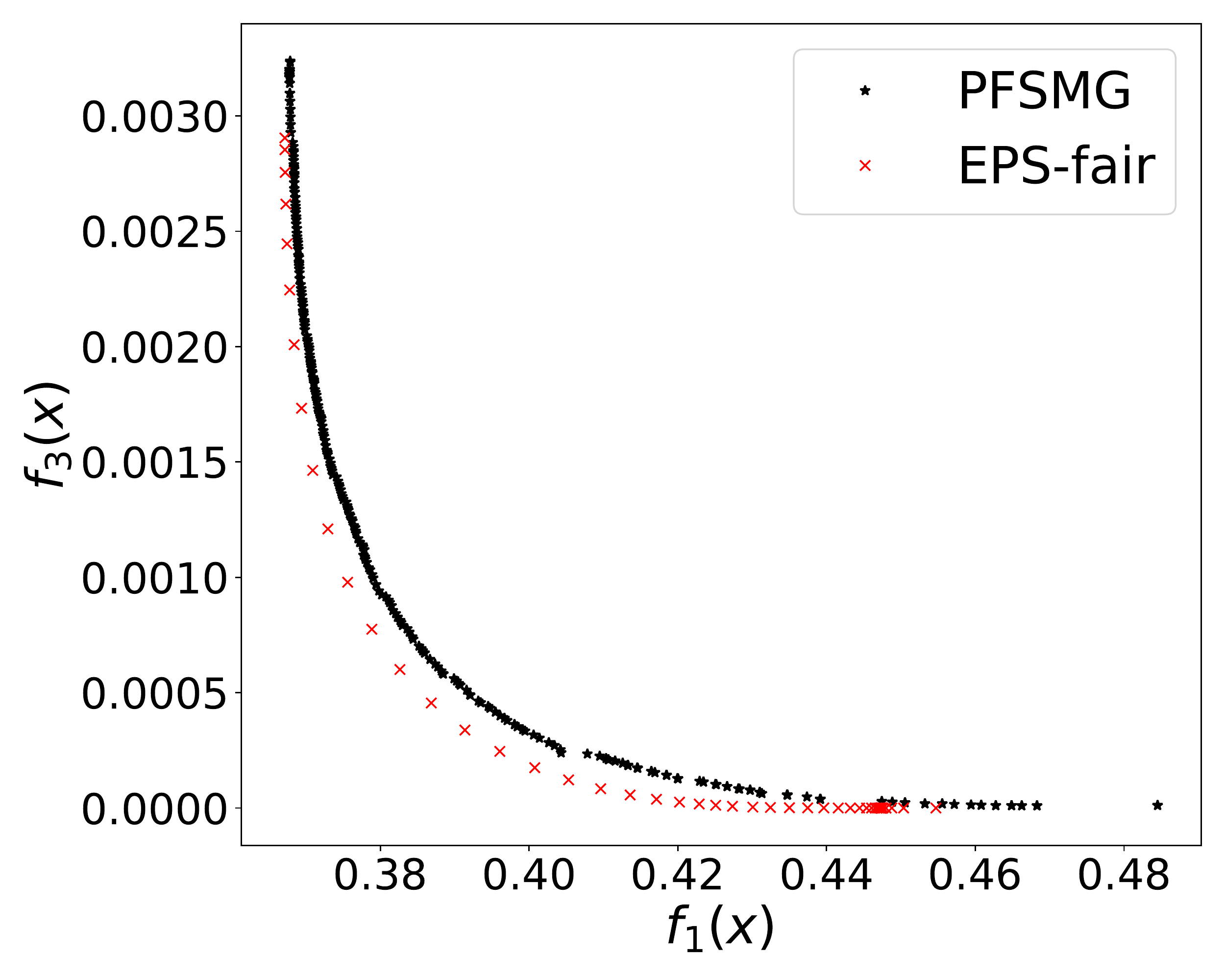}}
  \caption{Pareto front comparison for Adult dataset w.r.t. race. Parameters used in PF-SMG: $p_1 = 3, p_2 = 2, \alpha_0 = 3.0$ and multiplied by $1/3$ every $100$ iterates of SMG, $b_{1, k} = 50\times 1.012^k$, and $b_{2, k} = 30\times 1.012^k$. \label{res:Adult_race_nineSeeds}}
\end{figure}

\subsection{Streaming data}
\label{appendix:streaming_data}
\begin{figure}[H]
  \centering
  \subfloat[2,000 samples.]{\includegraphics[width = 0.33\textwidth]{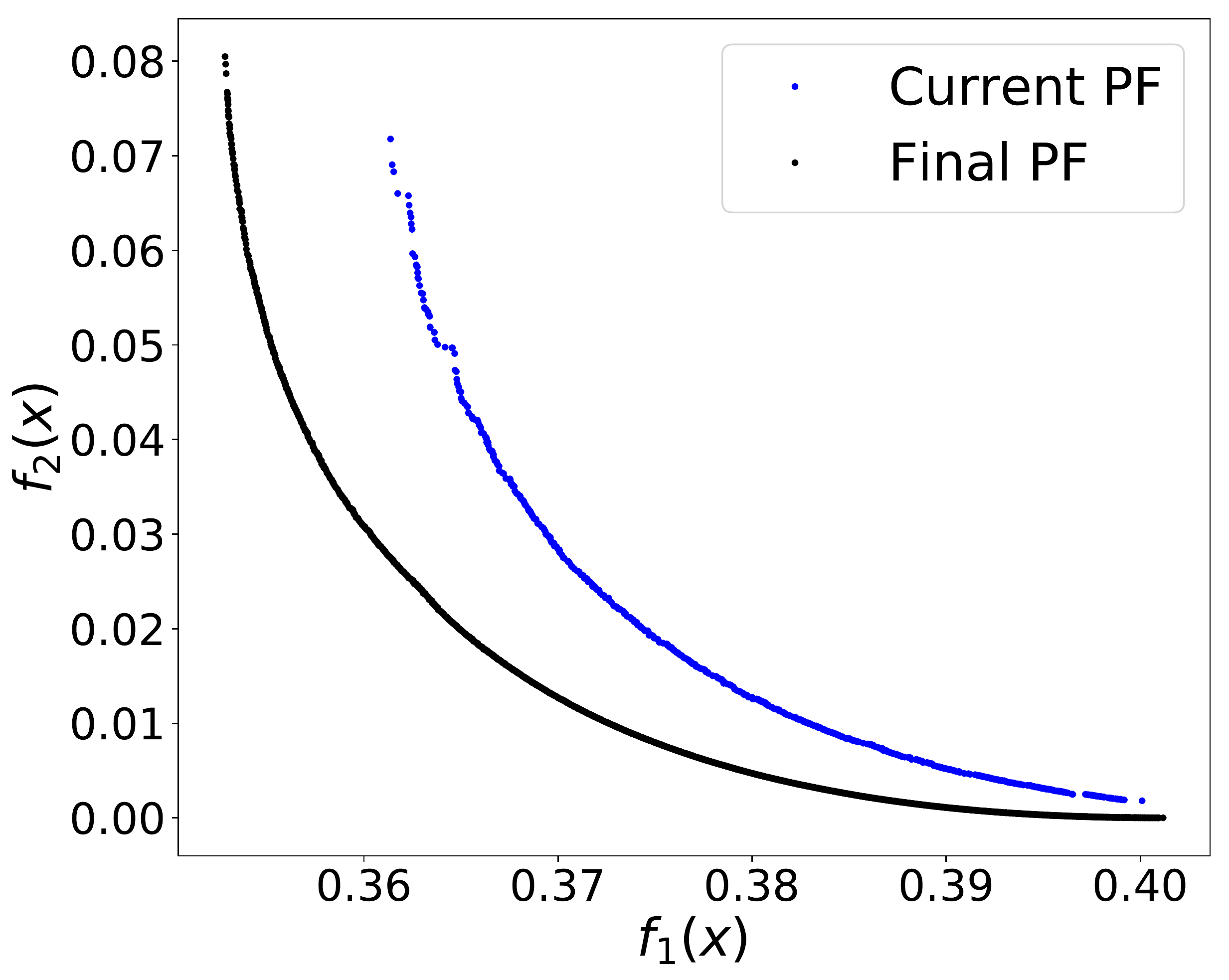}}
  \subfloat[4,000 samples.]{\includegraphics[width = 0.33\textwidth]{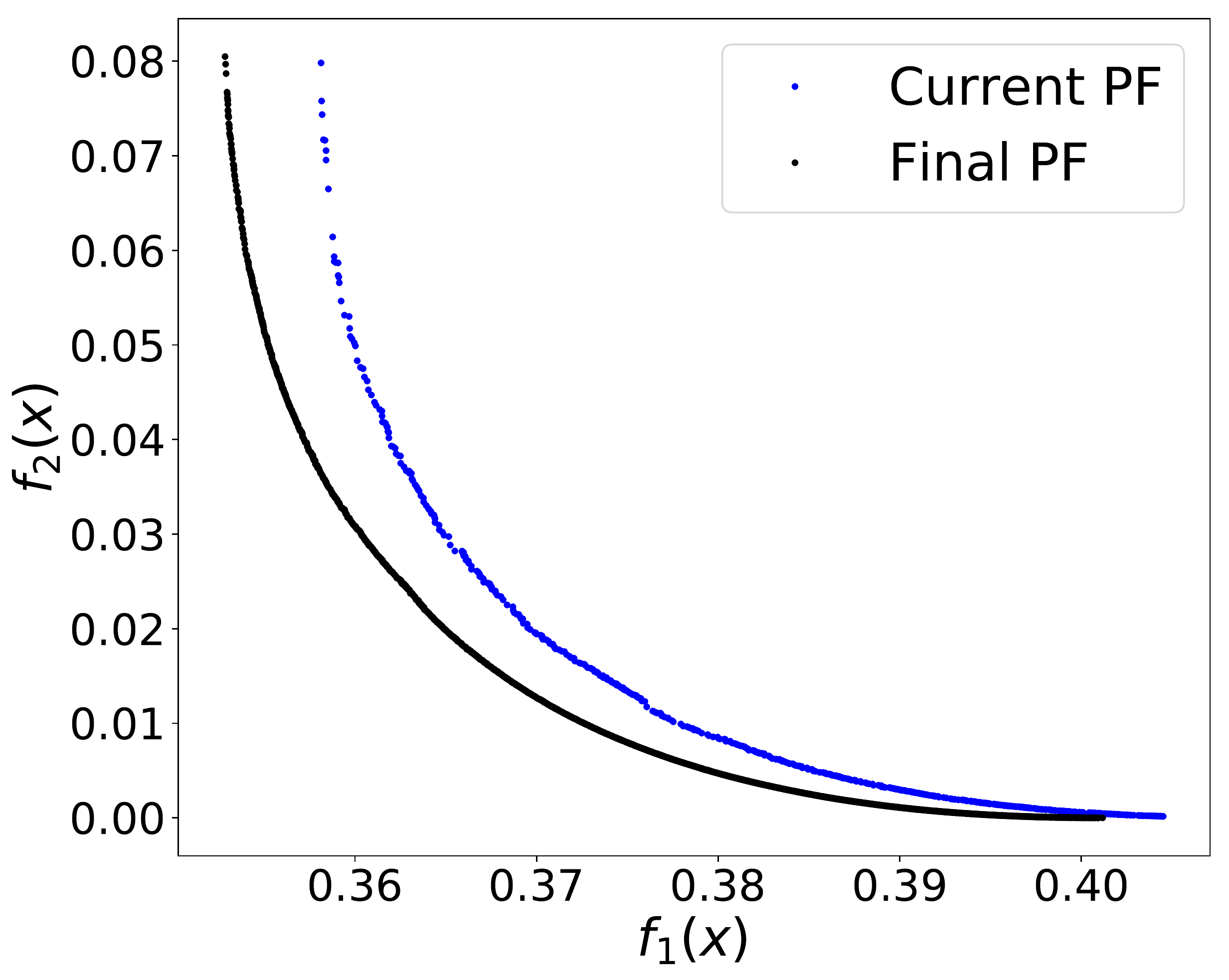}}
  \subfloat[6,000 samples.]{\includegraphics[width = 0.33\textwidth]{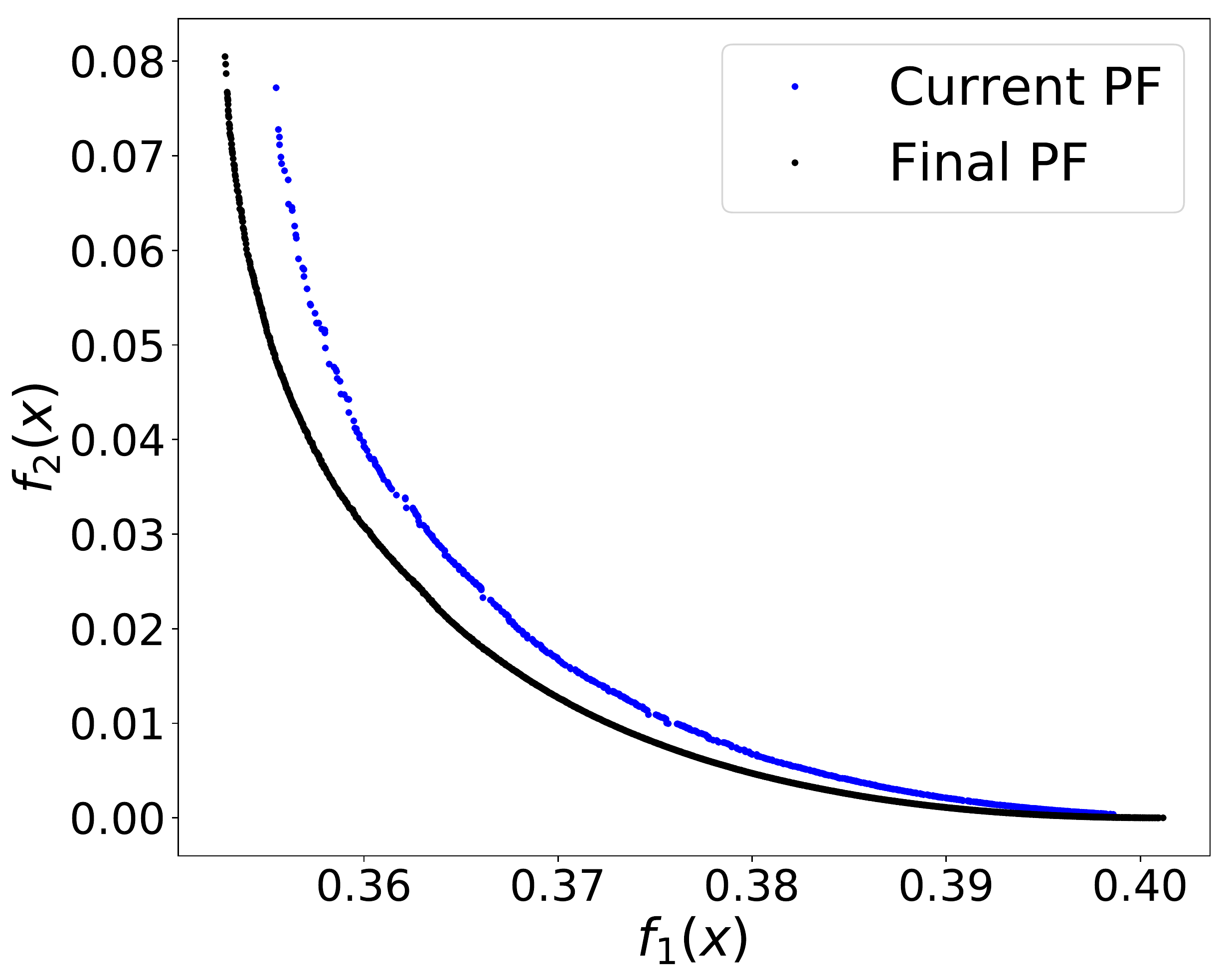}} \\
  \subfloat[8,000 samples.]{\includegraphics[width = 0.33\textwidth]{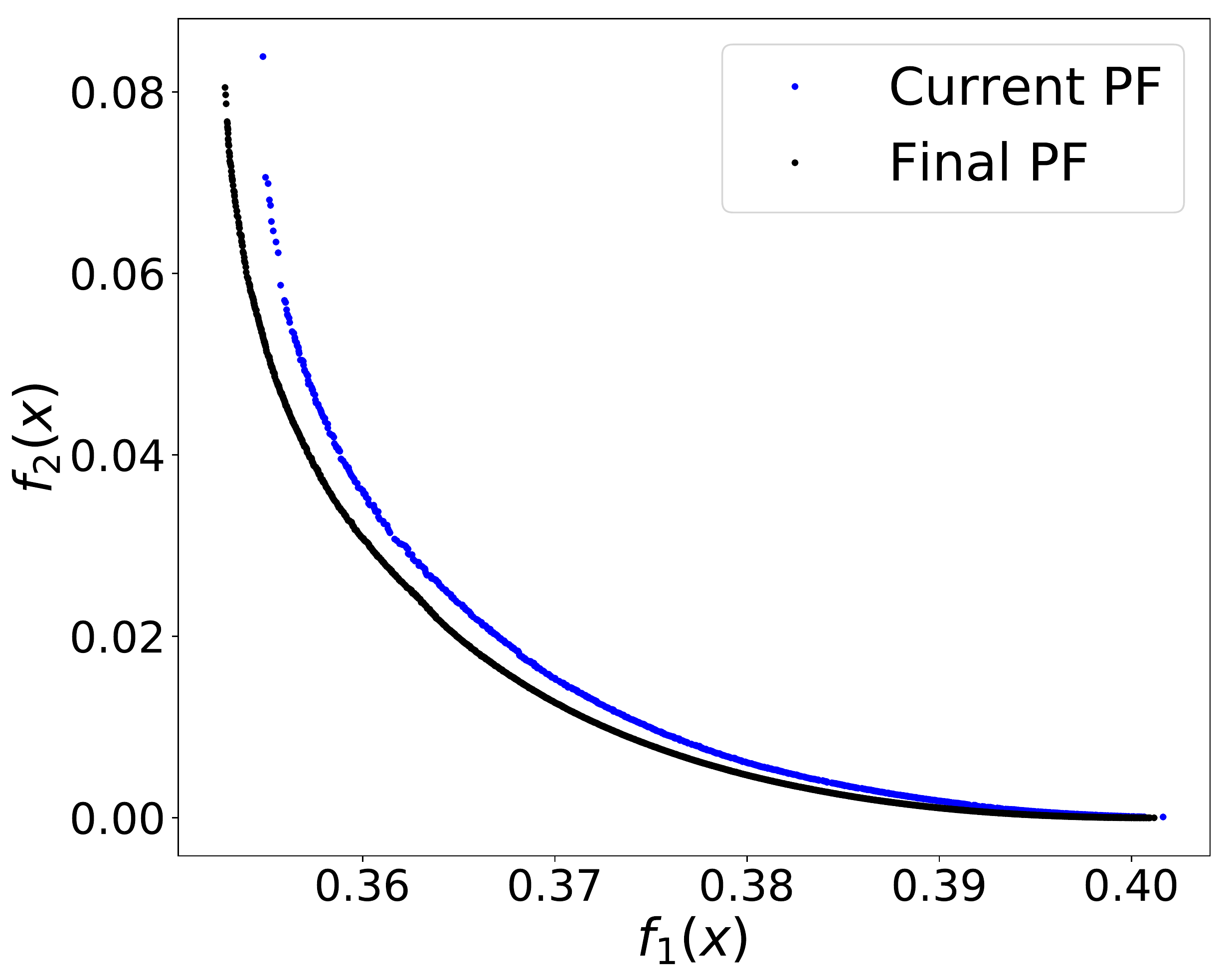}}
  \subfloat[10,000 samples.]{\includegraphics[width = 0.33\textwidth]{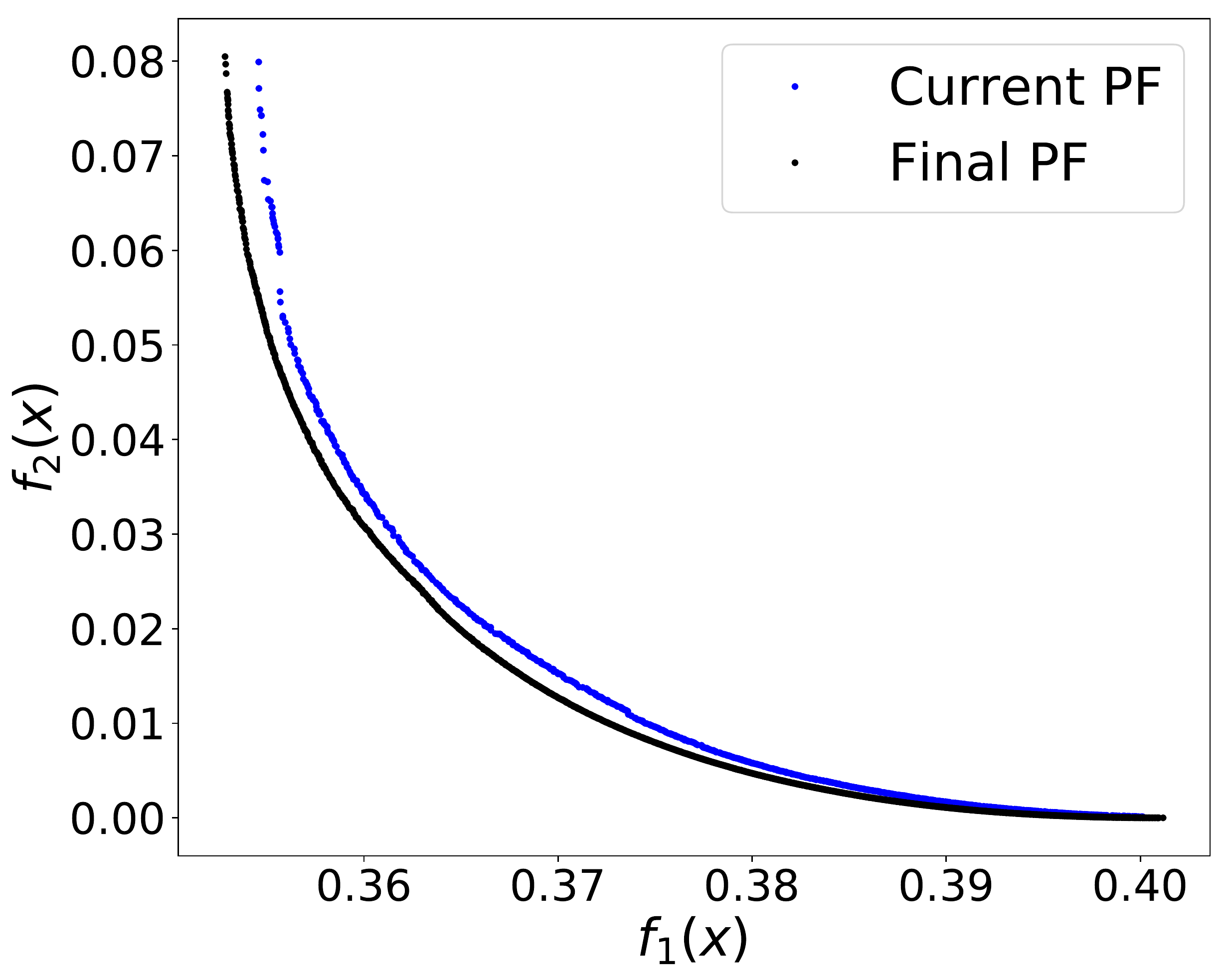}}
  \subfloat[12,000 samples.]{\includegraphics[width = 0.33\textwidth]{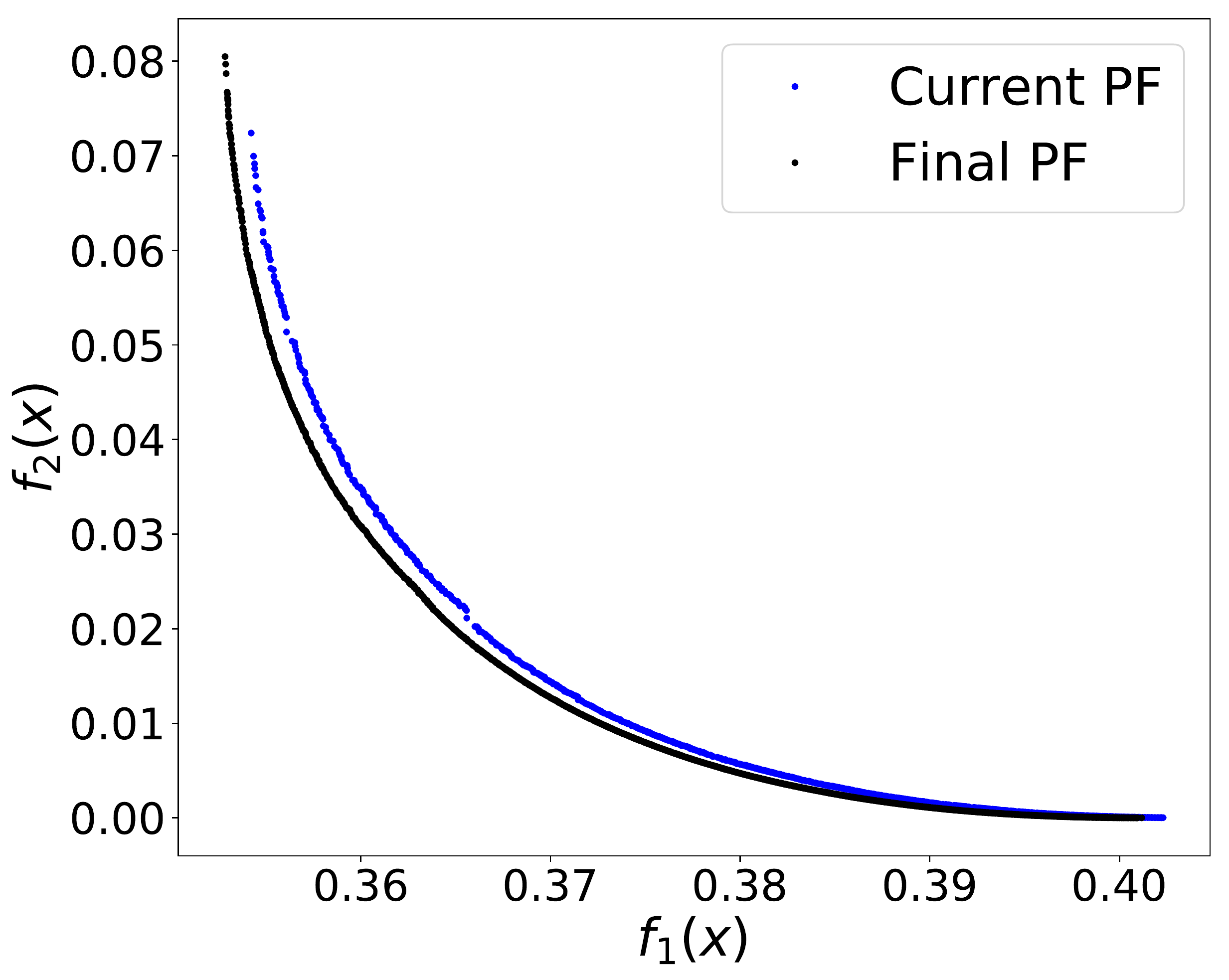}}
  \caption{Updating Pareto fronts using streaming data\label{res:Adult_gender_stramingData}.}
\end{figure}

\end{document}